\documentclass[11pt]{amsart}
\usepackage[margin=1in]{geometry}   %
\usepackage{adjustbox}
\usepackage{amsmath,amssymb,bm}   %
\usepackage{graphicx}
\usepackage{booktabs}
\usepackage{multirow}
\usepackage{subcaption}
\usepackage{xcolor}
\usepackage[numbers,sort&compress]{natbib}
\usepackage{hyperref}
\usepackage{siunitx}
\usepackage{enumitem}
\graphicspath{{figs/}}
\hypersetup{colorlinks=true,linkcolor=blue!50!black,citecolor=blue!50!black,urlcolor=blue!50!black}

\newlength{\figmaxh}
\newcommand{\figor}[2][\linewidth]{\includegraphics[width=#1,height=\figmaxh,keepaspectratio]{#2}}
\newcommand{\stackh}[1]{\setlength{\figmaxh}{#1\textheight}}

\newcommand{\lrnr}{LRNR}
\newtheorem{theorem}{Theorem}\newtheorem{proposition}[theorem]{Proposition}\newtheorem{corollary}[theorem]{Corollary}
\theoremstyle{remark}\newtheorem{remark}[theorem]{Remark}
\DeclareMathOperator{\dist}{dist}   %
\DeclareMathOperator{\diag}{diag}
\newcommand{\R}{\mathbb{R}}
\newcommand{\mb}{\bm{\mu}}
\newcommand{\bW}{\mathbf{W}}\newcommand{\bU}{\mathbf{U}}\newcommand{\bV}{\mathbf{V}}\newcommand{\bB}{\mathbf{B}}
\newcommand{\bP}{\mathbf{P}}\newcommand{\bD}{\mathbf{D}}\newcommand{\bI}{\mathbf{I}}\newcommand{\bS}{\mathbf{S}}
\newcommand{\bz}{\mathbf{z}}\newcommand{\bb}{\mathbf{b}}\newcommand{\bs}{\mathbf{s}}\newcommand{\bc}{\mathbf{c}}
\newcommand{\bu}{\mathbf{u}}\newcommand{\bv}{\mathbf{v}}\newcommand{\bh}{\mathbf{h}}\newcommand{\br}{\mathbf{r}}
\newcommand{\by}{\mathbf{y}}\newcommand{\bx}{\mathbf{x}}\newcommand{\bzero}{\mathbf{0}}
\newcommand{\bUbar}{\bar{\mathbf{U}}}\newcommand{\bsbar}{\bar{\mathbf{s}}}
\newcommand{\btheta}{\bm{\theta}}\newcommand{\bxi}{\bm{\xi}}\newcommand{\bSigma}{\bm{\Sigma}}

\title{Spatially local mini-batching for implicit neural representations}

\author{Minseok Choi}
\address[M.~Choi]{Pohang University of Science and Technology (POSTECH), Republic of Korea}
\email{mchoi@postech.ac.kr}

\author{Swarup Dhar}
\address[S.~Dhar]{Washington University in St.\ Louis, USA}
\email{d.swarup@wustl.edu}

\author{Heejae Kwon}
\address[H.~Kwon]{Pohang University of Science and Technology (POSTECH), Republic of Korea}
\email{heejaekwon@postech.ac.kr}

\author{Kookjin Lee}
\address[K.~Lee]{Arizona State University, USA}
\email{klee263@asu.edu}

\author{Donsub Rim}
\address[D.~Rim]{Department of Mathematics, Washington University in St.\ Louis, USA}
\email{rim@wustl.edu}

\date{November 16, 2026}

\begin{document}

\begin{abstract}\sloppy
Neural surrogates based on implicit neural representations (INRs) 
are typically trained on mini-batches
spread over the whole domain, since a generic INR loses accuracy when every point of
a batch lies in one small region. We show that low-rank neural
representations (\lrnr{}s), INRs whose weight matrices are products of shared
bases with coefficients emitted by a small hypernetwork, can be trained with spatially local
mini-batches. Each step uses the points of one time snapshot of an adaptive-mesh
numerical solution inside one small cell, with snapshots and cells drawn according to the
refinement of the mesh. We posit that such training 
is enabled by the structural property
of \lrnr{}s, as a local step reaches the shared bases only through the $r$
coefficients of each layer, and keeps one cell from rewriting
the network. By a Kolmogorov-width argument, we show that when restricted to a small cell, the hidden states of LRNRs lie close
to a low-dimensional subspace, which suggests the local steps only affect
low-rank subspaces of the shared bases.
On six hyperbolic-conservation-law problems with solutions computed using adaptive mesh
refinement, one local training recipe reaches 1.1--2.3\% relative $L^2$ error with a tenth of the
parameters of a dense network of the same width. 
Finally, we provide a comparison by training a hypernetwork-SIREN, a DeepONet,
and a Fourier neural operator (FNO) of the same size with the same local batches
and training budget. We observe that these baselines reach 16--78\% where the
\lrnr{} reaches 4.9\%. Interestingly, we also observe that local training has a
significant effect on the per-cell low-dimensional structure even in a wide
generic hypernetwork-SIREN.
\end{abstract}

\maketitle

\section{Introduction}
\label{sec:intro}

Neural surrogates for the solutions of partial differential equations are commonly trained by
stochastic gradient methods on mini-batches, random subsets of the training data whose loss
estimates the full loss. For implicit neural representations (INRs), also called coordinate
networks, the data are point samples of the solution, and the mini-batch 
usually draws its points
independently and uniformly over the physical domain. Physics-informed networks draw collocation points
uniformly at random or take them from a fixed grid \citep{raissi2019physics,lu2021deepxde}, INRs
fitted to data draw a random subset of the sample points \citep{sitzmann2020siren}, and operator
networks take as input whole fields or random points scattered over them
\citep{lu2021deeponet,li2021fno}. Each such batch informs the network about every region of the
domain at every step.

For solutions with localized features, such as shocks, contact discontinuities or thin
roll-ups, the common adaptive refinement strategies keep the batch spread over the domain and change the position of
the points: residual-based adaptive sampling, for instance, moves collocation points toward
large residuals \citep{lu2021deepxde,wu2023comprehensive}. However, whenever locality is used, it is usually used
in the architecture rather than the batch. For example, domain-decomposition networks assign a separate
network to each subdomain \citep{jagtap2020xpinns,moseley2023fbpinns}, and grid-based encodings
give each region of the domain its own parameters \citep{muller2022instant}.

\begin{figure}[t]
\centering\figor{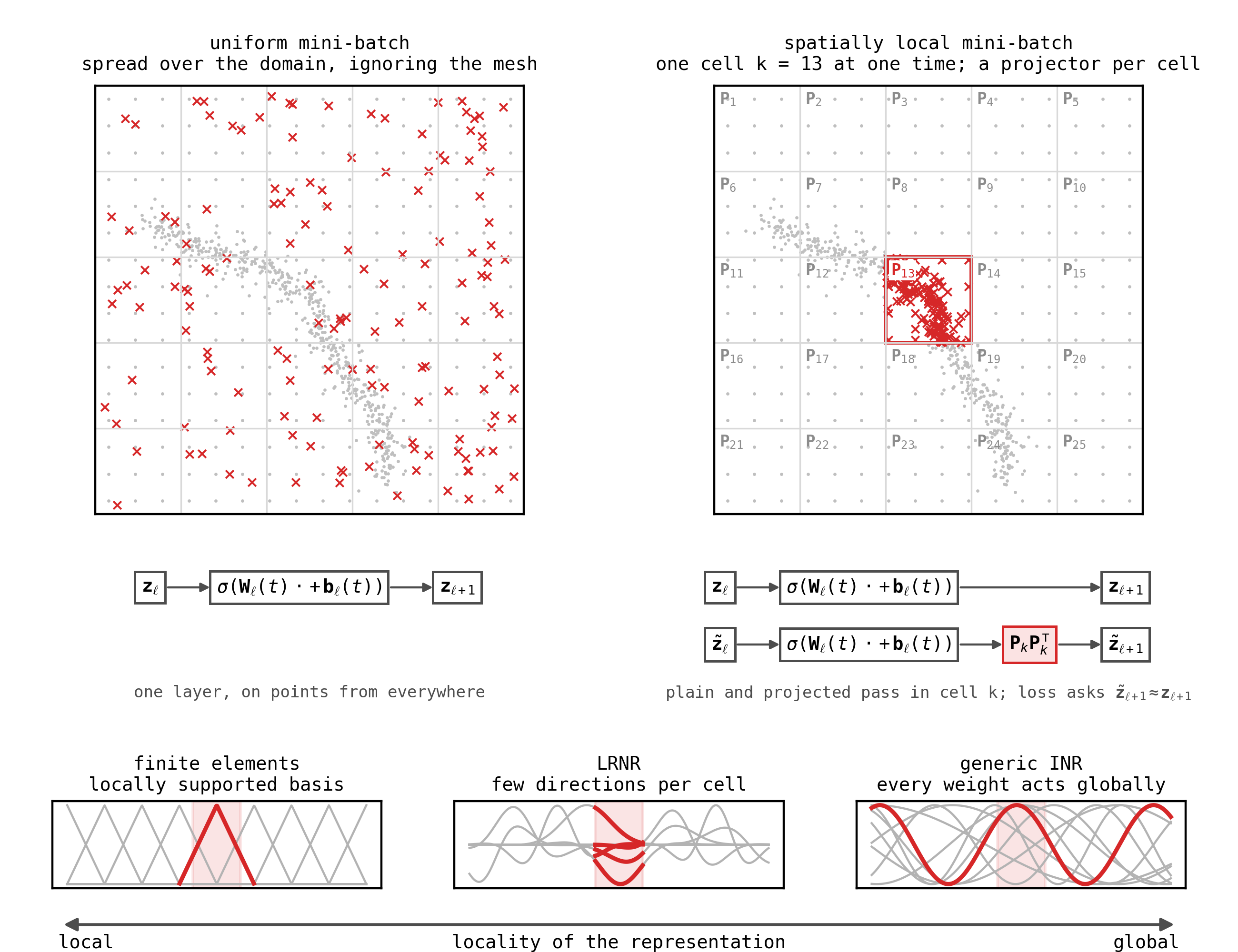}
\caption{Top: the sampled points (crosses) of a uniform mini-batch spread evenly over the
domain and ignore the adaptive mesh (gray dots, left), while a spatially local mini-batch takes
all the points of one time snapshot inside one cell, here $k=13$; each of the 25 cells has
its own projector $\bP_1,...\,,\bP_{25}$ (right). Below
each, one layer of the computation: the plain layer on the uniform batch, and on the local batch
the plain layer together with a twin pass that projects the layer's output with
$\bP_k\bP_k^\top$ after the activation, the loss asking the two outputs to agree. Bottom: locality of the
representation, from finite elements, whose basis functions have local support, through the
\lrnr{}, whose hidden states are close to a few directions within each cell, to a generic INR,
in which every weight acts on the whole domain; in each panel the highlighted curves are basis functions
that are active in the shaded cell.}
\label{fig:intro}
\end{figure}

In this work, we study a spatially local mini-batch that is drawn in two stages: first a small region of
the domain at one time (here, a cell of a tiling, in one time snapshot of the solution), then
all the sample points inside it (Figure~\ref{fig:intro}, top row). Averaged over the draw of regions, its loss is an unbiased
estimate of a weighted full loss, as in importance-sampled stochastic gradient descent
\citep{needell2016sgd}, so local mini-batching optimizes the same objective as uniform
mini-batching. The points of each step, however, are spatially local and correlated, and a single gradient contains information about one region only. The closest practice we know of is
in graph learning, where Cluster-GCN draws each mini-batch from one cluster of a partitioned
graph \citep{chiang2019clustergcn}. For a single INR that covers the whole domain, this form of
mini-batching is rarely used, to the best of our knowledge.

The low-rank neural representation (\lrnr{}) is an INR $u(\bx;\mb)$, a multilayer perceptron 
that maps a point of the domain to the conserved variables, whose weight
matrices are rank-$r$ products $\bW_\ell(\mb)=\bU_\ell \diag(\bs_\ell(\mb)) \bV_\ell^{\top}$ of
shared bases and coefficient vectors $\bs_\ell(\mb)\in\R^r$ produced by a small hypernetwork. The bases are learned once, and the \lrnr{} depends on the parameters $\mb$ of the problem only through the $2rL$ coefficients. In general $\mb$ can be a collection of any physical parameters; in this paper we focus on
the case in which $\mb=(t)$ is time alone. A trained \lrnr{} is therefore a time-dependent curve in an
$r$-dimensional family of networks.

We train LRNRs with spatially local mini-batches. We show that local
mini-batching is effective for the \lrnr{}, and we give an argument for why its hidden states are
locally low-dimensional. 
Generic architectures are less accurate under local training. Three standard architectures, hypernetwork-SIREN \cite{sitzmann2020siren}, DeepONet \cite{lu2021deeponet}, 
and Fourier neural operator (FNO) \cite{li2021fno}, trained with the
same local batches and the same budget end at least a factor of three above the \lrnr{}, which
we attribute to each local step changing weights that the rest of the domain depends on
(Section~\ref{sec:baselines}). 
Global batches remain the more accurate per step for every model, the \lrnr{}
included. However, the \lrnr{}'s discrepancy in accuracy between global and
local training stays roughly constant with respect to the width, whereas for the
wider hypernetwork-SIREN the gap between the two grows. That is, under 
spatially local batches the \lrnr{} loses less accuracy than the baselines.

Local mini-batches may be useful in settings that an accuracy comparison at a fixed number of steps
does not capture. Solver output is often stored and produced by region, as the patches of an
adaptive mesh or the subdomains of a parallel solver, and a local batch reads one local region
whereas a global batch needs points from the whole domain. Since the hidden states of a cell are
low-dimensional, a local step might also be computed with a reduced per-cell network. Although we did not measure these costs, which is left as an open problem (Section~\ref{sec:discussion}), these advantages can potentially outweigh the loss in accuracy per step.

In numerical analysis, locality in the representation of spatial fields is a nearly universally accepted principle, where it is built into the basis (Figure~\ref{fig:intro}, bottom row). 
A finite element consists of a cell, a finite-dimensional space 
of shape functions supported on the cell, and a set of nodal variables, such as point values. These variables are unisolvent for the shape functions: the values of the nodal variables determine a linear combination of the shape functions uniquely 
\citep{ciarlet2002finite,brenner2008mathematical}. Thus, each global degree of freedom belongs to a
few cells, and the basis function assigned to it is supported on those cells only. The data on
one cell therefore determine the representation there, and changing them does not change the field further away. This locality of the representation is also tied closely to the locality of the
differential operators: a derivative at a point depends only on the field near that point, so a
discretized operator couples the neighboring degrees of freedom only.

An INR has no analogous structure. Physics-informed networks pair
a local operator with a global representation, since the residual at a point is local but its
gradient reaches every weight, and their training difficulties are well documented
\citep{wang2021understanding,krishnapriyan2021characterizing}; domain decomposition is one
proposal to amend this mismatch \citep{moseley2023fbpinns}. In an INR, every weight influences the field
globally, so a gradient step
computed on one cell changes the field in all the others. The \lrnr{} does not localize its
weights either; but the hidden states have a local structure. The \lrnr{} can be placed between the two extremes of local and global (see Figure~\ref{fig:intro} for an illustration). Within a cell, the hidden states of each
layer lie close to a subspace of low dimension (Section~\ref{sec:theory}), and a basis of that
subspace, sampled from the trained model, plays the role of the element's space of shape
functions, while the coefficients $\bs_\ell(t)$ and the bases stay
shared by all cells. The FastLRNR reduction (Section~\ref{sec:fast}) uses this structure. For each cell it selects a set of neurons whose values determine the projected hidden
state, an interpolation set that plays the role of unisolvent nodal variables. This turns the cell's subspace into a small local network.

Our data are numerical solutions of hyperbolic conservation laws, systems of the form
$\partial_t u+\nabla\cdot f(u)=0$, whose solutions propagate at finite speeds, and nonlinear conservation laws develop
discontinuities such as shocks even from smooth initial data. The Euler equations of gas
dynamics, Burgers' equation, linear acoustics and advection are examples. We compute the solutions with adaptive-mesh finite-volume solvers
\citep{berger1984adaptive,berger1989local,berger1998amrclaw,leveque2002finite}, whose meshes
follow the structure of the flow, with fine patches at shocks, contacts and
roll-ups and coarse ones where the solution is smooth. The adaptive mesh implicitly indicates where the solution is difficult to
represent, and our sampling procedure uses this information during training.

\subsubsection*{Origins of the \lrnr{}}
The \lrnr{} architecture originated from efforts to achieve model reduction for
transport-dominated problems. A linear reduced
model approximates the set of solutions by a fixed subspace, and its error is bounded below by
the Kolmogorov $n$-width of that set. For transport and wave problems the width decays slowly,
$\sim n^{-1/2}$ for a moving discontinuity
\citep{ohlberger2016reduced,welper2017interpolation,greif2019decay}, so no reduced basis of small
dimension captures a traveling shock. There are works that remove the transport before applying the reduction:
transformed snapshot interpolation \citep{welper2017interpolation}, 
shifted proper orthogonal decomposition
transport reversal \citep{rim2018reversal}, 
\citep{reiss2018shifted} and registration methods \citep{taddei2020registration}. In another
approach the solution is transformed into coordinates where its dynamics are one-dimensional. The Radon
transform splits a multidimensional hyperbolic problem into one-dimensional ones
\citep{rim2018radon}, in the sense of the translation representation of
Lax and Phillips~\citep{laxphillips1989scattering}; and monotone rearrangement can be used to interpolate between displaced
one-dimensional profiles \citep{rim2018displacement}. Transported
subspaces \citep{rim2023mats} combined such ideas into a nonlinear reduced
model, by imposing low-rankness on the transport map. 
Performance bounds for reduced models of parametric transport \citep{rim2023performance}
quantified what general Lipschitz continuous nonlinear reductions can achieve. 

Neural networks became a focus in model reduction, and early works 
used autoencoders \citep{lee2020model}. 
Implicit neural representations were found to be well-suited to model reduction and operator learning, 
as they represent a field as a
function of the spatial coordinate and can therefore be trained on snapshots from different
discretizations, including adaptive meshes and point clouds
\citep{chen2023crom,pan2023nif}.
The work \citep{rim2024lrnr} showed theoretically that LRNRs can efficiently
represent entropy solutions of hyperbolic conservation laws with arbitrary shock
interactions and rarefactions, revealing a low-dimensional structure in
hyperbolic problems.  Therein \lrnr{}s were shown to generalize reduced models
such as transported subspaces: a \lrnr{} learns a varying transported subspace.
The measure of complexity then changes from the
$n$-width of the solution set to the width of the coefficient space. The number
of LRNR coefficients that depend on the physical parameters, the ``coefficient
width'' \citep{cho2025hyperbolic}, stays small
where the $n$-width decays slowly. Low-rank physics-informed networks with hypernetwork
coefficients \citep{cho2023hyperlrpinn} and FastLRNR \citep{cho2024fastlrnr}, which reduces a
trained \lrnr{} for fast evaluation and sparse backpropagation, followed. 
Then it was shown in \cite{cho2025hyperbolic} that \lrnr{}s can be learned from data alone on one- and two-dimensional
wave problems, including periodic acoustics, periodic Burgers and two-dimensional advection
computed with adaptive mesh refinement in Clawpack \cite{clawpack}, training each network on whole time
snapshots with a loss over all spatial points, and observed the low-rank tensor structure and the
interpretable modes that arise in the trained models. The present paper works with the same type
of data but replaces this global training with spatially local training. 

\subsubsection*{Related work}
INRs with sine activations \citep{sitzmann2020siren} represent 
PDE solutions as functions of spatio-temporal variables; hypernetworks \citep{ha2017hypernetworks} are separate networks whose outputs are INRs' weights. The \lrnr{} combines the two
through a low-rank factorization of every weight matrix, a more
strict requirement than the low-rank adaptation
of large models \citep{hu2022lora}; its hypernetwork emits only the coefficients of shared bases.
Operator-learning architectures, the DeepONet \citep{lu2021deeponet} and the FNO \citep{li2021fno}, 
are the natural baselines for a surrogate parametrized by time,
and we compare against them in Section~\ref{sec:baselines}.

Architectures of the same form, with shared bases in every layer and coefficients produced for
each instance, have also been used to adapt a network to a new problem. A graph hypernetwork that
reads the terms and couplings of an equation from an operator graph extends the low-rank
physics-informed networks above to families of PDEs with varying structure, including a
model of a capacitively coupled plasma \citep{jing2026graphhyper}. For structure-preserving
dynamics, a modulation $\bW+\sum_i d_i\bu_i\bv_i^{\top}$ with shared $\bu_i,\bv_i$ and coefficients
$d_i$ from a latent code adapts Hamiltonian and GENERIC networks to a new system from a few
trajectories while preserving their invariants \citep{jing2026metasp}. In computer vision, bases
taken from the singular directions of existing fine-tuned models, with only diagonal coefficients
trained for a new task, give few-shot adaptation with few trainable parameters
\citep{park2026bolt}. In these works the coefficients simply identify an
instance of the INR; in the \lrnr{}, they are related to the physical
parameters (via the hypernetwork).

Some INRs that are local by construction achieve the locality by dividing the domain among many small networks or local
codes: KiloNeRF splits a scene among thousands of tiny networks \citep{reiser2021kilonerf},
local implicit grids and deep local shapes store a latent code per cell
\citep{jiang2020local,chabra2020deep}, and ACORN adapts a block decomposition of the domain to
the signal \citep{martel2021acorn}. For partial differential equations, domain-decomposed
physics-informed networks train a separate network on each subdomain with interface conditions
\citep{jagtap2020cpinn,jagtap2020xpinns,shukla2021parallel}, combine local networks through
partitions of unity or multilevel windows
\citep{moseley2023fbpinns,dolean2024multilevel,kharazmi2021hpvpinn}, learn the partition itself
\citep{lee2021pou}, or couple subdomain networks by Schwarz iterations \citep{li2020d3m}. These
approaches give each region its own parameters; the \lrnr{} keeps a single network and obtains
locality from the structure of its hidden states.

On the reduction side, the per-cell FastLRNR surrogate \citep{cho2024fastlrnr} is a neural analog
of projection-based model reduction with empirical interpolation
\citep{barrault2004eim,chaturantabut2010deim}. Cohen and DeVore~\citep{cohen2016widths}
established the decay of the $n$-width \citep{pinkus1985nwidths} for images of balls under
holomorphic maps; our Theorem~\ref{thm:width} is an elementary, entrywise instance of that
result tailored to activations. The data come from Clawpack's adaptive mesh refinement
\citep{clawpack,mandli2016clawpack,berger1998amrclaw}, and the Euler quadrant problem is that of
Schulz-Rinne, Collins and Glaz~\citep{schulzrinne1993quadrants}.

\subsubsection*{Summary of Contributions}
\begin{enumerate}[nosep]
 \item A Kolmogorov-width theorem for the hidden states of analytic networks
   (Section~\ref{sec:modeltheory}): the width of $\{\sigma(\bU\bc)\}$ over a ball of coefficient
   vectors decays like $\exp(-c n^{1/q})$ in the dimension $n$; inside a cell, with $d$
   parameters, like $\exp(-c n^{1/d})$ at a rate set by the cell size. Projection errors
   propagate through layers with bounded factors, so projecting at every layer is consistent.
 \item A training procedure for \lrnr{}s from adaptive-mesh solver data (Section~\ref{sec:trainresults}): local (frame, cell)
   steps, refinement-driven sampling, an optional per-cell projection penalty, proposed to make the local subspaces explicit, a log-domain
   sparsity penalty on the coefficient spectrum, and a staged schedule with snapshot
   averaging (Section~\ref{sec:training}). One local training recipe fits six different problems to
   1.1--2.3\% relative $L^2$ error (Section~\ref{sec:results}).

 \item An analysis of the trained models (Sections~\ref{sec:analysis} and
 \ref{sec:fast}): a significant-coefficient study showing the trained coefficient rank is small but 
   is fully used and the first layer untruncatable; a hyper-mode decomposition of the
   hypernetwork's output showing one or two modes account for 99\% of its variance, with
   perturbations along them producing physically coherent changes, which extends to locally
   trained models the low-rank mode structure reported for globally trained \lrnr{}s
   \citep{cho2025hyperbolic}. A per-cell reduction
   (FastLRNR) is shown to reproduce the most complicated cell of a problem with 5--21\% of the
   floating point numbers where the cell holds enough solver points, and over-fits the other times.
 \item A controlled comparison (Section~\ref{sec:baselines}) with a hypernetwork-SIREN, a DeepONet and an FNO at matched parameter count, matched data and matched steps under local and global
   batches and two samplers, including width sweeps of the hypernetwork-SIREN and the \lrnr{},
   learning-rate stability, and the dimension of the hidden states per cell.
\end{enumerate}

\section{The low-rank neural representation and its local structure}
\label{sec:modeltheory}

\subsection{The model}
\label{sec:model}

\subsubsection*{Implicit neural representation}
The \lrnr{} architecture we will use throughout is a family of INRs $u(\bx;\mb):\R^{n_{\rm in}}\to\R^{n_{\rm out}}$ with $L$ layers, each hidden layer containing $M_\ell=M$ neurons, and a linear output layer:
\begin{align}
 \bz_0 &= \phi(\bx), \quad
 \bz_{\ell+1} = \sigma_\ell\!\left(\bW_\ell(\mb) \bz_\ell + \bb_\ell(\mb)\right),\quad \ell=0,...\,,L-1,
 \quad u = \bz_L.
\label{eq:trunk}
\end{align}
In this work, we use $\sigma_\ell(\cdot)=\sin(\omega_\ell \cdot)$ for the hidden layers and the identity for the
output layer. The first hidden layer uses a higher frequency $\omega_0$, and in other layers $\omega_\ell=1$.

\subsubsection*{Input features}
In this work, we focus on data with two spatial dimensions,
but the coordinates are lifted to Chebyshev polynomial features of total degree 1 to $p$ on the domain mapped to $[-1,1]^2$: $\phi(\bx)\in\R^{n_{\rm in}}$ (all products $T_i(x)T_j(y)$ with $1\le i+j\le p$, so $n_{\rm in}=(p+1)(p+2)/2-1$). Chebyshev polynomials were introduced to provide the high-frequency sine layer with a smooth low-frequency input; other orthogonal polynomials or Fourier features would serve the same purpose. In our experiments the lift improved the accuracy on some problems but not on others (Appendix~\ref{sec:design}); the models reported here use $p=9$. Time is not an INR input; it influences the network only
through the coefficients.

\subsubsection*{Low-rank, parameter-dependent weights}
Every weight matrix and bias is a rank-$r$ expansion in shared bases,
\begin{align}
 \bW_\ell(\mb) = \bU_\ell \diag\!\big(\bs_\ell(\mb)\big) \bV_\ell^{\top},\quad
 \bb_\ell(\mb) = \bB_\ell \bs^{\rm b}_\ell(\mb),
\label{eq:lowrank}
\end{align}
with column bases $\bU_\ell\in\R^{M_{\ell+1}\times r}$, row bases $\bV_\ell\in\R^{M_\ell\times r}$, bias bases
$\bB_\ell\in\R^{M_{\ell+1}\times r}$ and coefficient vectors $\bs_\ell(\mb),\bs^{\rm b}_\ell(\mb)\in\R^r$. A hidden layer stores $3Mr$ basis entries in place of a dense $M\times M$ block and its bias.

\subsubsection*{Hypernetwork}
The coefficients are the output of a small multilayer perceptron,
$(\bs_\ell,\bs^{\rm b}_\ell)_{\ell=0}^{L-1}=f_{\rm hyper}(\mb)\in\R^{2\times r\times L}$, with $\tanh$ hidden units and
either a softplus or an identity output (Table~\ref{tab:hyper}, Appendix~\ref{sec:design}). For the time-only examples $\mb=(t)$, so the hypernetwork is a map from one number to $2rL$ coefficients.

\subsubsection*{Per-cell projection penalty (optional)}
During training every hidden layer also has, for each cell $k$ of a two-level tiling of
the domain, an orthonormal projector $\bP_{k,\ell}\in\R^{M \times r_s}$. A twin forward
pass replaces the hidden state by $\bP_{k,\ell}\bP_{k,\ell}^{\top}\bz_\ell$ inside cell $k$ and the training
loss asks, at a small weight, that the output not change (Section~\ref{sec:training}). The
projectors are not part of the model at evaluation time. Its purpose is to make the per-cell subspaces of Section~\ref{sec:theory} explicit during training: projectors that adapt to each cell would give local steps a per-cell target and provide the cell bases that the FastLRNR reduction of Section~\ref{sec:fast} constructs after training. In our experiments the penalty is met mostly through directions shared by all cells, with a weak cell-adapted component, and the accuracy does not depend on it (Section~\ref{sec:results}); we report it as a proposal, and setting $\lambda_s=0$ gives plain local training.

\subsection{A Kolmogorov-width argument for local projections}
\label{sec:theory}

The training procedure in Section~\ref{sec:training} and the FastLRNR reduction in
Section~\ref{sec:fast} rely on a structural property of LRNRs: 
inside a cell of the domain, the hidden
states of a trained \lrnr{} lie close to a subspace of $\R^M$ whose dimension is much smaller compared to the
width $M$. This section gives an elementary explanation. A hidden state
is an analytic activation applied to a rank-$r$ pre-activation, and the image of a small ball
under an analytic map is, up to an exponentially small error, contained in a polynomial
subspace whose dimension is set by the size of the ball and the rank, not by the width. We aim at a
simple statement to demonstrate the concept, rather than a sharp estimate.

\subsubsection*{Setting}
Fix a hidden layer and drop its index. Its pre-activation at a point $\bx$ and time $t$
is, by \eqref{eq:lowrank},
\begin{equation}
 \bW(t) \bz + \bb(t) = \bU \diag(\bs(t)) \bV^{\top}\bz(\bx,t) + \bB \bs^{\rm b}(t)
 = \bUbar \bc(\bx,t),
 \quad \bUbar = [\bU \mid \bB]\in\R^{M\times q},\quad q = 2r,
\label{eq:Uc}
\end{equation}
with the \emph{coefficient vector}
$\bc(\bx,t) = \big(\diag(\bs(t))\bV^{\top}\bz(\bx,t);\ \bs^{\rm b}(t)\big)\in\R^{q}$:
every pre-activation the layer ever produces is a combination of the $q=100$ columns of $\bUbar$.
The hidden state is $\sigma(\bUbar\bc)$ with $\sigma$ applied entrywise ($\sigma=\sin(\omega \cdot)$
in the \lrnr{}). Write $\bu_i^{\top}$ for the rows of $\bUbar$. For a set $S_c\subset\R^q$ of coefficient
vectors with a center $\bc_0$, call
\begin{equation}
 \rho = \sup_{\bc\in S_c}\ \max_{i}\ |\bu_i^{\top}(\bc-\bc_0)|
\label{eq:radius}
\end{equation}
its \emph{pre-activation radius}: the largest excursion of any neuron's argument from its
value at the center. The Kolmogorov $n$-width of a bounded set $S\subset\R^M$ is
\begin{equation}
 d_n(S) = \inf_{\dim L = n}\ \sup_{\bs\in S} \dist(\bs, L),
\end{equation}
the error of the best $n$-dimensional linear approximation of $S$, that is, the error of the best projector of
rank $n$. Since the entries of a sine network's hidden state are of size one, $d_n(S)/\sqrt{M}$
reads as a per-entry (that is, relative) error.

\begin{theorem}[Image of a ball under an analytic activation]
\label{thm:width}
Let $\sigma$ be real-analytic on an interval containing all pre-activations, let
$S_c\subset\R^q$ have center $\bc_0$ and pre-activation radius $\rho$, and let
$S=\{\sigma(\bUbar\bc): \bc\in S_c\}\subset\R^M$. For $K\ge0$ let $\tau_K(\rho)$ bound the degree-$K$
Taylor remainder of $\sigma$ over the relevant arguments,
\begin{equation}
 \Big|\sigma(z_0+h)-\sum_{k=0}^{K}\frac{\sigma^{(k)}(z_0)}{k!}
    h^k\Big|\ \le\ \tau_K(\rho)
 \quad\text{for all } |h|\le\rho,
 \quad z_0\in\{\bu_i^{\top}\bc_0\}.
\end{equation}
For $n\ge1$ let $K(n)=\max\{k\ge0:\binom{q+k}{q}\le n\}$ be the largest degree whose monomials
in $q$ variables fit in $n$ dimensions. Then
\begin{equation}
 d_n(S) \le\ \sqrt{M} \tau_{K(n)}(\rho).
\label{eq:widthbound}
\end{equation}
For $\sigma=\sin(\omega \cdot)$, $\tau_K(\rho)\le(\omega\rho)^{K+1}/(K+1)!$. For $\sigma$
analytic and bounded by $C_\sigma$ on the complex disc of radius $\varrho>\rho$ around every $z_0$,
$\tau_K(\rho)\le C_\sigma (\rho/\varrho)^{K+1}/(1-\rho/\varrho)$.
\end{theorem}
The proof is in Appendix~\ref{sec:proofs}.

\begin{remark}[Explicit dependence on $n$]
\label{rem:explicit}
A closed form in $n$ can be derived from \eqref{eq:widthbound}. By the
arithmetic-geometric mean inequality,
$\binom{q+k}{q}=\frac{1}{q!}\prod_{j=1}^{q}(k+j)\le\frac{1}{q!}\big(k+\tfrac{q+1}{2}\big)^{q}$,
so every $k\le(q! n)^{1/q}-\frac{q+1}{2}$ has its monomials fit in $n$ dimensions. So we have
\begin{equation}
 K(n)\ \ge\ \Big\lfloor (q! n)^{1/q}-\tfrac{q+1}{2}\Big\rfloor
\end{equation}
whenever the right-hand side is non-negative. With Stirling's lower bound
\begin{equation}
 (K+1)!\ \ge\ \sqrt{2\pi(K+1)} \Big(\frac{K+1}{e}\Big)^{K+1}
\end{equation}
and $m=K(n)+1$, the two cases of the theorem read
\begin{align}
 \sigma=\sin(\omega \cdot):\quad&
 d_n(S)\ \le\ \sqrt{M} \frac{1}{\sqrt{2\pi m}}\Big(\frac{e \omega\rho}{m}\Big)^{m},
\label{eq:widthsine}\\
 \sigma\ \text{analytic, } |\sigma|\le C_\sigma \text{ on discs of radius } \varrho>\rho:\quad&
 d_n(S)\ \le\ \sqrt{M} \frac{C_\sigma}{1-\rho/\varrho}\Big(\frac{\rho}{\varrho}\Big)^{m},
\label{eq:widthanalytic}
\end{align}
with $m\ge(q! n)^{1/q}-(q+1)/2$. The width therefore decays at least like
$\exp(-c n^{1/q})$ in $n$, faster than any power of $n$ for fixed $q$, and for
the sine faster still, like $\exp(-n^{1/q}\log n^{1/q})$ up to constants. The
two inequalities are nearly tight for $K$ large compared with $q$, and cost
about one Taylor degree at the small $K$ of the illustration below.
\end{remark}

\begin{remark}[Scope of the theorem]
\label{rem:global}
The dimension count depends on the rank $q=2r$ and on the Taylor degree, but not on
the width; $\sqrt{M}$ is only the norm of an $M$-vector. But the number of monomials in $q=100$ variables is large ($\binom{q+k}{q}=101$, $5151$, $176{,}851$ for $k=1,2,3$), so $K(n)=1$ for
$101\le n<5151$ and the explicit form \eqref{eq:widthsine} gives no information for any $n$ below the
width (i.e., $(100! n)^{1/100}<50$ for $n\le M$). With $\omega=1$ the theorem implies that $101$
dimensions capture the hidden states to a per-entry error $\rho^2/2$.
Over the whole domain the pre-activations of a sine layer span several radians,
($\rho\gtrsim1$) so the hidden states of the trained
models over the full domain are not necessarily low-rank. The bound becomes nontrivial when $q$ is
a small number and $\rho$ is small (Corollary~\ref{cor:cell}).
\end{remark}

\begin{corollary}[A cell over time]
\label{cor:cell}
Suppose that inside a cell $\Omega_k$ of the domain the coefficient vector is an analytic
function of $d$ parameters, $\bc=\bc(\btheta)$, $\btheta\in\Theta_k\subset\R^d$. Then the proof of
Theorem~\ref{thm:width}, applied to the composite functions $g_i(\btheta)=\sigma(\bu_i^{\top}\bc(\btheta))$
expanded in $\btheta$ about the cell center $\btheta_k$, gives for every $n\ge1$
\begin{equation}
 d_n(S_k)\ \le\ \sqrt{M} \tilde\tau_{K_d(n)},\quad
 K_d(n)\ \ge\ \Big\lfloor (d! n)^{1/d}-\tfrac{d+1}{2}\Big\rfloor,\quad
 S_k=\{\sigma(\bUbar\bc(\btheta)):\btheta\in\Theta_k\},
\end{equation}
where $\tilde\tau_K$ bounds the degree-$K$ Taylor remainder in $\btheta$ of the $g_i$ over
$\Theta_k$: monomials in the $d$ parameters replace monomials in the $q$ coefficients. If the
$g_i$ are analytic on a polydisc of radius $\varrho_k$ about $\btheta_k$ and $\Theta_k$ has
radius $h_k<\varrho_k$, then $\tilde\tau_K\lesssim(h_k/\varrho_k)^{K+1}$ and
$d_n(S_k)\lesssim\sqrt{M} (h_k/\varrho_k)^{(d! n)^{1/d}-(d-1)/2}$: exponential decay in
$n^{1/d}$ at a rate set by the cell size.
\end{corollary}

\begin{remark}[Time-dependence]
For the time-dependent \lrnr{} the parameters are $\btheta=(x,y,t)$, so $d=3$: the coefficient vector
$\bc(\bx,t)=(\diag(\bs(t))\bV^{\top}\bz(\bx,t);\bs^{\rm b}(t))$ is analytic in
$\bx$ through the Chebyshev features and the sine layers below, and in $t$ through the
hypernetwork. The monomial count is then modest, e.g., $\binom{3+k}{3}=4,10,20,35,...\,,120,165,220,286$
for $k=0,...\,,9$ (i.e., $K_3(n)\approx(6n)^{1/3}-2$), and the remainder decays geometrically in the ratio of the cell size to
the analyticity radius of the layer. For a sine layer of frequency $\omega$ that radius is of
order $1/\omega$ in the directions along which the pre-activation changes, so a cell of width
$h$ yields a factor of order $\omega h$ per Taylor degree. For example, in the first hidden
layer ($\omega_0=30$) and a fine cell of width $0.1$ of the unit square, $\omega h\approx3$;
Lagrange's remainder $3^{k+1}/(k+1)!$ is $0.4\%$ at $k=10$ ($\binom{13}{3}=286$ dimensions)
and $10^{-5}$ at $k=14$ ($680$ dimensions). Note that the inequality
\eqref{eq:widthsine} of Remark~\ref{rem:explicit} with $q$ substituted by $d=3$
gives the same decay. 

The rank-$400$ projectors of the training procedure and the FastLRNR cell bases of $67$--$300$ directions that agree with the full model to $0.5$--$1.5\%$ (Section~\ref{sec:fast}) roughly illustrate this.
The interior layers have $\omega=1$ and bounded inputs, so their local radius is smaller and
fewer dimensions suffice; the first layer, the one with the large frequency, is also the layer
whose coefficients cannot be truncated (Section~\ref{sec:coef}). The 
the coefficient trajectory in $t$-direction is nearly affine in $t$
(Section~\ref{sec:hyper}), so the $t$-dependence of $\bc$ is close to a single
polynomial degree.
\end{remark}

\subsubsection*{Layers and the projection $\bV_{\ell+1}^{\top}$ of the next layer}
Two facts make the layer-wise picture consistent. Let $\bz=\sigma(\bUbar\bc)$ be the hidden state of
layer $\ell$ and let the next layer act on it through its row basis $\bV'=\bV_{\ell+1}\in\R^{M\times r}$,
its coefficients $\bD=\diag(\bs_{\ell+1}(t))$ and its stacked basis $\bUbar'=[\bU_{\ell+1} \mid \bB_{\ell+1}]$,
so that the next coefficient vector is $\bc'=(\bD\bV'^{\top}\bz; \bs^{\rm b}_{\ell+1})$ and the next hidden
state $\bz'=\sigma(\bUbar'\bc')$.

\begin{proposition}[Propagation through a layer]
\label{prop:layers}
Let $\sigma$ be $\omega$-Lipschitz ($|\sigma'|\le\omega$).
\begin{enumerate}[nosep,label=(\alph*)]
 \item \emph{The local radius propagates.} At fixed $t$, for two points $\bx,\bx'$ of a cell,
 \begin{equation}
  \|\bc'(\bx,t)-\bc'(\bx',t)\|_2\ \le\ \omega \|\bD\bV'^{\top}\|_2 \|\bUbar\|_2
     \|\bc(\bx,t)-\bc(\bx',t)\|_2 ,
 \end{equation}
 so a cell that is small for one layer is small for the next, up to the norms of the trained
 bases.
 \item \emph{Projection errors propagate at the same rate, and only $\bV'^{\top}\bz$ is relevant.} If $\bz$ is
 replaced by $\bP\bz$ for an orthogonal projector $\bP$ with $\|(\bI-\bP)\bz\|_2\le\varepsilon$, the next
 hidden state changes by at most
 \begin{equation}
  \|\bz'-\tilde\bz'\|_2\ \le\ \omega \|\bUbar'\|_2 \|\bD\|_2 \|\bV'^{\top}(\bI-\bP)\|_2 \varepsilon ,
 \end{equation}
 and in particular by zero if the range of $\bP$ contains the $r$ columns of $\bV'$.
\end{enumerate}
\end{proposition}
The proof is in Appendix~\ref{sec:proofs}.

By (b), if each layer's hidden state is projected onto its own local subspace with error $\varepsilon_\ell$, the output error is at most $\sum_\ell\big(\prod_{j>\ell}\omega_j\|\bUbar_j\|_2\|\bD_j\bV_j^{\top}\|_2\big)\varepsilon_\ell$, a sum with bounded factors, so projecting at every hidden layer is consistent. The existence
of the local subspace (Theorem~\ref{thm:width}, Corollary~\ref{cor:cell}) makes the hidden
state reconstructible from a few of its entries. The FastLRNR reduction uses this by evaluating only a selected subset of a layer's neurons in a cell,
reconstructing the cell's hidden state through the local basis, and passing on $\bV'^{\top}$ of the
reconstruction; (b) bounds how the reconstruction error travels through the remaining
layers.

\subsection{Existence of local projections}
\label{sec:existence}
These results apply to any network with analytic activations. In every cell of a tiling fine enough that the
pre-activation radius of a layer is small, the layer's hidden states over the cell and over
the whole time range lie within a distance of a subspace of dimension $n$ that decays like $\exp(-c n^{1/d})$;
the dimension is determined by the cell size relative to the layer's analyticity scale
($\omega h$ for a sine layer) and by the number of parameters the coefficients depend on,
not by the width. A spatially local projector of small rank therefore exists at every hidden
layer, and projecting at every layer at once costs a sum of bounded terms. The same argument
applies to the hypernetwork-SIREN and DeepONet baselines in Section~\ref{sec:baselines}, which
are analytic networks too. What distinguishes \lrnr{} is the rank-$r$ coefficient structure
\eqref{eq:Uc}, which makes every layer's pre-activations a $q$-dimensional family and gives the
local subspaces a concrete form.
The FastLRNR reduction in Section~\ref{sec:fast} is
the constructive counterpart.

\subsection{Local projection in the training procedure}
\label{sec:assume}
The optional projection penalty of Section~\ref{sec:model} is based on this structure. It uses a
two-level tiling of the domain into cells, a projector $\bP_{k,\ell}$ of rank $r_s$ for every
cell $k$ and every hidden layer $\ell$, and a projected twin pass that asks the network's output
not to change when each hidden state is replaced by its projection. Projectors that captured each cell's subspace would be the training-time counterpart of the FastLRNR bases of Section~\ref{sec:fast}. 
The penalty was introduced to assist its learning, but technically, in 
the \lrnr{} the
penalty can be met without local information, because each layer reads its input
through an $r$-dimensional row basis shared by all cells. In fact, the training procedure we adopted learns projectors that are mostly global, 
and training without the penalty reaches the same accuracy 
(Section~\ref{sec:training}). The orthogonality penalty on the bases is meant to keep the propagation constants of
Proposition~\ref{prop:layers} moderate; at the weight used its pull is weak
(Section~\ref{sec:training}). By Corollary~\ref{cor:cell}, a coarse cell has a larger $h_k$ and needs more dimensions than a fine one, which the two levels of the tiling reflect; randomly shifting the cell boundaries prevents the subspaces from specializing
to fixed between-cell boundaries.

\section{Training and results}
\label{sec:trainresults}

\subsection{Training from adaptive-mesh solver data}
\label{sec:training}

\subsubsection*{Data}
Each problem is one trajectory of a Clawpack/AMRClaw run with three refinement levels,
written at 81 equally spaced times $t_k$ over the given time interval (Table~\ref{tab:examples}). A
frame (snapshot) is the set of cell centers of every patch that is not covered by a finer patch, with
the conserved variables at those centers. The finest patches are located where the solver found
localized features (shocks, contacts, roll-ups), so the point density of a frame is the solver's
own metric of where the solution has details. The network is trained on the scattered points as they are, without interpolation to a grid, and all errors in this paper are measured on
them, with each point weighted by the area of its cell (the square of its nearest-neighbor
spacing), so that the reported relative $L^2$ errors are errors on the domain rather than
on the mesh. Unless stated otherwise, they are errors on the training data, at the solver's points of the training frames; the error at times between the training frames is reported in
Section~\ref{sec:results}.

\subsubsection*{Spatially local mini-batches}
\begin{figure}[tp]
\centering
\figor{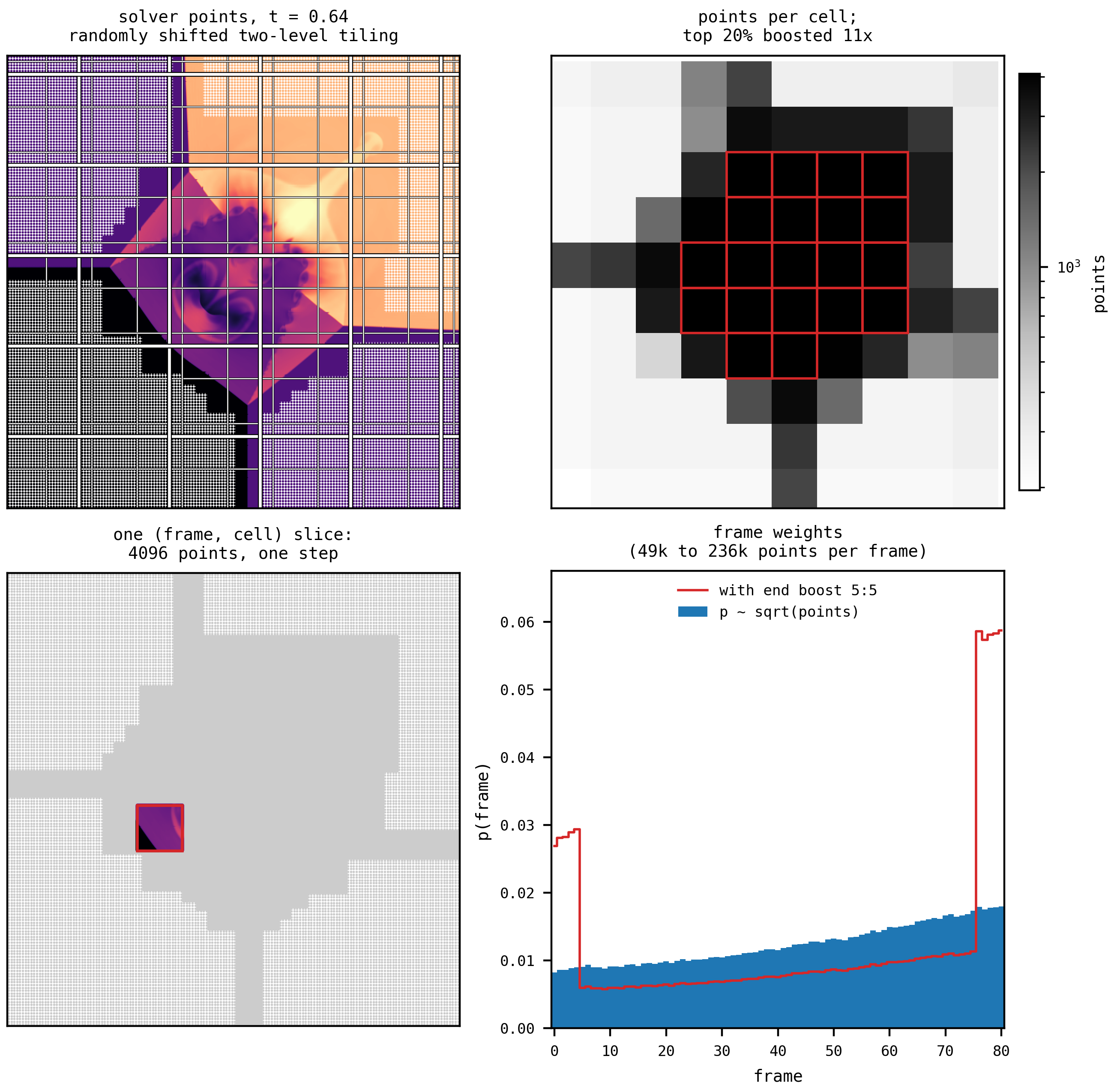}
\caption{The local spatial sampling, on frame 64 ($t=0.64$) of the Euler quadrants.
Top left: the solver's adaptive points (171{,}000 here, colored by density) under the two-level
tiling: 25 coarse cells (thick) and 100 fine cells (thin) on $[0,1]^2$, each level
translated by its own random offset of up to a quarter cell, redrawn every time the tiling is
applied. Top right: the number of solver points in each fine cell (gray, log scale) is the
sampler's score; the cells at or above its 0.8 quantile with at least 100 points (red, 20 of
100) are used in the draw list ten extra times. Bottom left: one step is one such draw, the points of
one frame inside one cell (here 4{,}100 points at the shock--contact junction), on which one
gradient step is taken with that cell's projector resident. Bottom right: the frame is drawn with
probability proportional to $\sqrt{N_k}$ (blue; 49{,}000 to 236{,}000 points per frame) and,
in the final stages, five times more often within five frames of either end (red).}
\label{fig:sampling}
\end{figure}

A step of the training procedure is one gradient step on the points of one frame that fall into one
cell of a tiling of the domain. The tiling has two levels, $n\times n$ fine and $(n/2)\times(n/2)$ coarse square cells. Each level is translated by its own random offset of up to a fraction $\delta$ of a cell every time the tiling is applied, so that the cell boundaries move, and neighboring projectors overlap. The two levels give two scales of local step. A fine cell isolates the most refined features with a small batch, and a coarse cell (four fine cells) exposes the network to the same features together with their neighborhood, across fine-cell boundaries; the random shifts keep the cell boundaries from becoming fixed seams in the model. A spatially local mini-batching step is given by:
\begin{enumerate}[nosep]
 \item draws a frame (snapshot) with probability proportional to $\sqrt{N_k}$, where $N_k$ is the
   number of points in frame $k$, and a cell from a list that contains every cell once and
   the cells above the $q$-quantile of the per-cell point count (with at least $N_{\min}$ points) $\beta$ extra times, so that the solver's most refined cells are visited $\beta+1$ times as often as the others;
 \item takes the frame's points inside the cell (cached per (frame, cell); the cache is
   cleared every 500 steps so that the random shifts are refreshed);
 \item[(2-1)] (with the projection penalty) loads that cell's projectors into the active parameters (one set is resident at a
   time);
 \item runs the plain forward pass and, with the projection penalty, the projected twin pass;
 \item forms the loss \eqref{eq:loss} and steps two Adam optimizers \citep{kingma2015adam}: one on the global
   parameters (bases and hypernetwork) and, with the projection penalty, one, at a tenth of the rate, on the resident projector.
\end{enumerate}
The frame draw is tempered ($\sqrt{N_k}$ vs. $N_k$) so that the late, richly
refined frames of a problem with developing structure are visited more often; e.g., in the final stages of the three scalar and acoustic problems the
first and last five frames are drawn $\beta_{\rm end}$ times more often still, because the error of the
plain ladder was concentrated at the two ends of the time range (Section~\ref{sec:results}).
Figure~\ref{fig:sampling} shows the ingredients on one frame of the Euler problem.

\subsubsection*{Loss and regularizers}
For the sampled slice with points $\{\bx_i\}$ and solver values $\{\by_i\}$,
\begin{align}
 \mathcal{L} &= \underbrace{\tfrac1n\sum_i \|u(\bx_i)-\by_i\|^2}_{\text{misfit}}
 + \lambda_s \underbrace{\tfrac1n\sum_i \|u_P(\bx_i)-u(\bx_i)\|^2}_{\text{projection misfit}}\nonumber\\
 &\quad + f(e)\Big[\lambda_{\rm sp} \mathcal{R}_{\rm sparse} + \lambda_{o} \mathcal{R}_{\rm ortho}
 + \lambda_{o,s} \mathcal{R}_{\rm ortho,P}\Big],
\label{eq:loss}
\end{align}
where $u_P$ is the output of the twin pass with every hidden state replaced by its projection; the weights are listed in Table~\ref{tab:hyper}, and $\lambda_s=\lambda_{o,s}=0$ gives plain local training.
\begin{itemize}[nosep]
 \item $\mathcal{R}_{\rm ortho}=\frac{1}{r^2L}\sum_\ell\big(\|\bU_\ell^{\top}\bU_\ell-\bI\|_F^2+\|\bV_\ell^{\top}\bV_\ell-\bI\|_F^2+\|\bB_\ell^{\top}\bB_\ell-\bI\|_F^2\big)$
   penalizes the departure of the bases from orthonormality; $\mathcal{R}_{\rm ortho,P}$ is the same
   for the resident projector, which it keeps orthonormal.
 \item $\mathcal{R}_{\rm sparse}$ asks the coefficient spectrum of every layer to decay
   geometrically along the rank axis:
   \begin{align}
   \mathcal{R}_{\rm sparse}=\mathrm{mean}_{(\ell,\ \bs\text{ or }\bs^{\rm b})}\sum_{j=1}^{r-1}
   \mathrm{ReLU}\!\Big(\log\gamma+\log|s_{j+1}|-\log|s_j|\Big),
   \end{align}
   a penalty on every consecutive pair whose ratio $|s_{j+1}|/|s_j|$ exceeds $1/\gamma$.
   Working in the log domain makes the penalty homogeneous of degree zero in the
   coefficients, so it reorders the spectrum without shrinking it (the plain form collapses
   the output). The ratio $\gamma=D^{-1/(r-1)}$ is set from a total dynamic range $D$ demanded across the spectrum, so the penalty means the same thing at every rank.
 \item $f(e)$ is a schedule for the regularizers: a sinusoidal cycling between 0 and 1 of
   period $T_f$ during the first half of a stage, whose amplitude decays to zero over
   the second half so that the stage ends at $f=1$ and its snapshots are comparable.
\end{itemize}
The projection misfit is not scheduled; it is on at full weight from the first step.

\subsubsection*{Stages, schedules and averaging}
A stage is $S$ steps, a whole number of regularization periods $T_f$. A ladder is a sequence of stages, each initialized from the
previous stage's model, of two kinds: \emph{cosine stages}, learning rate $\eta$ with cosine annealing \citep{loshchilov2017sgdr} to zero (continuation stages restart at a reduced peak $\eta_2$), and \emph{constant-rate stages}, a constant rate $\eta_c$ with a snapshot written every $S_{\rm snap}$ steps, the stage's model being the average of its snapshots \citep{izmailov2018swa}. The ladders switch
from cosine to constant-rate stages once the per-stage gain of the cosine stages falls below
about 5\%, and where several constant-rate stages fluctuate around a floor, the snapshots of all
of them are averaged into one model. Snapshot averaging lowers the error by 20\% once a ladder has stopped descending and raises it while the ladder still descends (Appendix~\ref{sec:design}).

\subsubsection*{Hyperparameters}
Table~\ref{tab:hyper} lists the settings shared by all experiments; with them the INR stores $1.20\times10^6$ basis entries as 32-bit floating point numbers (4.8~MB), a tenth of the $1.2\times10^7$ of a dense network of the same widths; the stage ladders and
training histories of the final models are in Appendix~\ref{sec:recipe}.
\begin{table}[h]
\centering\small
\begin{tabular}{@{}lp{0.72\linewidth}@{}}
\toprule
Setting & Value \\
\midrule
INR & $L=5$ layers, width $M=2000$, $\sin(\omega \cdot)$, $\omega_0=30$, $\omega_{\ell>0}=1$ \\
Rank & $r=50$ \\
Features & Chebyshev products $T_i(x)T_j(y)$, $1\le i+j\le9$ ($p=9$, $n_{\rm in}=54$) \\
Hypernetwork & $2\times128$ identity ($3\times5$ softplus, Euler) \\
Tiling & $n=10$: 25 coarse + 100 fine cells, random shift $\le$ a quarter cell ($\delta=1/4$) \\
Projector rank & $r_s=400$ ($800$ for periodic acoustics) \\
Sampler & Frames $\propto\sqrt{N_k}$; cells above the $q=0.8$ quantile with at least $N_{\min}=100$ points $\times11$ ($\beta=10$); end frames $\times5$ ($\beta_{\rm end}=5$) (final stages, all but Euler) \\
Stage & $S=$ 97{,}400 steps ($=1000\pi\times31$), regularization period $T_f=6283$, snapshots every $S_{\rm snap}=10^4$ steps \\
Loss weights & $\lambda_s=10^{-3}$, $\lambda_{\rm sp}=10^{-3}$, $\lambda_o=10^{-7}$, $\lambda_{o,s}=10^{-2}$, $D=10^3$ \\
Learning rates & $\eta=10^{-4}$, continuation peak $\eta_2=5\times10^{-5}$ cosine / $\eta_c=$ $10^{-5}$ constant \\
\bottomrule
\end{tabular}
\caption{Settings shared by all experiments.}
\label{tab:hyper}
\end{table}

\subsection{Problems, accuracy and learned solutions}
\label{sec:results}

\begin{table}[h]
\centering\small
\begin{tabular}{@{}p{0.235\linewidth}>{\raggedright\arraybackslash}p{0.335\linewidth}rrrr@{}}
\toprule
Problem & Equations, domain, horizon & \# of Vars. & Stages & Error & Pre-boost \\
\midrule
Euler quadrants & Euler, $[0,1]^2$, $t\le0.8$ & 4 & 20 & 1.1\% & n/a \\
Periodic acoustics & Acoustics, $[-1,1]^2$ periodic, $t\le0.75$ & 3 & 24 & 1.7\% & 2.5\% \\
Swirling advection & Advection, $[0,1]^2$, $t\le4$ & 1 & 24 & 1.7\% & 2.9\% \\
Periodic Burgers & Burgers, $[0,1]^2$ periodic, $t\le2.5$ & 1 & 23 & 1.5\% & 2.3\% \\
Shock--bubble & Axisymmetric Euler + tracer, $[0,2]\times[0,0.5]$, $t\le0.6$ & 5 & 15 & 1.8\% & 2.7\% \\
Acoustics with inclusions & Variable-coefficient acoustics, $[0,1]^2$, $t\le0.9$ & 3 & 12 & 2.3\% & 3.4\% \\
\bottomrule
\end{tabular}
\caption{The six problems and the area-weighted relative $L^2$ error of the trained models
on the training data, at the solver's points (rms over the times $t_0,t_{20},t_{40},t_{60},t_{80}$). The pre-boost column
is the floor of each ladder before the end frames were over-sampled; the Euler ladder was
trained without that over-sampling, so it has no pre-boost value.}
\label{tab:examples}
\end{table}

\subsubsection*{Problems}
The six problems are listed in Table~\ref{tab:examples}.
\begin{itemize}
 \item \emph{Euler quadrants.} The four-shock Riemann configuration of
   Schulz-Rinne, Collins and Glaz~\citep{schulzrinne1993quadrants}, on a $40^2$ base grid with two levels of $4\times$
   refinement and 81 frames to $t=0.8$. The most difficult region is the Kelvin--Helmholtz roll-up along
   the contact.
 \item \emph{Periodic acoustics.} A ring of pressure launched at $(-0.5,0)$ on the periodic square,
   which focuses, re-expands and crosses its periodic images.
 \item \emph{Swirling advection.} A flat interface wound up to $t=2$ and unwound to $t=4$.
 \item \emph{Periodic Burgers.} A square of $u=1$ that steepens into diagonal shocks and rarefies
   behind them.
 \item \emph{Shock--bubble interaction.} A planar shock at $x=0.2$ travels down an axisymmetric tube
   and hits a bubble of gas ten times lighter than its surroundings. The bubble is compressed,
   its interface rolls up into a vortex ring, and a passive tracer marks the bubble's gas. The axisymmetric Euler equations evolve density, axial and radial momentum and energy, with the
   tracer as a fifth variable, and the domain is rescaled to the unit square for training.
 \item \emph{Acoustics with inclusions.} A pulse enters through the left boundary of a uniform medium
   holding two discs, a slow dense one (impedance 12, sound speed 0.3) and a fast one (impedance
   10, sound speed 1.5). The pulse is partly reflected by the slow disc and focused inside it,
   runs ahead through the fast one, and leaves through the right boundary. The field is zero at
   $t=0$, and the solver's mesh grows from 2{,}500 to 114{,}000 points as the scattered fronts
   spread.
\end{itemize}
Advection, acoustics and Burgers use a $50^2$ or $40^2$ base grid with two further levels of
refinement. The last two problems are adaptive-mesh versions of two problems of the PyClaw
gallery \citep{clawpack,mandli2016clawpack}, run with AMRClaw \citep{berger1998amrclaw}. The
shock--bubble run has an $80\times20$ base grid with two levels of $2\times$ and $4\times$
refinement, flagged on the density and the tracer only, for 13{,}000--35{,}000 points per
frame.

The gallery drives the inclusions problem with a square wave of duration 0.05. We use a cosine
hump of the same peak and width 0.1 at half height,
$u_n(t)=\tfrac12\bigl(1-\cos(\pi t/0.1)\bigr)$ for $t<0.2$. With the square wave the same
recipe reaches 4.5\%, and with a hump of half the width 3.9\%, so the thinness of the pulse, more
than its jumps, sets the difficulty, and we report the problem whose features the solver resolves well.

\subsubsection*{Accuracy}
The Euler model was trained in 20 stages: cosine stages from scratch, a tempered sampler from
stage 13, and three constant-rate stages whose snapshot average is the final model, at 1.1\%.
The other five were trained by one command string without per-problem changes: one cosine
stage, then constant-rate averaged stages, then constant-rate stages that over-sample the end
frames (from stage 17 for acoustics, advection and Burgers, from stage 7 for shock--bubble and
inclusions). The final model is the snapshot average of the last stage or stages. Each ladder
was stopped once it had been flat for three stages, except the inclusions ladder, which was
still gaining 0.1 points per stage.

Over-sampling the end frames lowered the floor of each of these five ladders by a third or more, from 2.5, 2.9,
2.3, 2.7 and 3.4\% to 1.7, 1.7, 1.5, 1.8 and 2.3\% for acoustics, advection, Burgers,
shock--bubble and inclusions (Table~\ref{tab:examples}). All six models end between 1.1 and
2.3\%. The sampled-cell misfit falls by a factor of 2--17 over a ladder, most of it in the
first three stages; the training histories are in Appendix~\ref{sec:recipe}.

\begin{figure}[tp]
\centering
\figor{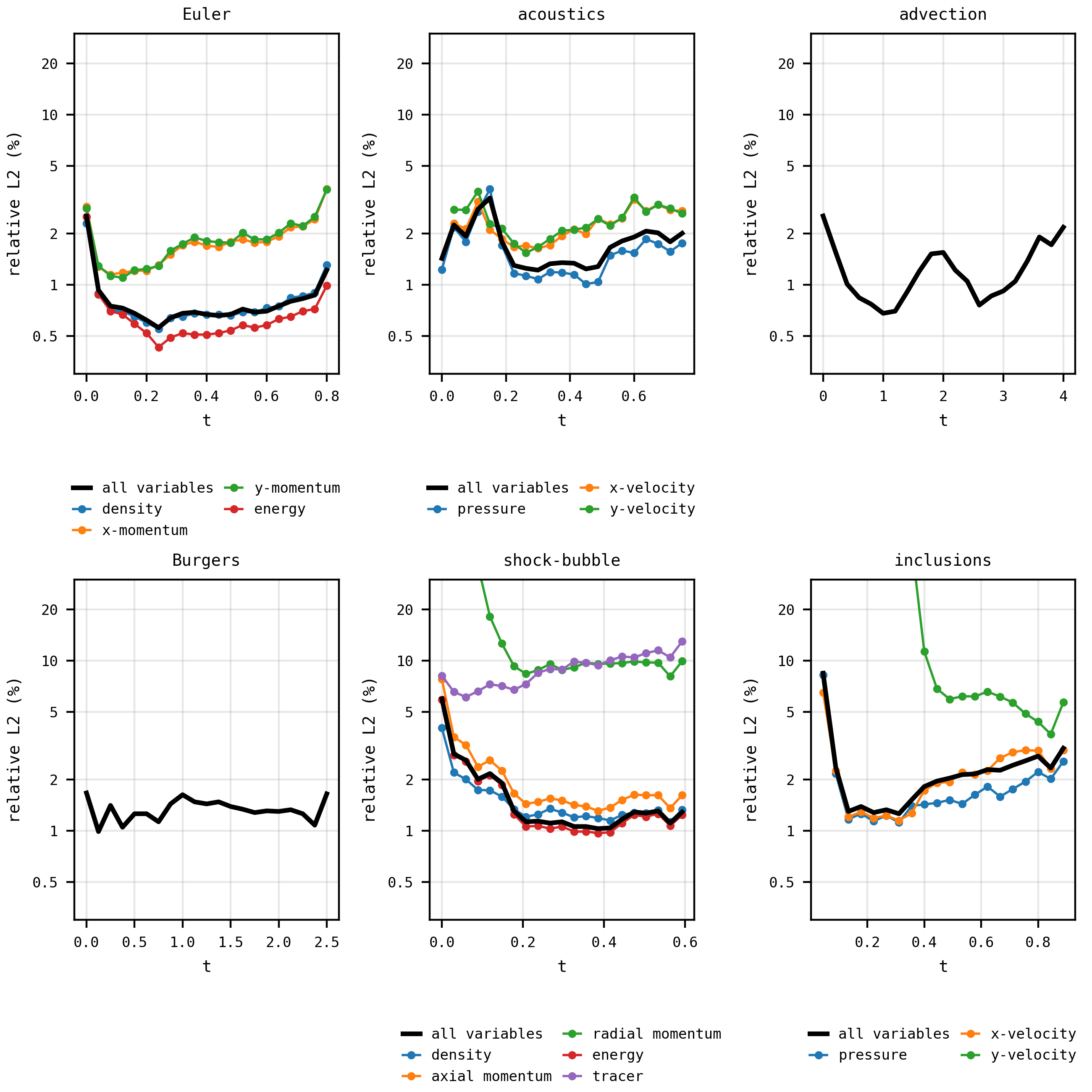}
\caption{Relative $L^2$ error by time of the six final models, at every fourth frame: all
variables together (black) and, for the problems with several variables, each variable. A
variable's error is left out at the times where the variable itself is (nearly) zero: the
radial momentum before the shock reaches the bubble, the inclusions' fields at $t=0$ and their
$y$-velocity before the pulse reaches the discs.}
\label{fig:pertime}
\end{figure}

The per-time error of the six final models (Fig.~\ref{fig:pertime}) shows where the lower limit is.
\begin{itemize}
 \item \emph{Euler, acoustics, advection and Burgers.} The interior times are at 0.6--2\%, with one
   excursion to 3.2\% at the acoustic focus ($t\approx0.15$), and the end points $t=0$ and
   $t=T$ at 1.2--2.5\%, down from 3--6\% before the end frames were over-sampled. Per
   variable, the Euler density and energy are at 0.4--1.3\% and the two momenta, which are
   small over much of the domain, at 1--3.6\%.
 \item \emph{Shock--bubble.} 6\% at $t=0$, where the incoming shock and the bubble are
   discontinuities on a quiet background, then 1.0--1.3\% once the shock has entered the
   bubble. The density, axial momentum and energy are at 1--1.6\%. The two small or thin variables have larger errors. The radial momentum, small next to the axial, is at about 10\%, and the
   tracer, a thin rolled-up sheet by the end, grows from 6\% to 13\%.
 \item \emph{Inclusions.} The only model whose error grows with time. After 8\% while the pulse is
   entering, it is at 1.2--1.4\% while the pulse crosses the uniform medium and rises to
   2--3\% once both discs have scattered it and the reflections fill the domain. The pressure
   and the $x$-velocity follow the total; the $y$-velocity, which only the scattering creates,
   is at 4--7\%.
\end{itemize}

\subsubsection*{Error between training times}
\begin{table}[h]\centering
\begin{tabular}{@{}lrr@{}}\toprule
Problem & 81 training times & 80 halfway times \\\midrule
Euler quadrants & 0.8\% & 0.8\% \\
Periodic acoustics & 1.8\% & 1.8\% \\
Swirling advection & 1.3\% & 1.6\% \\
Periodic Burgers & 1.4\% & 1.8\% \\
Shock--bubble & 1.4\% & 1.4\% \\
Acoustics with inclusions & 2.1\% & 2.1\% \\
\bottomrule\end{tabular}
\caption{Relative $L^2$ error of the final models (area-weighted, root mean square over the frames) at the 81 training times and at the 80 times halfway between them. The halfway solutions were computed by restarting the solver from the training run at the preceding training time and advancing it by half an output interval.}\label{tab:heldout}\end{table}
To measure the error away from the training times, we computed reference solutions at the 80
times halfway between the training frames and evaluated the final models there without retraining
(Table~\ref{tab:heldout}). Each halfway solution was obtained by restarting the solver from a
checkpoint of the training run at the preceding training time and advancing it by half an output
interval, so that it differs from the training run only in its last time step. A rerun of the
solver with twice the output frequency is not a suitable reference: in the periodic acoustics
problem every output interval is a single time step on the coarsest level, at a Courant number near
0.9, and halving every step adds numerical dissipation that changes the solution by up to 7\% at
the final time. The halfway error equals the error at the training times on the Euler, acoustics,
shock--bubble and inclusions problems and is 0.3 and 0.4 points higher on advection and Burgers.
The models are therefore about as accurate between the training times as at them.
Section~\ref{sec:baselines} repeats its comparisons at these times.

\subsubsection*{Learned solutions}
Figures~\ref{fig:story_euler}--\ref{fig:zoom_inclusions} (acoustics, advection, Burgers and
shock--bubble in Appendix~\ref{sec:morefigs}) show the solver's field and the \lrnr{}'s side
by side at four times, with their difference, on the full domain and inside a window on the
most demanding region. The solver is drawn as the nearest adaptive-mesh point on a $600^2$
uniform grid, so its panels have the solver's own resolution. The \lrnr{} is evaluated on the
same grid and, in the window, on its own $600^2$ uniform grid, so the zoom is not a crop of the
coarse view. The difference is the model's error at the solver's points, transferred to the grid by the same nearest-point map, and the percentage above each error panel is the area-weighted relative
$L^2$ error of the shown variable at those points.
\begin{itemize}
 \item \emph{Euler.} The model reproduces the four constant states at $t=0$ with the discontinuities in
   place (2.3\% of the density, almost all of it in a thin band along the initial jumps), is at
   0.6--0.7\% at $t=0.27$ and $0.53$ and at 1.3\% at the final time. The outer shocks, the
   reflected shocks and the contact are at the solver's positions throughout. Inside the
   window, the Kelvin--Helmholtz rolls at $t=0.53$ and the train of billows along the slip
   line with its secondary rolls at $0.80$ are present with their shapes and spacing. The error
   there is 0.8--1.5\%, a speckle the
   size of the finest solver cells in the billow cores and along the shock.
 \item \emph{Inclusions.} In Fig.~\ref{fig:story_inclusions}, the pressure is at 1.2--1.6\% until
   $t=0.69$ and rises to 2.6\% at the end, as the pulse transmitted through the fast disc runs
   ahead of the plane front and the reflections from both discs cross. The window on the slow,
   dense disc shows the part of the pulse that enters it and focuses inside, at 1.6--2.1\% once
   the pulse has arrived; the difference there is a fine ring pattern on the focusing front and
   its reflections.
 \item \emph{Acoustics.} In Fig.~\ref{fig:story_acoustics} (Appendix~\ref{sec:morefigs}), the four times shown are at 1.1--1.8\%. The largest error occurs at the focus, $t=0.15$ (3.6\%, Fig.~\ref{fig:pertime}), the sharpest feature of the trajectory, which falls between them. The crossings of the periodic images are followed closely, and the difference in
   the window, at 0.8--1.3\%, is a faint set of rings, the fronts being a few solver cells
   thinner or thicker.
 \item \emph{Advection.} The errors are largest at the end states, 2.5\% at $t=0$, where the interface has
   zero thickness, and 2.2\% at $t=4$, where the solver's own numerical diffusion has smeared
   it a little, with 0.8\% at the two times between.
 \item \emph{Burgers.} 1.3--1.7\% at every time, the difference a thin line along the moving shock, with no visible difference in the rarefaction.
 \item \emph{Shock--bubble.} Figure~\ref{fig:story_shockbubble} (Appendix~\ref{sec:morefigs}) is drawn at the physical aspect of
   the tube. The density error is 4.0\% at $t=0$, almost all of it on the incoming shock and the
   bubble's edge, and 1.2--1.3\% afterward. The transmitted and reflected shocks, the bubble compressed at
   $t=0.20$ and the vortex ring it rolls into are in place at every time. Inside the window that
   follows the ring the error is 0.9--2.1\% from $t=0.20$, concentrated
   on the ring's thin trailing filament and on the leading shock.
\end{itemize}

\begin{figure}[tp]
\centering
\figor{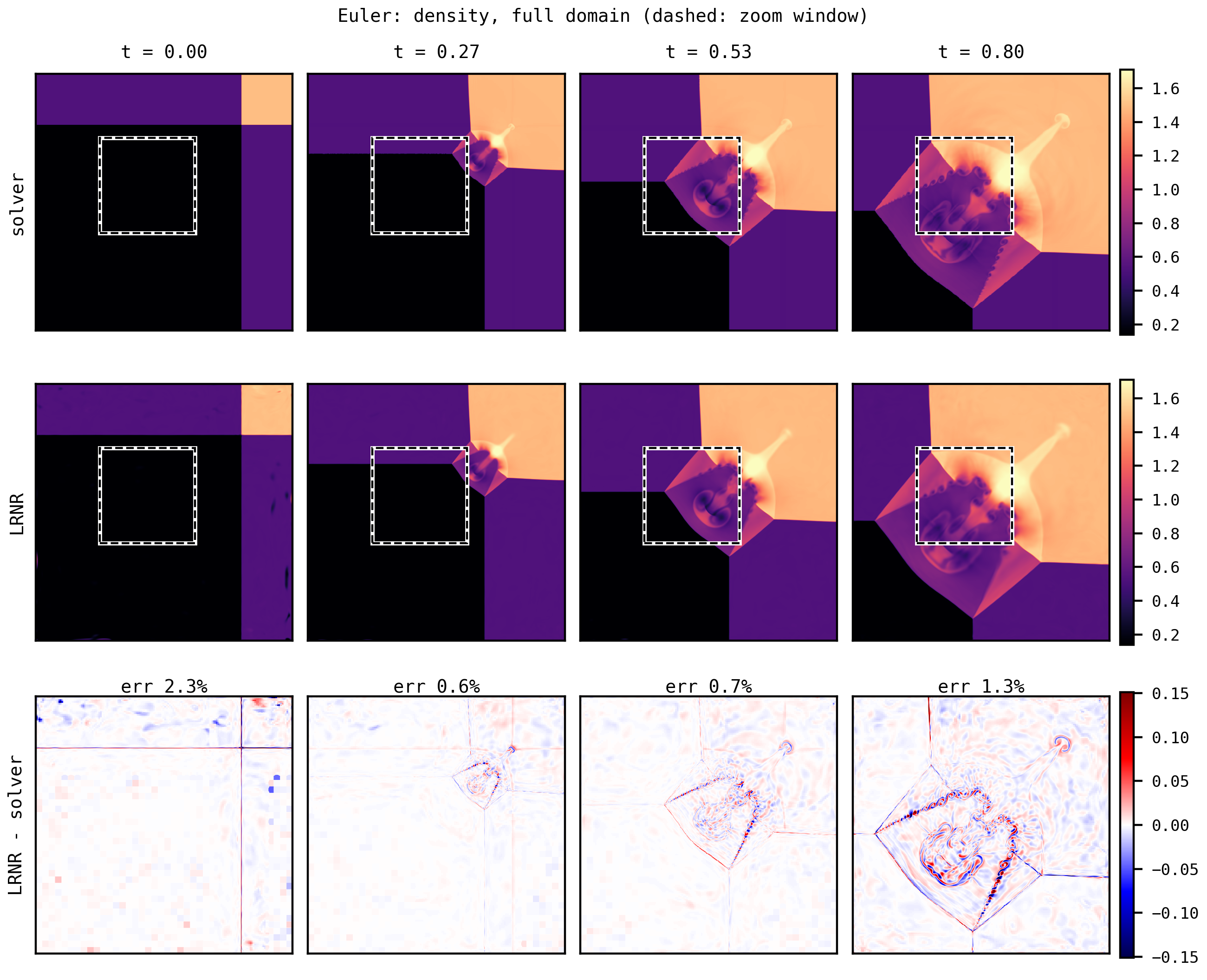}
\caption{Euler quadrants, density: solver (top row), \lrnr{} (middle) and their difference
(bottom) at four times on the full domain. The dashed square is the window of
Fig.~\ref{fig:zoom_euler}.}
\label{fig:story_euler}
\end{figure}

\begin{figure}[tp]
\centering
\figor{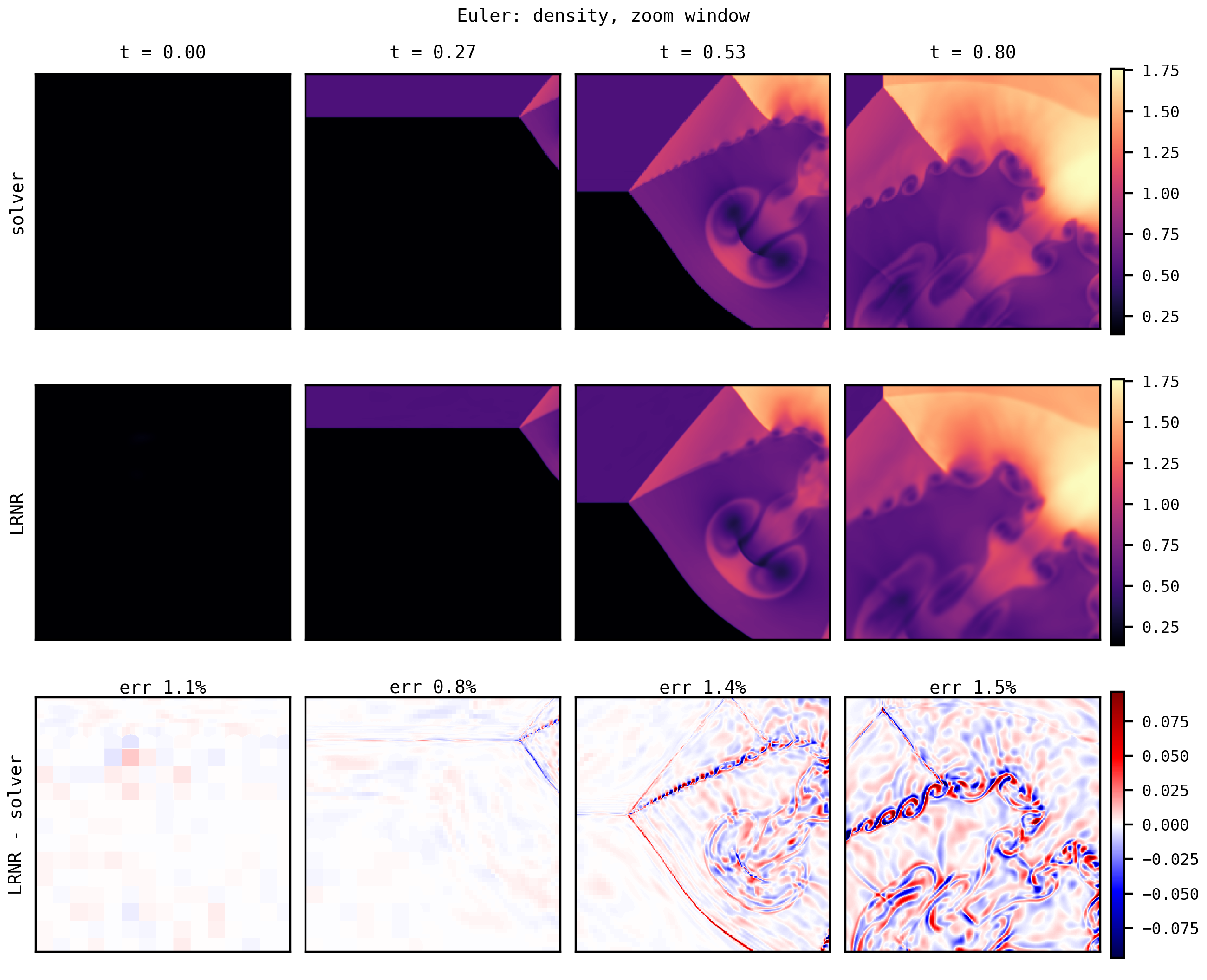}
\caption{As Fig.~\ref{fig:story_euler}, inside the Kelvin--Helmholtz window
$[0.25,0.62]\times[0.38,0.75]$.}
\label{fig:zoom_euler}
\end{figure}

\begin{figure}[tp]
\centering
\figor{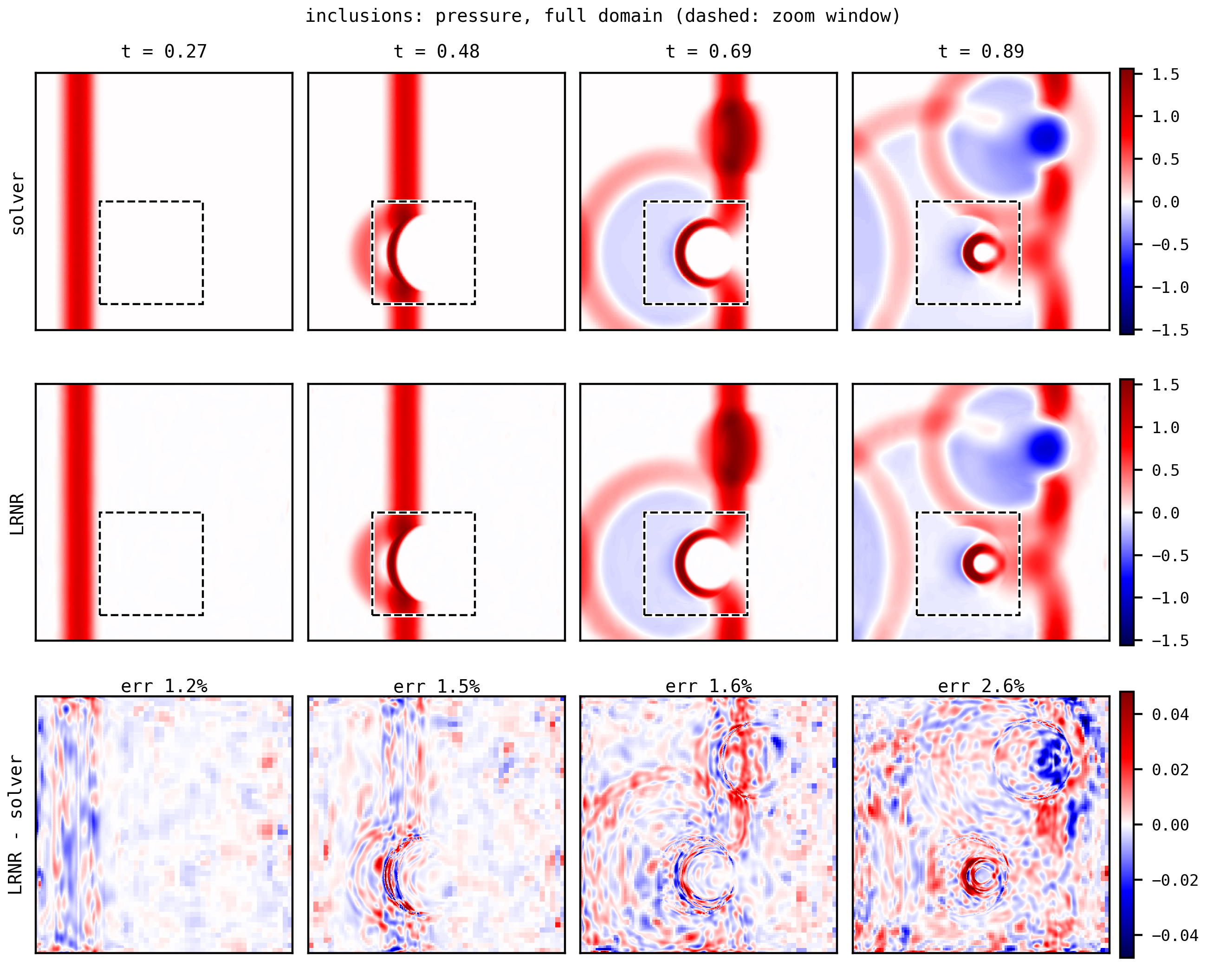}
\caption{Acoustics with inclusions, pressure: as Fig.~\ref{fig:story_euler}, from $t=0.27$
(the field is zero until the pulse arrives).}
\label{fig:story_inclusions}
\end{figure}

\begin{figure}[tp]
\centering
\figor{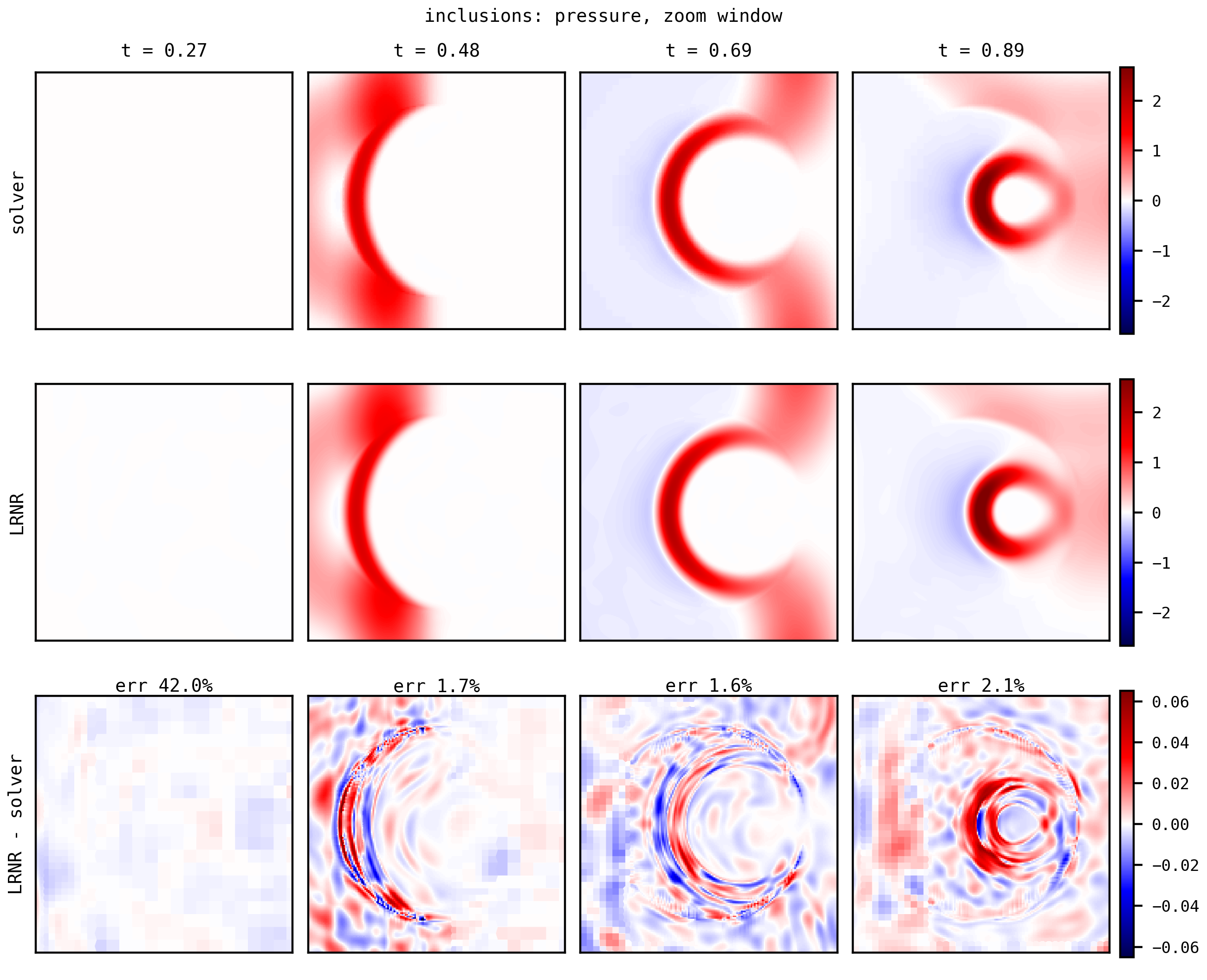}
\caption{As Fig.~\ref{fig:story_inclusions}, inside the window on the slow, dense disc and its
surroundings, $[0.25,0.65]\times[0.1,0.5]$, which the pulse has barely reached at $t=0.27$ (hence
the large relative error of that panel on a near-zero field).}
\label{fig:zoom_inclusions}
\end{figure}

\subsubsection*{Effect of the regularizers}
Neither penalty on the model directly produces the structure that its form describes; 
both nevertheless
affect the trained model measurably. The sparsity penalty targets a geometric decay of the
coefficient spectrum of every layer along the rank axis. This decay does not occur inside the network. The trained models spread their hidden layers over nearly all 50 coefficients but decay in the
output layer (Section~\ref{sec:coef}), and dropping the trailing coefficients by index still raises the error even when the bases are made orthonormal (Appendix~\ref{sec:design}). The scale-invariant logarithmic form is necessary, since the plain form
collapsed the output. A weight of $10^{-3}$ is more accurate than $10^{-2}$ and also leaves a
smaller spectral tail, while $10^{-2}$ trades accuracy for compressibility, saving 125 of the 1000
ranks of a rank-200 model for 8\% in error. The penalty decreases over every ladder
(Appendix~\ref{sec:recipe}), so the coefficients are reordered toward decay throughout
training.

At the weight used, the orthogonality penalty on the bases is mild, and the bases remain far from orthonormal, at an unweighted penalty of 10 to 50. Larger weights improve the structure at a cost in accuracy. At $10^{-4}$ it has no measurable effect; at $10^{-2}$ it halves the penalty, raises the
error slightly and lets a per-layer rank allocation keep about twenty fewer ranks; and enforcing
orthonormality costs 34\% in error without making the coefficient spectrum decay. Because the
bases are not orthonormal, the coefficients of the trained models are ranked by their actual
contribution rather than by their magnitude (Section~\ref{sec:coef}). The
orthonormality penalty on the projector holds its unweighted value at $10^{-5}$--$10^{-6}$; the
stored projectors of the periodic-acoustics ladder have singular values between 0.88 and 1.33, so
$\bP_k\bP_k^{\top}$ is close to, but not exactly, a projection.

\subsubsection*{Effect of the projection penalty}
\begin{table}[h]\centering
\begin{adjustbox}{max width=\linewidth}\begin{tabular}{rl*{5}{r}rr}\toprule
Width & & \multicolumn{5}{c}{Seed} & Mean & Paired ratio \\
\cmidrule(lr){3-7}
 & & 0 & 1 & 2 & 3 & 4 & & \\\midrule
500 & projector & 13.04\% & 9.00\% & 8.51\% & 8.76\% & 9.49\% & 9.76\% & \\
 & no projector & 16.35\% & 9.31\% & 8.08\% & 9.12\% & 8.82\% & 10.34\% & $1.04\ [0.89, 1.20]$ ($n=5$) \\
1000 & projector & 6.92\% & 6.67\% & 6.61\% & 6.31\% & 7.41\% & 6.79\% & \\
 & no projector & 7.29\% & 6.80\% & 6.61\% & 6.80\% & 6.55\% & 6.81\% & $1.00\ [0.91, 1.10]$ ($n=5$) \\
\bottomrule\end{tabular}\end{adjustbox}
\caption{Local training with and without the projector over five seeds (periodic acoustics, uniform sampling, learning rate $10^{-4}$, 400{,}000 steps). A seed fixes the initialization and the order of the draws for both arms, so the comparison is paired. Means are over the seeds both arms have. Paired ratio: the geometric mean over seeds of the no-projector error divided by the projector error, with its 95\% $t$ interval.}\label{tab:proj_seeds}\end{table}
The penalty was introduced with two aims, to steer locally sampled training toward per-cell subspaces and to provide bases for the per-cell reduction.
The projection misfit was designed as a per-cell constraint, in which each cell has its own projectors and the penalty asks the output not to change when each hidden state is replaced by its projection.
In the \lrnr{} it can be met without local information. Layer $\ell$ reads its input only through
the row basis, $\bV_\ell^{\top}\bz_\ell$, which every cell shares, and a projector whose range
contains the range of $\bV_\ell$ satisfies $\bV_\ell^{\top}\bP_{k,\ell}\bP_{k,\ell}^{\top}\bz_\ell=\bV_\ell^{\top}\bz_\ell$
exactly, in every cell. With $r_s=800$ far above $r=50$ every projector has room for this global
solution, and training moves toward it.

The trained projectors of the periodic-acoustics ladder show this. They stay at their random
initialization in all but a few of their 800 directions; per cell, 7 to 9 directions rotate by more
than $5^\circ$, and the largest angle is $15$--$22^\circ$. The rotated directions are shared across
cells and point toward the row basis of the next layer; seen through that basis, 75--89\% of the
change is common to all cells. A cell's twin pass is accordingly as accurate with a neighbor's or a
distant cell's projector as with its own. The penalty leaves a weak local trace. The cell-specific part of the change correlates with the cell's own hidden-state statistics in the first
and last hidden layers (correlation 0.10--0.13, against about zero with a distant cell's), and in the
last hidden layer it varies smoothly from cell to cell. In absolute terms the penalty is met loosely.
At the end of the ladder the projected pass reproduces the plain pass to about 45\% (root-mean-square
difference relative to the root-mean-square solution) while the plain pass fits the data to 6\%;
untrained random projectors give 98\%.

Our experiments show that the penalty, while it does not reduce the accuracy, does not improve locally
sampled training overall. Without it (no
projection misfit and no projector orthonormality penalty), the four-stage ladder ends at 5.0\%
against 4.9\% with adaptive sampling and at 5.6\% against 6.1\% with uniform sampling, and the
single-seed width sweep of Table~\ref{tab:lrnr_width} shows no consistent direction. Over five
paired seeds the two arms differ by a factor of 1.04 (95\% interval 0.89--1.20) at width 500 and of
1.00 (0.91--1.10) at width 1000 (Table~\ref{tab:proj_seeds}), and the per-cell hidden-state dimension
of the first three hidden layers agrees within about 10\% with and without the penalty. The per-cell
reduction in Section~\ref{sec:fast} therefore samples its own bases from the trained model.
How to make the projectors effective, so that they shape locally sampled training instead of
leaving only subtle effects, is an open problem. Projectors that encode local information would have
to be denied the global route, for instance by building them from each cell's own hidden states, as the FastLRNR's sampled bases are built, so that the two would coincide (Section~\ref{sec:discussion}).

\section{Analysis of the trained models: rank and hyper-modes}
\label{sec:analysis}

\subsection{Rank usage}
\label{sec:coef}

For each layer $\ell$ the contribution of coefficient $i$ is the operator norm of its
rank-one term over the trajectory, $c_{\ell,i}=\max_t|s_{\ell,i}(t)| \|\bU_{\ell,i}\| \|\bV_{\ell,i}\|$,
with $\bU_{\ell,i}$, $\bV_{\ell,i}$ the $i$-th columns of the bases. Ranking the coefficients by $c_{\ell,i}$ and keeping the leading $k$ in every layer gives a
truncation ladder whose error against the solver measures the model's effective rank.

\begin{table}[h]\centering
\begin{adjustbox}{max width=\linewidth}\begin{tabular}{lrrrlrrrrrrrrrrr}\toprule
 & & & \multicolumn{2}{c}{Hypernetwork} & Dense & \multicolumn{5}{c}{Channels for 90\%, layer} & \multicolumn{5}{c}{Channels for 99\%, layer} \\
\cmidrule(lr){4-5}\cmidrule(lr){7-11}\cmidrule(lr){12-16}
Model & Parameters & MB & Parameters & Shape & equivalent & 1 & 2 & 3 & 4 & 5 & 1 & 2 & 3 & 4 & 5 \\\midrule
Euler & 1,206,140 & 4.60 & 3,040 & 3$\times$5 & 12,124,004 & 39 & 39 & 34 & 23 & 14 & 48 & 49 & 47 & 38 & 33 \\
Acoustics & 1,267,756 & 4.84 & 64,756 & 2$\times$128 & 12,122,003 & 42 & 44 & 43 & 36 & 5 & 49 & 50 & 50 & 48 & 6 \\
Advection & 1,267,556 & 4.84 & 64,756 & 2$\times$128 & 12,118,001 & 43 & 35 & 32 & 24 & 7 & 50 & 45 & 43 & 38 & 19 \\
Burgers & 1,267,556 & 4.84 & 64,756 & 2$\times$128 & 12,118,001 & 39 & 35 & 17 & 18 & 20 & 48 & 45 & 33 & 33 & 34 \\
Shock-bubble & 1,267,956 & 4.84 & 64,756 & 2$\times$128 & 12,126,005 & 40 & 43 & 44 & 44 & 40 & 49 & 50 & 50 & 50 & 49 \\
Inclusions & 1,267,756 & 4.84 & 64,756 & 2$\times$128 & 12,122,003 & 35 & 23 & 20 & 17 & 12 & 46 & 37 & 35 & 33 & 22 \\
\bottomrule\end{tabular}\end{adjustbox}
\caption{Parameter counts (bases + hypernetwork), storage in float32, the dense MLP of the same widths for comparison, and the number of coefficient channels (of 50) that carry 90\% and 99\% of the contribution mass per layer, ranked by $c_{\ell,i}=\max_t|s_{\ell,i}|\,\|\bU_{\ell,i}\|\,\|\bV_{\ell,i}\|$.}\label{tab:coef}\end{table}

\begin{table}[h]\centering
\begin{adjustbox}{max width=\linewidth}\begin{tabular}{lrrrrrrrrrrr}\toprule
keep per layer & 50 & 49 & 48 & 47 & 45 & 42 & 37 & 25 & 12 & 6 & 4 \\\midrule
Euler & 0.017 & 0.040 & 0.242 & 0.246 & 0.311 & 0.349 & 0.441 & 0.681 & 0.680 & 0.739 & 0.799 \\
acoustics & 0.015 & 0.149 & 0.354 & 0.852 & 0.935 & 1.003 & 1.015 & 0.982 & 0.933 & 0.967 & 0.961 \\
advection & 0.035 & 0.105 & 0.114 & 0.216 & 0.779 & 0.760 & 0.807 & 0.908 & 1.023 & 1.159 & 1.167 \\
Burgers & 0.020 & 0.021 & 0.024 & 0.055 & 0.123 & 0.526 & 0.711 & 0.811 & 1.240 & 1.017 & 1.088 \\
shock-bubble & 0.037 & 0.071 & 0.107 & 0.287 & 0.304 & 0.340 & 0.574 & 0.710 & 0.833 & 0.924 & 0.941 \\
inclusions & 0.018 & 0.018 & 0.018 & 0.022 & 0.160 & 0.701 & 0.923 & 1.059 & 1.155 & 1.079 & 1.087 \\
\bottomrule\end{tabular}\end{adjustbox}
\caption{Truncation ladder: relative $L^2$ error (rms over frames 0/20/40/60/80) when only the top-$k$ channels of \emph{every} layer are kept.}\label{tab:trunc}\end{table}

\begin{figure}[tp]
\centering\stackh{0.36}
\figor{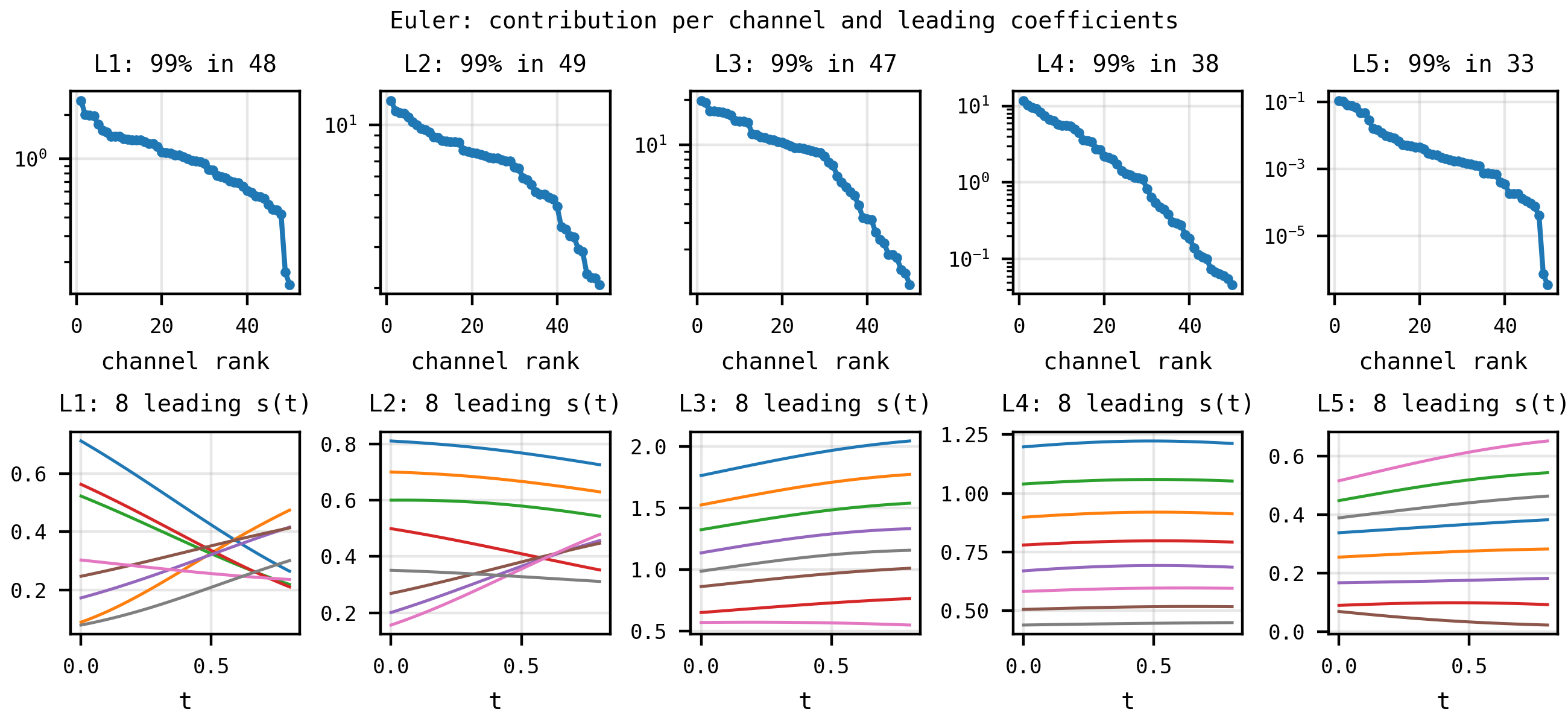}\\[4pt]\figor{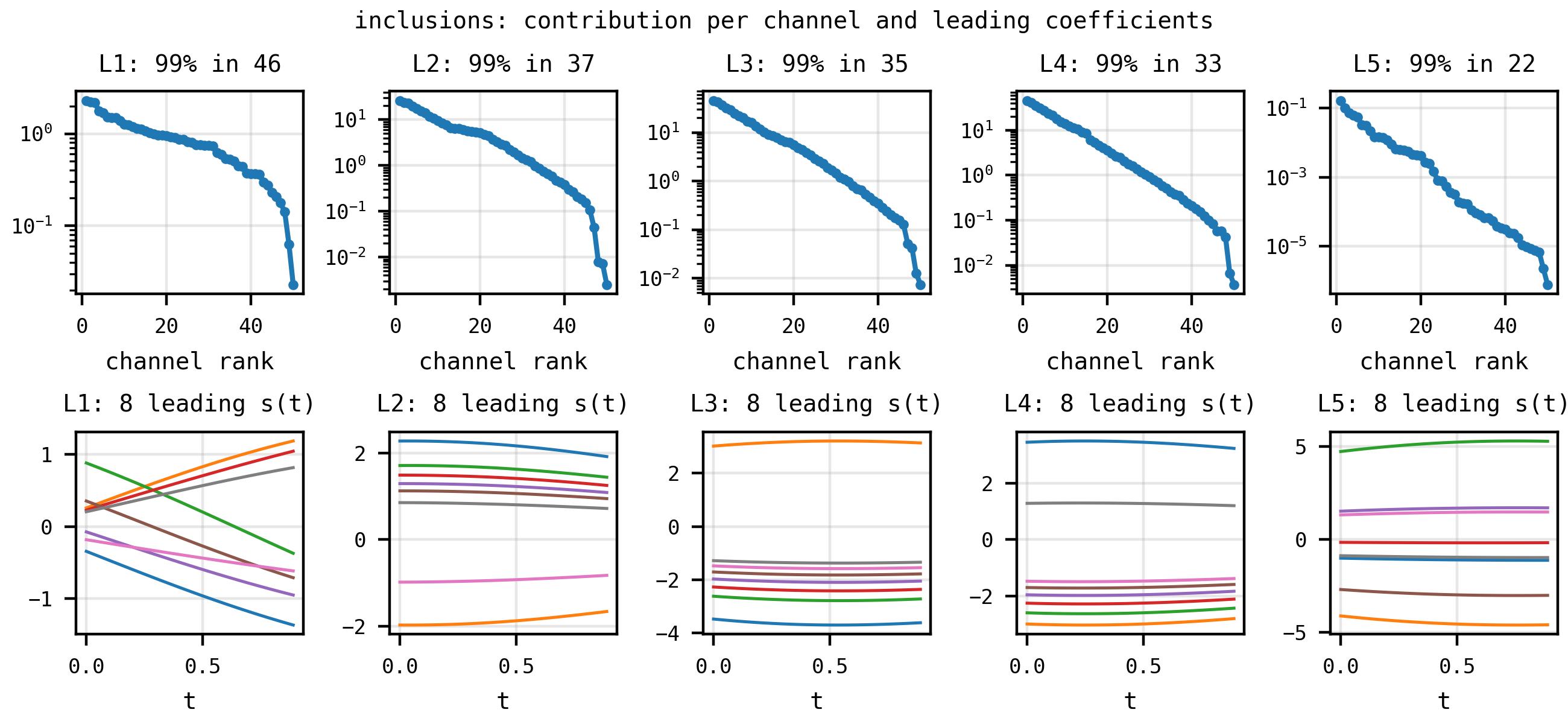}
\caption{Per layer, the ranked contribution of the 50 coefficients (top) and the time
courses $s_{\ell,i}(t)$ of the eight leading coefficients (bottom), for the Euler and inclusions
models (acoustics and advection in Fig.~\ref{fig:coef_b}, Burgers and shock--bubble in
Fig.~\ref{fig:coef_new}, Appendix~\ref{sec:morefigs}).}
\label{fig:coef}
\end{figure}

\subsubsection*{Parameters and storage}
Every model has $1.2$--$1.27\times10^6$ parameters (4.6--4.8~MB in float32), of which
$1.2\times10^6$ are the three bases of the five layers and the rest the hypernetwork
($3{,}040$ for the Euler model, $64{,}756$ for the $2\times128$ networks of the other five); a
dense MLP of the same widths would have $1.21\times10^7$ weights, ten times more, and would
have no time dependence at all (Table~\ref{tab:coef}).

\subsubsection*{Coefficient contributions}
The contribution ranking (Figs.~\ref{fig:coef}, \ref{fig:coef_b} and \ref{fig:coef_new}, Table~\ref{tab:coef}) of the
Euler, acoustics, advection and Burgers models is flat in the hidden layers and steep only in
the output layer: in the first three hidden layers 34--44 of the 50
coefficients are needed for 90\% of the contribution mass and 47--50 for 99\%; the fourth hidden
layer needs 18--36; the output layer 5--20. The sparsity penalty, which asks for a geometric
decay along the coefficient index, has produced a decay only where the output is formed; inside
the INR the models spread the representation over nearly all coefficients with a dynamic
range of $\max_t|s_{\ell,i}|$ of 3--30 within a layer. The shock--bubble and inclusions models depart
from this in opposite directions (Figs.~\ref{fig:coef} and \ref{fig:coef_new}). The shock--bubble model, which forms five
output variables, needs 40--44 coefficients for 90\% in \emph{every} layer, the output layer
included. Beyond the first layer, the inclusions model decays the most steeply of the six, with 17--23 coefficients for
90\% in layers 2--4 and 12 in the output layer, with three or four coefficients per layer that are below $10^{-3}$ of the largest at all times, while its first layer needs 35.

\subsubsection*{Truncation}
The truncation results (Table~\ref{tab:trunc}) show that keeping the top 37 of 50
coefficients in every layer raises the relative error from 1.5--3.7\% to 44\% (Euler), 57\%
(shock--bubble), 71\% (Burgers), 81\% (advection), 92\% (inclusions) and 101\% (acoustics), and
every smaller $k$ is at or above the field itself. Keeping 45 costs 12\% (Burgers), 16\%
(inclusions), 30\% (shock--bubble), 31\% (Euler), 78\% (advection) and 94\% (acoustics).
Dropping one coefficient per layer already takes acoustics from 1.5\% to 15\%, advection
from 3.5\% to 10.5\%, the shock--bubble model from 3.7\% to 7.1\% and Euler from 1.7\% to
4.0\%, and only Burgers and inclusions absorb it (these ladders weight every solver point
equally, hence the larger baseline than Table~\ref{tab:examples}). This is
consistent with the rank comparison in Appendix~\ref{sec:design}, where rank 50 equalled rank
200: on these problems 50 is the effective rank, which does not compress further. 

The layer-wise ladder
(Table~\ref{tab:trunc_layer} in Appendix~\ref{sec:morefigs}), which truncates one layer at a
time and leaves the others intact, shows that the \emph{first} layer is the most sensitive. Dropping its
five weakest coefficients alone costs 93\% (acoustics), 78\% (advection), 31\% (Euler) and 12\%
(Burgers), 30\% (shock--bubble) and 16\% (inclusions), whereas layers 2--5 tolerate the loss of
five coefficients at no more than twice the error on every model. At larger cutoffs the models differ. On Euler, acoustics, advection and Burgers, layers 2--5 lose five to ten coefficients at no more than
twice the error and layers 4--5 keep 30 of 50 at $+0$--$50\%$, while keeping 30 coefficients costs
a factor of 1.6--2.9 in layers 2--5 of the inclusions model and 4.5--7.9 in layers 4--5 of the
shock--bubble model, whose every layer is flat. The first layer multiplies the 54
Chebyshev features, and every one of its 50 coefficients contributes spectral content that the rest of the network needs; deeper in the INR the representation is compressible to 30--40
coefficients one layer at a time. In these examples, the ten-fold saving over a
dense network therefore comes from the factorization itself, further reduction
comes from the per-cell route in Section~\ref{sec:fast}, whose local bases need
far fewer directions than the global rank, as Section~\ref{sec:theory} predicted.

Three consequences regarding the layer-wise sensitivity
follow from Table~\ref{tab:trunc_layer}. First, the rank is a per-layer
quantity, but the first layer influences it. The deeper layers each tolerate the loss of five to
ten coefficients on their own at no more than twice the error, yet keeping 37
coefficients in every layer at once costs 44--101\% as
the losses compound from layer to layer. A smaller rank in the deeper layers is therefore not
available after training; it would have to be trained in, for example as rank 50 in the first
layer and 30--40 below it (we have not tested
that design). Second, the time dependence is concentrated in the first layer. The hyper-mode
analysis in Section~\ref{sec:hyper} finds 42--78\% of the energy of the leading modes in the
first layer's coefficients on the Euler, acoustics and inclusions models, so truncating that
layer removes spectral content and time dependence together. An error made in the first layer passes through every
later sine layer, each with its own amplification factor, whereas an error in the output layer
reaches the output directly (Proposition~\ref{prop:layers}). 
Third, the global sensitivity need not hold for a single cell. Inside a cell the width argument bounds the dimension of every layer's hidden states by
the cell size, not by the layer's position, and in the 100-neuron study in
Section~\ref{sec:fast} giving the first layer more neurons than the others was not more accurate than the best even allocation. Per-layer budgets therefore have a larger effect on the global model, where the first layer must keep its full rank, than on the per-cell leaves.

\subsection{Hyper-modes of the time dependence}
\label{sec:hyper}

The hypernetwork's output over the horizon, all coefficients $(\bs_\ell,\bs^{\rm b}_\ell)_\ell$ stacked, is
a curve $\bs(t)\in\R^{2rL}$ (500 numbers for $r=50$, $L=5$). Its singular value decomposition about the mean,
$\bS-\bsbar=\bU\bSigma\bV^{\top}$ with $\bS\in\R^{81\times500}$, gives the \emph{hyper-modes} $\bv_k$
(directions in coefficient space) and their time courses $\sigma_k \bu_k(t)$. The number of
modes needed to represent $\bs(t)$ is a second measure of the model's dimension, independent of
the rank, and a perturbation of the coefficients along a mode, $\bs(t)+\epsilon \sigma_k \bv_k$,
is a perturbation of the \emph{solution} whose physical meaning can be inspected. A decomposition of this kind, with interpretable modes, was reported for \lrnr{}s trained globally on whole snapshots~\citep{cho2025hyperbolic}; here it is applied to models trained with local batches.

\begin{figure}[tp]
\centering
\figor{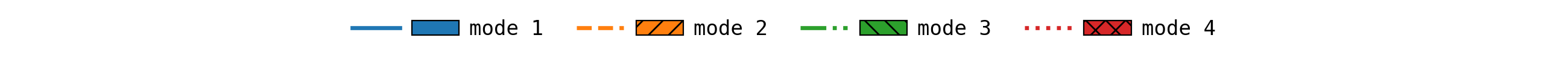}\\[0pt]\figor{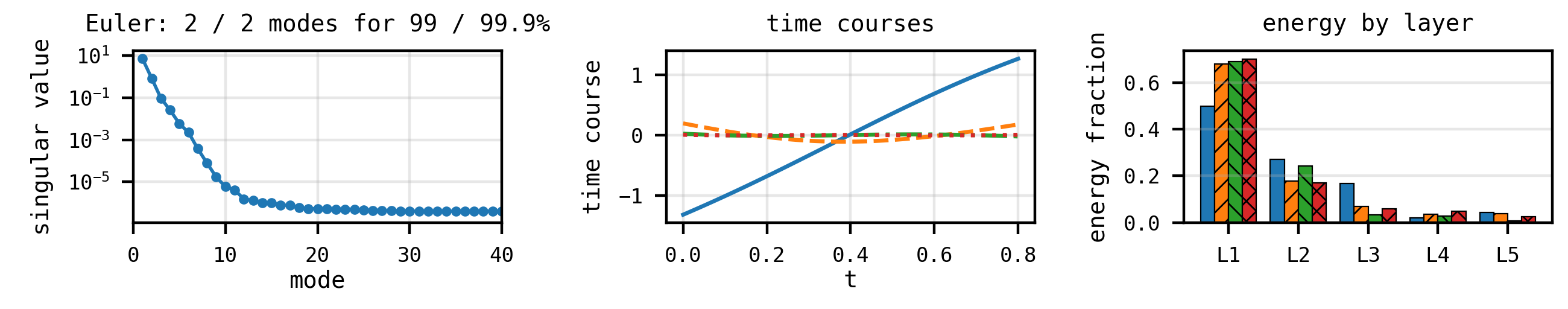}\\[1pt]\figor{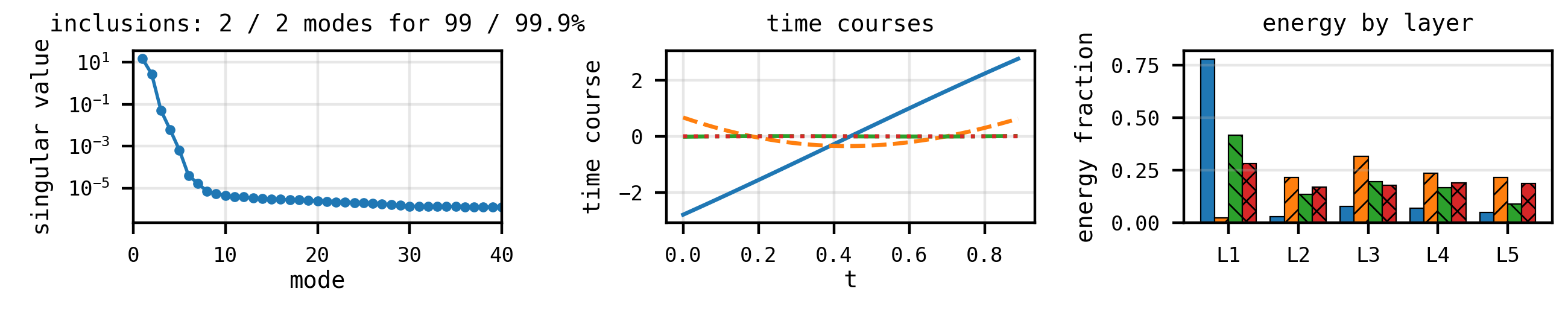}\\[1pt]\figor{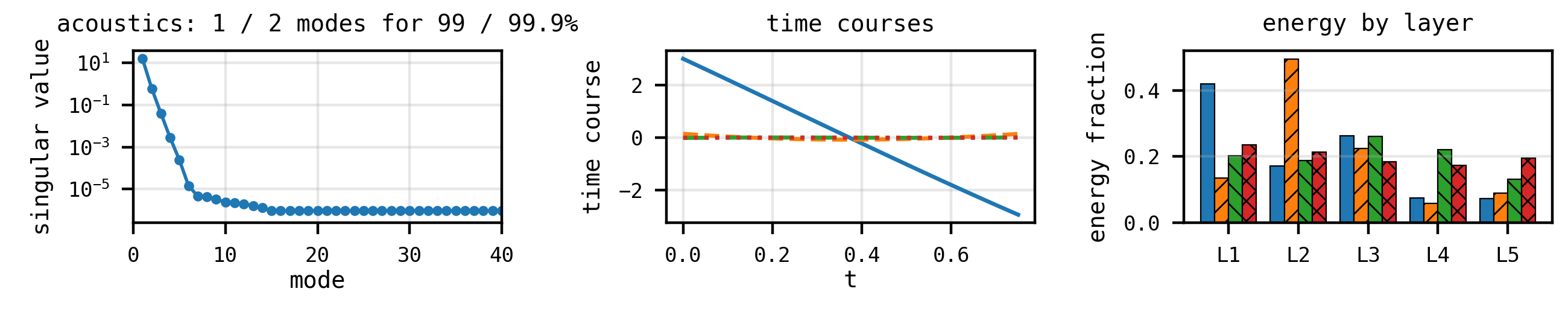}\\[1pt]%
\figor{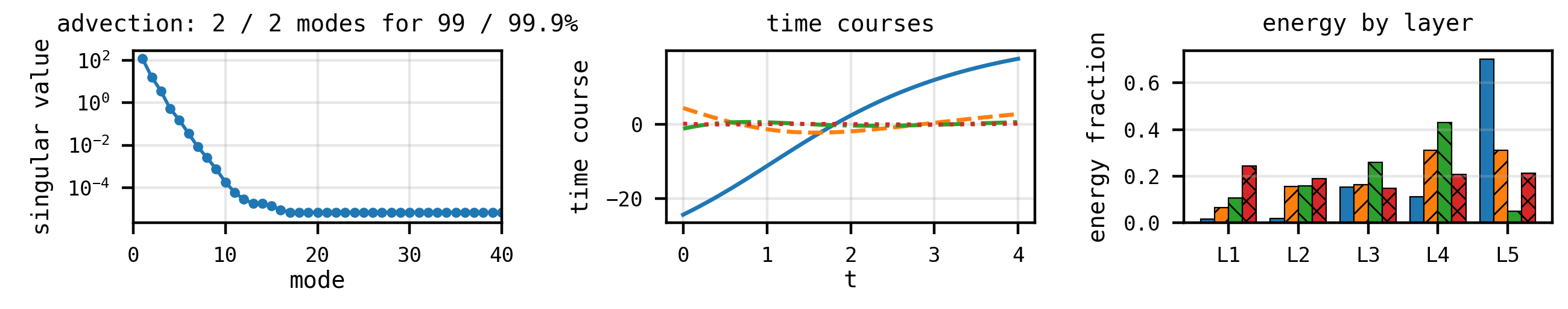}\\[1pt]\figor{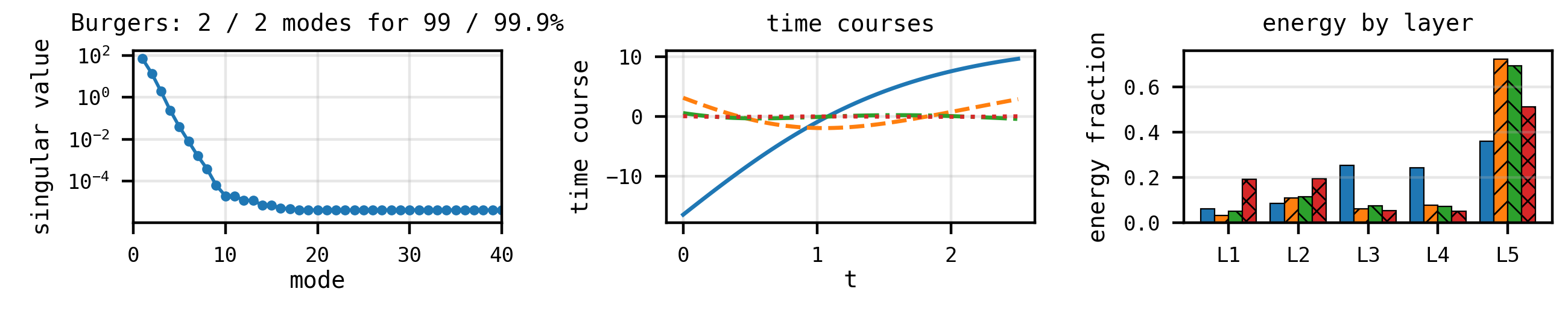}\\[1pt]\figor{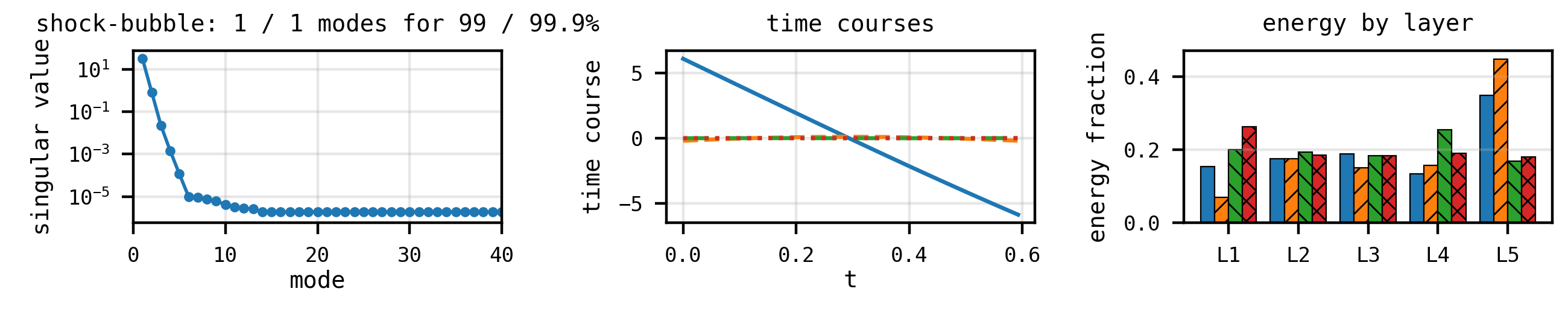}
\caption{Hyper-modes of the six models, one row each: the singular values of the coefficient
trajectory (left), the time courses of the four leading modes (middle, told apart by line style)
and the distribution of each mode's energy over the layers (right, told apart by hatching).}
\label{fig:hm_spectrum}
\end{figure}

\subsubsection*{Dominant directions of the coefficient trajectory}
For every model the singular values of $\bS-\bsbar$ fall by a factor of 5--38 from the first
to the second mode and by $10^2$--$10^4$ by the fourth (Table~\ref{tab:hm},
Fig.~\ref{fig:hm_spectrum}); one mode (acoustics, shock--bubble)
or two (Euler, advection, Burgers, inclusions) account for 99\% of the variance of $\bs(t)$, and at most
two account for 99.9\% (one on the shock--bubble model, whose second singular value is $1/38$ of the
first). Over 81 times the hypernetwork traces,
in its 500-dimensional output, a curve that is planar to three digits. The time courses of
the leading modes are smooth and monotone in $t$. Mode 1 is, to a good approximation, time
itself re-parametrized, and mode 2 its curvature. Fig.~\ref{fig:coef} shows the same coefficient by coefficient; the individual $s_{\ell,i}(t)$ are nearly affine in $t$, so the
hypernetwork, for all its capacity, produces an output close to a straight line through
coefficient space. The nonlinearity of the representation is in the INR, not in the map
from $t$ to the coefficients; the theoretical construction for entropy solutions in \citep{rim2024lrnr} also keeps the time dependence linear. The layers in which the modes act differ by problem. On Euler and acoustics, 42--70\% of the energy of every leading mode lies in the first layer's coefficients (the
modes re-weight the input frequencies), and on the inclusions model 78\% of mode 1; on
advection 70\% of mode 1 is from the output layer (the swirl is a re-weighting of fixed hidden
features); on Burgers and the shock--bubble the energy is spread across the layers, with a
third of the shock--bubble's mode 1 in the output layer.

\begin{table}[h]\centering
\begin{adjustbox}{max width=\linewidth}\begin{tabular}{l*{6}{S[table-format=1.2e-2]}cc}\toprule
Model & \multicolumn{6}{c}{Singular values of $\bS-\bsbar$} & \multicolumn{2}{c}{Modes for} \\
\cmidrule(lr){2-7}\cmidrule(lr){8-9}
 & {$\sigma_1$} & {$\sigma_2$} & {$\sigma_3$} & {$\sigma_4$} & {$\sigma_5$} & {$\sigma_6$} & 99\% & 99.9\% \\\midrule
Euler & 6.99e+00 & 8.27e-01 & 9.33e-02 & 2.56e-02 & 5.83e-03 & 2.23e-03 & 2 & 2 \\
Acoustics & 1.57e+01 & 6.12e-01 & 3.99e-02 & 2.78e-03 & 2.48e-04 & 1.43e-05 & 1 & 2 \\
Advection & 1.18e+02 & 1.63e+01 & 3.54e+00 & 5.23e-01 & 1.53e-01 & 3.54e-02 & 2 & 2 \\
Burgers & 7.00e+01 & 1.41e+01 & 1.96e+00 & 2.44e-01 & 3.95e-02 & 7.82e-03 & 2 & 2 \\
Shock-bubble & 3.17e+01 & 8.45e-01 & 2.29e-02 & 1.47e-03 & 1.18e-04 & 1.16e-05 & 1 & 1 \\
Inclusions & 1.48e+01 & 2.76e+00 & 5.13e-02 & 6.09e-03 & 6.56e-04 & 3.94e-05 & 2 & 2 \\
\bottomrule\end{tabular}\end{adjustbox}
\caption{Hyper-mode spectra of the coefficient trajectory $\bs(t)\in\mathbb{R}^{500}$ over the 81 times: the six leading singular values and the number of modes that carry 99\% and 99.9\% of its variance.}\label{tab:hm}\end{table}

\subsubsection*{Perturbations along the modes}
\begin{figure}[tp]
\centering\stackh{0.9}
\figor{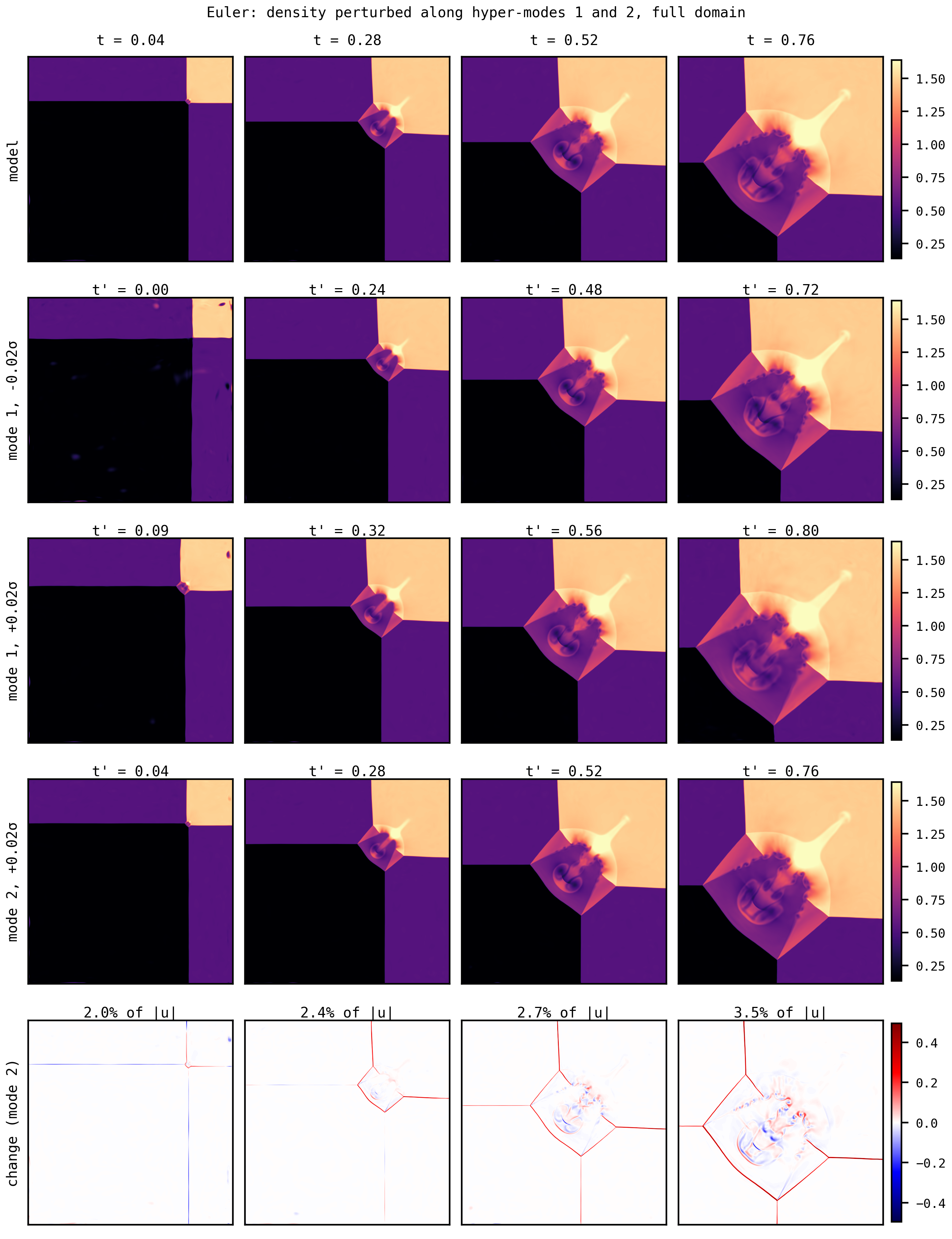}
\caption{Euler quadrants, density, perturbed along the hyper-modes at four times. Rows: the
model $u(t)$; the field at $\bs(t)-0.02 \sigma_1 \bv_1$ and at $\bs(t)+0.02 \sigma_1 \bv_1$ (the
title gives the time $t'$ on the model's own trajectory nearest to the perturbed field); the
field at $\bs(t)+0.02 \sigma_2 \bv_2$ and the change it makes, as a fraction of $\|u\|$.}
\label{fig:hm_perturb}
\end{figure}

The map from coefficients to fields is sensitive (a coefficient error of $7\times10^{-4}$
doubles the Euler error), so a step of one full $\sigma_k$ along a mode crosses the whole trajectory and leaves the solution manifold. The perturbations are therefore small,
$\epsilon=0.02 \sigma_k$, backward and forward along mode 1 (oriented so that $+$ advances
time) and forward along mode 2, and each perturbed field is compared with the model's own
trajectory to find the nearest time $t'$. The four times shown are chosen among the interior
frames so that every perturbed field stays within 15\% of the trajectory; a time at which a
step would leave the horizon or the solution manifold is not shown (this excludes the first
frames of the inclusions problem, where the pulse is still entering, and one late acoustics
frame). Figure~\ref{fig:hm_perturb} shows the Euler problem; Appendix~\ref{sec:morefigs} the
other five and the Kelvin--Helmholtz window.

A step of $0.02 \sigma_1$ is a step of about four frames along the model's own evolution. On
the Euler, acoustics, shock--bubble and inclusions problems the nearest trajectory time moves by
four frames (Euler four or five) at every one of the four times, backward for the backward step
and forward for the forward one, and the distance to that trajectory point is 0--14\% of the
field against a change of 16--118\% of it; on Burgers the step is three to six frames. The
shocks of the Euler problem advance with the billows developing behind them, the acoustic
rings expand and their periodic images cross earlier or later, the Burgers shock travels, the
shock--bubble's shock and roll-up advance together, and on the inclusions problem the plane
front and the scattered rings move as one. The swirling advection problem is an exception, because its flow reverses at $t=2$ and the interface at $t$ and at $4-t$ are near-mirror images, so
the coefficient curve is not monotone in time and a step along mode 1 can reach \emph{another phase of the swirl} (from $t=0.2$ a backward step reaches the state of $t=1.9$,
from $t=3.8$ a forward step reaches $t=2.05$), overall a cleanly wound or unwound interface,
rather than a smeared one.

Mode 2 is the small correction. Its $0.02 \sigma_2$ step changes the field by 0.3--1.1\%
(acoustics, shock--bubble, inclusions), 1.2--3.5\% (Euler), 1.8--7.2\% (Burgers) and 2.0--14\%
(advection), and leaves the nearest trajectory time where it was (within one frame, except twice on
advection, where it too reaches the mirror phase), so it is not a time step. The change maps show that it moves the fronts. On Euler the change
is a thin signed line along every shock (a re-timing of the shock positions), on acoustics a
set of concentric rings (the ring radii shift), on Burgers a ridge along the shock, and on
advection a two-signed band along the interface; on the shock--bubble a line along the leading
shock and the bubble's edge, and on the inclusions problem a two-signed doublet on every front,
strongest on the ring focusing in the slow disc.

\section{Per-cell reduction: FastLRNR}
\label{sec:fast}

A trained \lrnr{} is still a global network, in which every one of the 2000 neurons of a hidden layer is
evaluated at every point, although Section~\ref{sec:theory} shows that inside a small cell the
hidden states over the cell's points and over time lie close to a subspace of low dimension.
This section turns that observation into a computation. For a single cell we build a small
network, which we call the \emph{leaf}, 
that reproduces the trained model on the cell from a few hundred
neurons per layer. The construction follows the finite element pattern of the introduction. A local basis for the hidden states plays the role of the element's shape functions, and a set of
selected neurons whose values determine the coefficients in that basis plays the role of the
nodal variables.

\subsubsection*{Construction of a leaf}
A leaf is built in four steps.
\begin{enumerate}[nosep]
 \item Sample the hidden states of the trained model on the cell, at a few hundred times and
   points.
 \item Take the leading $r_{\rm cell}$ eigenvectors of their Gram matrix as a local basis for
   each layer. This is the subspace that Section~\ref{sec:theory} predicts.
 \item Select $m\approx1.5 r_{\rm cell}$ neurons per layer by oversampled empirical
   interpolation \citep{barrault2004eim,chaturantabut2010deim}, so that the whole hidden state
   can be reconstructed from the values of those neurons.
 \item Fold the resulting rank-$r_{\rm cell}$ layer maps into small dense matrices and fine-tune
   them on the cell, against the model and, where solver points are available, against the
   solver.
\end{enumerate}
The leaf stores $m\times r_{\rm cell}$ floating point numbers per layer in place of $2000\times50$. Its
coefficients remain linear in the global $\bs_\ell(t)$, so the hypernetwork is shared with the
full model and is not changed. The reconstruction error of one layer propagates through the remaining layers as bounded by Proposition~\ref{prop:layers}.

\subsubsection*{Choice of cells}
We reduce one cell per problem: the most complicated cell of a $10\times10$ tiling, measured by
the solver's own refinement, since a local basis has the most to capture there. On the
Euler problem this is the shock--contact junction, with 3{,}900 points per frame, $3.3\times$
the mean over cells. On advection it is the swirl center (800 points, $2.1\times$). The
acoustics and Burgers runs refine almost uniformly, so their chosen cells are only mildly above
the mean. On the shock--bubble problem the cell lies on the path of the vortex ring
($3.2\times$ the mean). On the inclusions problem it lies at the edge of the fast disc
($6.9\times$) and stays empty until the transmitted pulse reaches it at $t\approx0.5$.

\begin{table}[h]\centering
\begin{adjustbox}{max width=\linewidth}\begin{tabular}{lrrrrrrr}\toprule
 & Neurons & \multicolumn{2}{c}{Agreement} & \multicolumn{3}{c}{RMSE against the solver} & \\
\cmidrule(lr){3-4}\cmidrule(lr){5-7}
Cell & per layer & Frames & Held-out & Model & Leaf, frames & Leaf, held-out & Floats \\\midrule
Euler 77 & 100 & 1.50\% & 1.24\% & 0.0154 & 0.0331 & 0.0264 & 59,293 \\
Euler 77 & 150 & 1.02\% & 1.06\% & 0.0154 & 0.0284 & 0.0216 & 86,586 \\
Euler 77 & 200 & 0.75\% & 0.75\% & 0.0154 & 0.0255 & 0.0180 & 114,436 \\
Euler 77 & 300 & 0.55\% & 0.58\% & 0.0154 & 0.0225 & 0.0154 & 170,136 \\
Euler 77 & 450 & 0.47\% & 0.45\% & 0.0154 & 0.0194 & 0.0136 & 253,686 \\
Acoustics 87 & 100 & 2.02\% & 2.60\% & 0.0029 & 0.0046 & 0.0040 & 63,083 \\
Acoustics 87 & 150 & 1.62\% & 2.19\% & 0.0029 & 0.0034 & 0.0031 & 92,238 \\
Acoustics 87 & 200 & 1.28\% & 1.73\% & 0.0029 & 0.0026 & 0.0026 & 121,988 \\
Acoustics 87 & 300 & 1.11\% & 1.50\% & 0.0029 & 0.0020 & 0.0023 & 181,488 \\
Acoustics 87 & 450 & 0.97\% & 1.44\% & 0.0029 & 0.0021 & 0.0021 & 270,738 \\
Advection 45 & 100 & 1.44\% & 7.71\% & 0.0180 & 0.0968 & 0.0658 & 60,660 \\
Advection 45 & 150 & 1.19\% & 7.74\% & 0.0180 & 0.0881 & 0.0679 & 88,737 \\
Advection 45 & 200 & 0.99\% & 7.17\% & 0.0180 & 0.1017 & 0.0630 & 117,387 \\
Advection 45 & 300 & 0.88\% & 9.31\% & 0.0180 & 0.1217 & 0.0802 & 174,687 \\
Advection 45 & 450 & 0.90\% & 10.64\% & 0.0180 & 0.1481 & 0.0914 & 260,637 \\
Burgers 16 & 100 & 2.00\% & 12.37\% & 0.0162 & 0.0602 & 0.0309 & 59,050 \\
Burgers 16 & 150 & 1.81\% & 27.18\% & 0.0162 & 0.1161 & 0.0588 & 86,343 \\
Burgers 16 & 200 & 1.72\% & 22.08\% & 0.0162 & 0.0961 & 0.0491 & 114,193 \\
Burgers 16 & 300 & 1.58\% & 24.01\% & 0.0162 & 0.1067 & 0.0538 & 169,893 \\
Burgers 16 & 450 & 1.56\% & 24.84\% & 0.0162 & 0.1060 & 0.0533 & 253,443 \\
Shock--bubble 32 & 100 & 0.59\% & 1.02\% & 0.0454 & 0.0760 & 0.1093 & 63,699 \\
Shock--bubble 32 & 150 & 0.51\% & 0.85\% & 0.0454 & 0.0614 & 0.0953 & 93,050 \\
Shock--bubble 32 & 200 & 0.46\% & 0.62\% & 0.0454 & 0.0448 & 0.0684 & 123,000 \\
Shock--bubble 32 & 300 & 0.38\% & 0.55\% & 0.0454 & 0.0342 & 0.0606 & 182,900 \\
Shock--bubble 32 & 450 & 0.36\% & 0.52\% & 0.0454 & 0.0346 & 0.0611 & 272,750 \\
Inclusions 56 & 100 & 1.84\% & 2.81\% & 0.0107 & 0.0109 & 0.0153 & 59,842 \\
Inclusions 56 & 150 & 1.44\% & 2.29\% & 0.0107 & 0.0093 & 0.0138 & 87,429 \\
Inclusions 56 & 200 & 1.32\% & 1.97\% & 0.0107 & 0.0088 & 0.0121 & 115,579 \\
Inclusions 56 & 300 & 1.23\% & 2.01\% & 0.0107 & 0.0082 & 0.0128 & 171,879 \\
Inclusions 56 & 450 & 1.19\% & 1.88\% & 0.0107 & 0.0077 & 0.0115 & 256,329 \\
\bottomrule\end{tabular}\end{adjustbox}
\caption{FastLRNR leaves of the most complicated cell of each example (solver-matched fine-tune, 60{,}000 steps): neurons per layer (the first rung has 101), agreement with the full model (relative $L^2$ on the cell at the sampled frames and at the 20 held-out times), error against the solver's points in the cell (model vs leaf, at the evaluation frames and at the held-out times), and floating point numbers stored by the leaf (the full model: $1.20\times10^6$).}\label{tab:fast}\end{table}

\begin{figure}[tp]
\centering
\figor{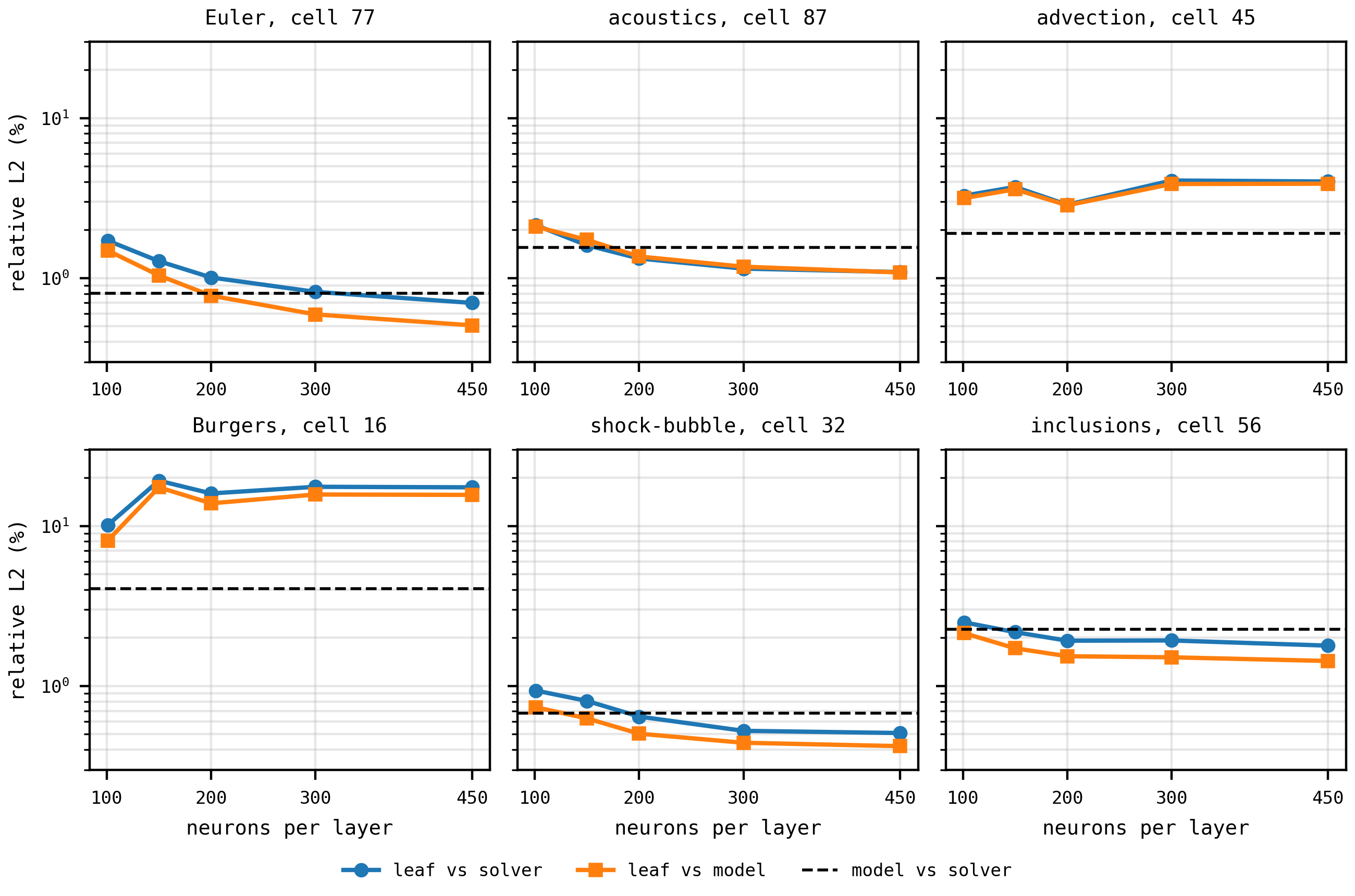}
\caption{The width ladder of the most complicated cell of each problem: the leaf's error against
the solver's points in the cell, its agreement with the full model, and the model's own error
(dashed), relative $L^2$ pooled over all times, against the neurons per layer. The leaves
themselves, their difference from the model and the errors over time are in
Figs.~\ref{fig:fast_cells_a}--\ref{fig:fast_cells_c} (Appendix~\ref{sec:morefigs}).}
\label{fig:fast}
\end{figure}

\subsubsection*{Width ladders}
Each cell was reduced at five widths, from 100 to 450 neurons per layer (basis ranks 67 to 300
with $1.5\times$ oversampling). The sampled basis is built from 61 of the 81 times, the
fine-tune is matched to the model and to the solver's points in the cell, and the remaining 20
times are held out (Table~\ref{tab:fast}, Fig.~\ref{fig:fast}). The results fall into two
regimes, depending on how much data the cell holds.

In cells with many solver points the reduction is accurate. On the Euler cell, with 73{,}000 points, the
leaf's agreement with the model improves from 1.5\% to 0.47\% along the ladder, and its error
against the solver at the held-out times falls from 0.026 to 0.0136, against the model's 0.0130.
At 450 neurons the leaf thus matches its model with 21\% of the floating point numbers. On the acoustics cell
the leaf is even slightly more accurate than its model at the held-out times from 200 neurons
on (0.0023 and 0.0021 against 0.0026).

In cells with few solver points the leaves over-fit the sampled times. The advection cell holds 4{,}400 points and
the Burgers cell 1{,}400. Their leaves agree with the model to 1--2\% at the sampled times but
only to 8--27\% at the held-out ones, their held-out error against the solver is three to six
times the model's, and it grows with width. A 60{,}000-step fine-tune of a 450-neuron leaf on 61
frames of a few thousand points memorizes those frames.

Two variants of the fine-tune reveal the cause (Table~\ref{tab:fast_variants} in
Appendix~\ref{sec:morefigs}). Removing the solver term does not help with the over-fitting
(advection 0.070 against 0.066, with the model at 0.020). A fine-tune a quarter as long helps
the wide leaves but still leaves them at two to three times the model's error. The
450-neuron advection leaf is better before any fine-tune (0.043, with 4.4\% agreement) than
after it (0.056, 6.4\%). On the two large cells, by contrast, the solver term lowers the error by 5--30\%.
The practical rule is to fine-tune against the solver where a cell holds tens of thousands of
its points, and to keep the sampled basis and interpolation set unchanged where it holds only a
few thousand.

The shock--bubble and inclusions cells lie between the two regimes and test this rule. The
shock--bubble cell holds 16{,}000 solver points at the five evaluation times and behaves like a
large cell. Its leaves agree with the model to 0.36--0.59\% at the sampled times and to
0.52--1.0\% at the held-out ones, are more accurate than the model against the solver at the
evaluation times from 200 neurons on (0.0448 and 0.0342 against 0.0454), and at the held-out
times reach 0.061 from 300 neurons, 7\% above the model's 0.057. The inclusions cell holds
9{,}900 points and also reduces accurately. Its leaves agree with the model to 1.2--1.8\% at the
sampled times and 1.9--2.8\% at the held-out ones, are more accurate than the model at the
evaluation times from 150 neurons on (0.0093 down to 0.0077 against 0.0107), and at the
held-out times are 6\% above the model at 200 neurons (0.0121 against 0.0114) and level with it
at 450 (0.0115).

\subsubsection*{Allocation of a 100-neuron budget}
The ladders fix the ratio of interpolation points to basis rank and fine-tune only the folded
rows. To see how much a small leaf can recover with other choices, we forced both large cells
to about 100 neurons per layer and compared six ways of spending that budget
(Table~\ref{tab:fast100} in Appendix~\ref{sec:morefigs}):
\begin{itemize}[nosep]
 \item the ladder's default, basis rank 67 with $1.5\times$ oversampling;
 \item a lower rank with more interpolation points;
 \item a higher rank, barely oversampled;
 \item a fine-tune of the column and row bases and the biases, rather than of the folded rows
   only;
 \item a longer and slower fine-tune;
 \item an uneven allocation that gives 150 neurons to the first layer, the least compressible
   one (Section~\ref{sec:coef}).
\end{itemize}
On the acoustics cell every variant is in a narrow band, 0.0032 to 0.0045 against the
model's 0.0026, at 5\% of the floating point numbers, and the slow fine-tune is best. On the Euler cell, fine-tuning the bases takes the 100-neuron leaf from 0.0264 to 0.0161 at the
held-out times. That is within 24\% of the model and close to the ladder's 450-neuron leaf, at a
fifth of its floating point numbers. On both cells, then, a 100-neuron leaf costs a factor of 1.2--1.5 in error
for a factor of 20 in storage, provided the fine-tune may change the bases.

\subsubsection*{Cost of evaluation}
Evaluating one leaf on its own cell takes 2.4--3.3~ms per frame, against 3.0--4.0~ms for the
full model on the same points. A single cell therefore gives no speed-up. The gain of the reduction lies in the
folded storage and in evaluating hundreds of cells in one packed kernel: the per-cell
leaves are up to $68\times$ faster than the full model at $1000^2$ points on the Euler problem.

\section{Comparison with hypernetwork-SIREN, DeepONet and FNO}
\label{sec:baselines}

To test whether spatially local mini-batching depends on the low-rank structure, we trained
three baselines of the \lrnr{}'s size ($1.25\times10^6$ parameters) on the periodic acoustics
data:
\begin{itemize}[nosep]
 \item a hypernetwork-SIREN, in which a $2\times128$ hypernetwork of $t$ emits every weight and
   bias of a four-hidden-layer sine network (the \lrnr{} without the low-rank factorization);
 \item a DeepONet \citep{lu2021deeponet}, in which a network of $t$ produces $p=100$
   coefficients per variable and a four-layer sine INR of $\bx$ produces $p$ basis functions;
 \item a two-dimensional FNO \citep{li2021fno} with four spectral
   layers of 12 modes on a $64\times64$ grid and $(x,y,t)$ as input channels.
\end{itemize}
The protocol is in Appendix~\ref{sec:bl_protocol}.

\subsubsection*{Architectures at a matched budget}
The matched parameter count gives the models very different shapes. The hypernetwork-SIREN
spends 128 parameters of its last hypernetwork layer on every weight it generates, so its INR
can be only 48 neurons wide. The DeepONet's INR is 608 wide and the FNO has 32 channels. The
\lrnr{} has a 2000-wide INR whose weights depend on $t$, because its hypernetwork emits
$2rL=500$ coefficients instead of $1.2\times10^7$ weights.

\subsubsection*{Stability under global and local sampling}
\begin{table}[h]\centering
\begin{adjustbox}{max width=\linewidth}\begin{tabular}{llccccccc}\toprule
 & & \multicolumn{5}{c}{20{,}000 steps, learning rate} & \multicolumn{2}{c}{400{,}000 steps, learning rate} \\
\cmidrule(lr){3-7}\cmidrule(lr){8-9}
Model & Sampler & $10^{-4}$ & $3\times10^{-4}$ & $10^{-3}$ & $3\times10^{-3}$ & $10^{-2}$ & $10^{-4}$ & $3\times10^{-4}$ \\\midrule
Hyper-SIREN & Global, adaptive & 26.8\% & 19.8\% & 100.8\% & 97.4\% & div. & 6.7\% & 5.9\% \\
Hyper-SIREN & Global, uniform & 17.2\% & 16.3\% & 96.0\% & 96.0\% & div. & 5.0\% & 6.0\% \\
Hyper-SIREN & Local, adaptive & 80.4\% & 85.2\% & 96.6\% & div. & div. & 16.3\% & 96.7\% \\
Hyper-SIREN & Local, uniform & 79.7\% & 96.1\% & div. & div. & div. & 24.9\% & -- \\
DeepONet & Global, adaptive & 89.6\% & 80.1\% & 85.7\% & div. & div. & -- & -- \\
DeepONet & Global, uniform & 84.4\% & 77.4\% & 78.1\% & div. & div. & 56.5\% & 47.5\% \\
DeepONet & Local, adaptive & 88.0\% & 94.1\% & 95.9\% & div. & div. & 77.1\% & 77.9\% \\
DeepONet & Local, uniform & 87.2\% & 92.2\% & 93.3\% & div. & div. & 76.1\% & 74.3\% \\
FNO & Global, adaptive & 36.3\% & 20.3\% & 16.3\% & 15.8\% & 16.0\% & -- & -- \\
FNO & Global, uniform & 38.3\% & 18.2\% & 16.3\% & 15.7\% & div. & 15.6\% & 15.6\% \\
FNO & Local, adaptive & 80.2\% & 50.0\% & 26.6\% & 46.8\% & 96.0\% & 16.8\% & 16.0\% \\
FNO & Local, uniform & 80.6\% & 46.5\% & 23.5\% & 31.4\% & 96.0\% & 17.1\% & 16.0\% \\
\midrule
\lrnr{} & Local, adaptive & 34.4\% & 63.3\% & 96.2\% & 96.2\% & div. & -- & -- \\
\lrnr{} & Local, uniform & 43.6\% & 78.4\% & 96.1\% & 96.1\% & div. & -- & -- \\
\bottomrule\end{tabular}\end{adjustbox}
\caption{Relative $L^2$ error of the baselines on the periodic acoustics problem (area-weighted, frames 0/20/40/60/80) after a 20{,}000-step learning-rate sweep with global or local batches, and after 400{,}000 steps. Adaptive sampling is the \lrnr{}'s sampler of record (frames by the square root of their point count, cells by point count with the dense cells boosted; global batches draw points uniformly within a frame); uniform sampling draws frames and cells uniformly and global points uniformly over the area; ``div.'' marks a loss that became non-finite or exceeded $10^3$ times its initial value. The \lrnr{} rows are the same 20{,}000-step sweep with local batches (the same divergence rule; the \lrnr{} runs are not stopped at divergence, and at $10^{-2}$ the loss returns to that of the zero predictor); along its ladder (a first cosine stage at $10^{-4}$, then constant $10^{-5}$) it reaches 11.2\% after 97{,}400 steps and 4.9\% after 389{,}600 steps.}\label{tab:baselines}\end{table}
A sweep of five learning rates over 20{,}000 steps (Table~\ref{tab:baselines}, first five
columns) yields distinguishable results for the baselines. With global sampling the FNO did not diverge at any learning rate and reached 16--20\%; the
hypernetwork-SIREN diverged at $10^{-2}$ and stalled at the zero predictor at $10^{-3}$ and
$3\times10^{-3}$, a usable window of one decade; the DeepONet diverged from $3\times10^{-3}$ up
and stayed above 80\%. With the local sampler, one cell of the
$10\times10$ tiling per step as for the \lrnr{}, all three degrade: the hypernetwork-SIREN from
20\% to 80\% at its best rate, the DeepONet from 80\% to 88\%, and the FNO, scored on grid
patches since it cannot consume a cell, from 16\% to 27\%.

\subsubsection*{Matched sampler, data and steps}
The main comparison trains each baseline in the same way as the \lrnr{}, one cell per
step with frame and cell drawn by the solver's point density, for 400{,}000 steps at the two
best local rates of the sweep. The reference is the \lrnr{}'s own ladder after the same number
of steps (stage 4).
The hypernetwork-SIREN reaches 16.3\% at $10^{-4}$ and stays at 97\% at $3\times10^{-4}$, the
FNO 16.0\% and 16.8\%, and the DeepONet 77--78\% (Table~\ref{tab:baselines}, last two columns). The \lrnr{}
is at 4.9\%, a factor of 3.3 below the best baseline. Trained globally, the baselines can fit the problem. With a budget six times larger, a single cosine cycle of $2.34\times10^6$
steps, the hypernetwork-SIREN reaches 3.7\% against the \lrnr{}'s final 1.7\%, a factor of two that we attribute to the wider INR and the
training procedure. The \lrnr{}'s advantage is therefore largest under local sampling.

\subsubsection*{Global batches for the \lrnr{}}
Global batches are more accurate for the \lrnr{} as well. The same four-stage ladder, with each
step taking 4{,}096 random points of one frame instead of the points of one cell (and no
projector), reaches 8.4, 2.1, 1.6 and 1.4\% after stages 1 to 4, against 11.2, 7.3, 5.4 and 4.9\%
with local batches; with uniform sampling the two ladders end at 1.2\% and 6.1\%. A global step
sees about sixteen times the data of a local one, spread over the domain, and in this comparison
it lowers the error by a factor of 3.5 to 5 at a fixed number of steps. Local mini-batching is
therefore not a more accurate way to train the \lrnr{}. Under local batches the \lrnr{} loses less accuracy than the baselines. At the matched size and with the same local steps, the
baselines' errors are three or more times the \lrnr{}'s. Settings in which a local step can
be worth this loss are discussed in Section~\ref{sec:discussion}.

\subsubsection*{Width of the hypernetwork-SIREN}
\begin{table}[h]\centering
\begin{adjustbox}{max width=\linewidth}\begin{tabular}{rr*{9}{r}}\toprule
 & & \multicolumn{5}{c}{Adaptive sampling} & \multicolumn{4}{c}{Uniform sampling} \\
\cmidrule(lr){3-7}\cmidrule(lr){8-11}
 & & \multicolumn{2}{c}{Global} & \multicolumn{2}{c}{Local} & & \multicolumn{2}{c}{Global} & Local & \\
\cmidrule(lr){3-4}\cmidrule(lr){5-6}\cmidrule(lr){8-9}\cmidrule(lr){10-10}
Width & Parameters & $10^{-4}$ & $3\times10^{-4}$ & $10^{-4}$ & $3\times10^{-4}$ & Local $\div$ global & $10^{-4}$ & $3\times10^{-4}$ & $10^{-4}$ & Local $\div$ global \\\midrule
48 & 964,531 & 6.7\% & 5.9\% & 16.3\% & 96.7\% & 2.8 & 5.0\% & 6.0\% & 24.9\% & 5.0 \\
96 & 3,695,203 & 3.3\% & 2.9\% & 11.1\% & 96.5\% & 3.7 & 3.0\% & 2.8\% & 14.4\% & 5.1 \\
192 & 14,506,435 & 1.9\% & 1.6\% & 7.3\% & 96.2\% & 4.7 & 1.4\% & 1.0\% & 7.7\% & 7.5 \\
384 & 57,528,451 & 1.1\% & 0.8\% & 6.3\% & 96.2\% & 7.9 & 1.1\% & 0.7\% & 9.0\% & 13.6 \\
\bottomrule\end{tabular}\end{adjustbox}
\caption{The hypernetwork-SIREN at four INR widths on the periodic acoustics problem after 400{,}000 steps, with global and with local batches under adaptive and under uniform sampling: relative $L^2$ error (area-weighted, frames 0/20/40/60/80) at two learning rates, and the ratio of the best local to the best global error. Width 48 is the matched-budget model of Table~\ref{tab:baselines}; the \lrnr{} ($1.25\times10^6$ parameters) reaches 4.9\% with the local sampler of record after the same number of steps.}\label{tab:bl_width}\end{table}
At the matched budget the hypernetwork-SIREN's INR is 48 wide, close to the \lrnr{}'s rank of
50, so it is possible that a narrow network tolerates local steps for the same reason the
\lrnr{} does. Table~\ref{tab:bl_width} widens it to 96, 192 and 384 at the same number of
steps. Under both samplers the error falls with width, but it falls faster under global
sampling: the best local error is 2.8 times the best global one at width 48, 3.7 times at 96,
4.7 times at 192 and 7.9 times at 384. The degrees of freedom that a wider network adds are the
ones that the local steps interfere with. In absolute terms the 384-wide model, with
$5.75\times10^7$ parameters, 46 times the \lrnr{}'s, reaches 6.3\% under local sampling against
the \lrnr{}'s 4.9\%. Under global sampling it reaches 0.79\%, with each step taking 4{,}096 points
spread over the domain, about sixteen times the data of a local step. With uniform sampling the
ratios are larger and grow faster with width (5.0, 5.1, 7.5 and 13.6), and the 384-wide model
reaches 9.0\% with local batches against 0.66\% with global ones.

\subsubsection*{Width of the \lrnr{}}
\begin{table}[h]\centering
\begin{adjustbox}{max width=\linewidth}\begin{tabular}{rr*{8}{r}}\toprule
 & & \multicolumn{4}{c}{Adaptive sampling} & \multicolumn{4}{c}{Uniform sampling} \\
\cmidrule(lr){3-6}\cmidrule(lr){7-10}
Width & Parameters & Local & No projector & Global & Local $\div$ global & Local & No projector & Global & Local $\div$ global \\\midrule
250 & 217,756 & 33.3\% & 18.0\% & 3.0\% & 11.1 & 67.9\% & 69.2\% & 3.2\% & 21.3 \\
500 & 367,756 & 6.9\% & 7.1\% & 1.5\% & 4.7 & 13.0\% & 16.4\% & 1.3\% & 10.1 \\
1000 & 667,756 & 5.1\% & 5.1\% & 0.9\% & 5.5 & 6.9\% & 7.3\% & 0.9\% & 7.5 \\
2000 & 1,267,756 & 4.7\% & 3.8\% & 0.9\% & 5.4 & 5.7\% & 5.8\% & 0.8\% & 7.2 \\
\bottomrule\end{tabular}\end{adjustbox}
\caption{The \lrnr{} at four widths on the periodic acoustics problem after 400{,}000 steps of one cosine run (rank 50, projector rank $0.4\times$ width): relative $L^2$ error (area-weighted, frames 0/20/40/60/80) with local batches at learning rate $10^{-4}$, with and without the projector (no projection misfit, no projector orthonormality penalty), and with global batches (best of $10^{-4}$ and $3\times10^{-4}$; global batches use no projector). Adaptive sampling is the sampler of record; uniform sampling draws frames and cells uniformly and global points uniformly over the area. Local runs at $3\times10^{-4}$ stall at 96\% at every width. One seed; Table~\ref{tab:proj_seeds} repeats widths 500 and 1000 over five seeds.}\label{tab:lrnr_width}\end{table}
Table~\ref{tab:lrnr_width} repeats the width sweep for the \lrnr{} at widths 250 to 2000, with
local batches with and without the projector and with global batches, under the adaptive
sampler of record and under uniform sampling. Global batches are more accurate for the \lrnr{}
too, but unlike the hypernetwork-SIREN's, its ratio of local to global error does not grow with
width once the width is 500 or more: 4.7, 5.5 and 5.4 with adaptive sampling and 10.1, 7.5 and
7.2 with uniform sampling at widths 500, 1000 and 2000. At width 250 the local errors are large (33\% and 68\%). The \lrnr{}'s local error falls with width, from 6.9\% at width 500 to 4.7\% at
width 2000 with adaptive sampling, while the hypernetwork-SIREN's falls from 16.3\% to 6.3\% at
sixty times the parameters. At comparable size and under local batches, the \lrnr{} at width 1000, with $6.7\times10^5$ parameters, reaches 5.1\%, against 16.3\% for the
hypernetwork-SIREN with $9.6\times10^5$. The projector does not change this. The no-projector columns differ from the local ones in both directions, by amounts comparable to the
seed-to-seed spread of Table~\ref{tab:proj_seeds}, except at width 250, where the run without it
is better under adaptive sampling (18\% against 33\%).

\subsubsection*{Uniform sampling}
Every comparison in this section was repeated with uniform sampling, in which frames and cells
are drawn uniformly and global batches draw their points uniformly over the area
(Tables~\ref{tab:baselines}, \ref{tab:bl_width} and~\ref{tab:lrnr_width}). The conclusions do not
change. Local training of the \lrnr{} is more accurate with the adaptive sampler. Its four-stage ladder ends at 4.9\% with the adaptive sampler against 6.1\% with uniform sampling, and the width-2000 run at 4.7\% against 5.7\%. It helps the local hypernetwork-SIREN still more, which reaches 24.9\% after
400{,}000 steps with uniform sampling against 16.3\%. The DeepONet and the FNO hardly depend on the
sampler (74--76\% and 16--17\% with local batches). With global batches uniform sampling is as good
as the adaptive one or slightly better, for the \lrnr{} (1.2\% against 1.4\% after four stages)
and for the hypernetwork-SIREN. Under uniform sampling, the \lrnr{} is still stable over a wider
range of learning rates than the INR baselines: it stalls at the zero predictor from $10^{-3}$ and
diverges only at $10^{-2}$, while the hypernetwork-SIREN with local batches diverges from
$10^{-3}$.

\subsubsection*{Comparisons between training times}
The runs of this section were also evaluated at the 80 times halfway between the training
frames, computed as for Table~\ref{tab:heldout}.
Tables~\ref{tab:baselines_heldout}, \ref{tab:bl_width_heldout} and~\ref{tab:lrnr_width_heldout}
in Appendix~\ref{sec:morefigs} give the results in the layout of Tables~\ref{tab:baselines},
\ref{tab:bl_width} and~\ref{tab:lrnr_width}. For 153 of the 160 runs the halfway error is within
5\% of the error at the 81 training times. The runs whose halfway error is higher by more than 5\%
are the six with training errors below 0.8\%: the \lrnr{} with global batches at widths 1000 and
2000 and learning rate $3\times10^{-4}$, from 0.63--0.76\% to 0.82--0.89\%, and the 384-wide
hypernetwork-SIREN with global batches, from 0.65\% to 0.74\% with uniform sampling and from
0.71\% to 1.0\% with adaptive sampling. Of the pairs of runs whose errors on the training data are
below 90\% and differ by more than 10\%, 99.9\% are in the same order at the halfway times, and
the reversals are all among runs with errors near 1\%. The conclusions of this section therefore
hold at the halfway times. In the matched comparison with local batches after 400{,}000 steps,
the \lrnr{} reaches 3.1\% at the halfway times, against 13.7\% for the hypernetwork-SIREN, 15.5\%
for the FNO and 74.9\% for the DeepONet. The ratio of the hypernetwork-SIREN's best local to best
global error grows with width, 2.7, 3.8, 4.6 and 5.1 at widths 48 to 384, while that of the
\lrnr{} does not, 4.6, 4.3 and 3.9 at widths 500, 1000 and 2000. The 384-wide hypernetwork-SIREN
with global batches reaches 0.74--1.0\% at the halfway times, against 0.82--0.89\% for the
\lrnr{} with global batches at width 2000 and 1.8\% for the final \lrnr{}.

\subsubsection*{Sensitivity to the learning rate}
Under local sampling the three INR-based models differ in the range of learning rates at which
they remain stable and in their accuracy within it (Table~\ref{tab:baselines}, 20{,}000 steps).
The hypernetwork-SIREN and the DeepONet diverge from $3\times10^{-3}$ up; the \lrnr{} remains
stable through $3\times10^{-3}$, with a loss below twenty times its initial value, and diverges
only at $10^{-2}$, where its loss reaches $10^4$ times its initial value within 300 steps. Where
any of the three trains, the \lrnr{} is the most accurate, with 34\% at $10^{-4}$ against 80\% and 88\%,
and 63\% at $3\times10^{-4}$ against 85\% and 94\%. At $10^{-3}$ and $3\times10^{-3}$ the \lrnr{}
is stable but stalls at the zero predictor, 96\%, as the other two do at $10^{-3}$, and it does
not leave that state when the cosine schedule brings the rate down to $10^{-4}$ and below over
the last 6{,}000 steps, rates at which the run started at $10^{-4}$ keeps descending. A direct
test confirms that the stall is permanent on this time scale. Continued from the stalled
$10^{-3}$ and $3\times10^{-3}$ runs at a constant $10^{-4}$, the rate at which a fresh run trains,
the \lrnr{} stays at 96\% through 100{,}000 further steps. Its useful
range is therefore $10^{-4}$ to $3\times10^{-4}$, and its ladder trains every stage at $10^{-4}$ or
$10^{-5}$, where it did not diverge in about a hundred stages. Local sampling also moves the hypernetwork-SIREN's best rate. Under global sampling it is $3\times10^{-4}$, and under local sampling the same rate leaves it at the zero predictor after 400{,}000 steps at every
width from 48 to 384 (Table~\ref{tab:bl_width}). The FNO is the exception to the pattern. It is not an INR; it acts on a fixed $64\times64$ grid, and under local sampling its loss is scored on
grid patches rather than on the solver's points of one cell. It is stable at every rate with
global sampling and at all but $10^{-2}$ with local sampling, and at $10^{-3}$ and
$3\times10^{-3}$ it is more accurate than the INR-based models.

\subsubsection*{Negative effect of local steps on the baselines}
A local step sees about 250 points on one patch of width 0.2 at one time. In the
hypernetwork-SIREN every weight of the INR is a hypernetwork output and receives the cell's
gradient; a sine network has no spatial locality in its parameters, so fitting one patch moves
every neuron, including those serving the 99 other cells, and the next step partly undoes it.
The DeepONet's INR is equally global, and its two networks, which meet only in a dot product,
pull the coefficients for a given $t$ toward each cell in turn. A spectral layer is global by
construction, so a patch loss reaches every retained Fourier mode.

We posit that the \lrnr{} tolerates the same steps for two structural reasons.
First, the gradient of a step
reaches the shared bases only through the $r=50$ coefficients of each layer, a bottleneck that
keeps one patch from rewriting the network. Second, the
hypernetwork sees the time of the step, so the coefficients represent the time dependence and the
bases only have to be consistent across cells. Together these amount to a block coordinate
descent over space and time under a shared low-rank constraint. The per-cell projector is not among the reasons, since without it the \lrnr{} trains to the same accuracy under local batches
(Table~\ref{tab:lrnr_width}). This account is our reading of the measured gap; the runs do not
prove it.

\subsubsection*{Hidden-state rank}
\begin{figure}[tp]
\centering\stackh{0.29}
\figor{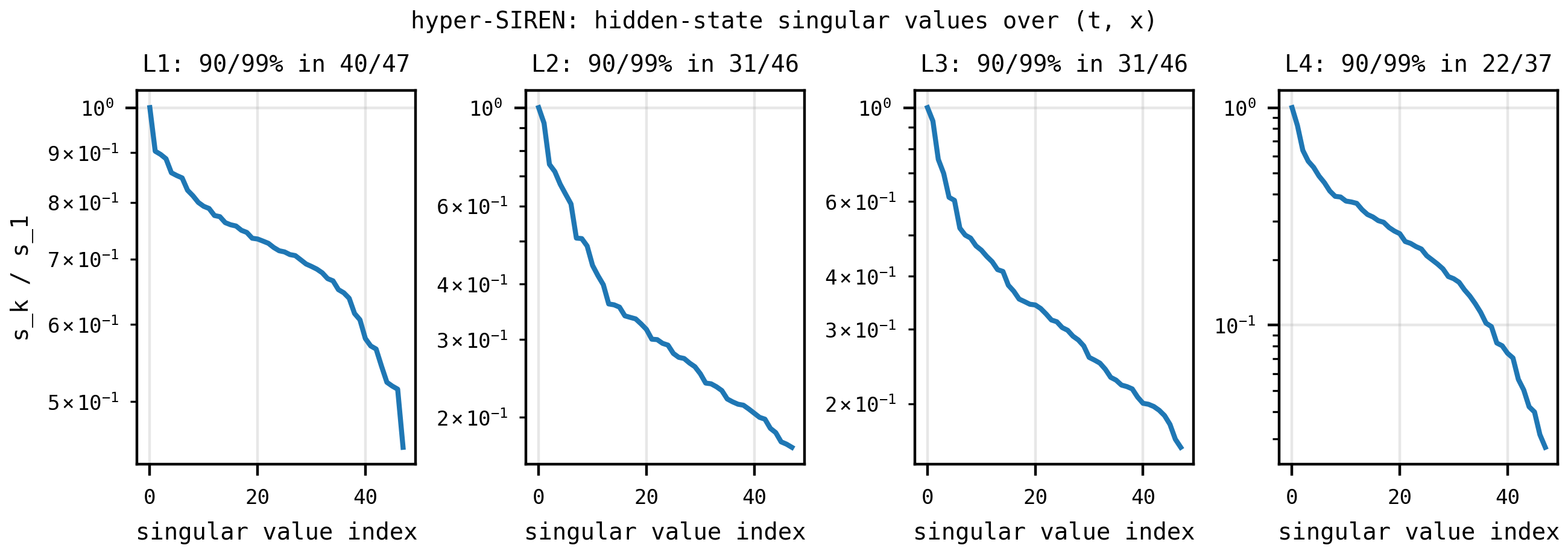}\\[2pt]\figor{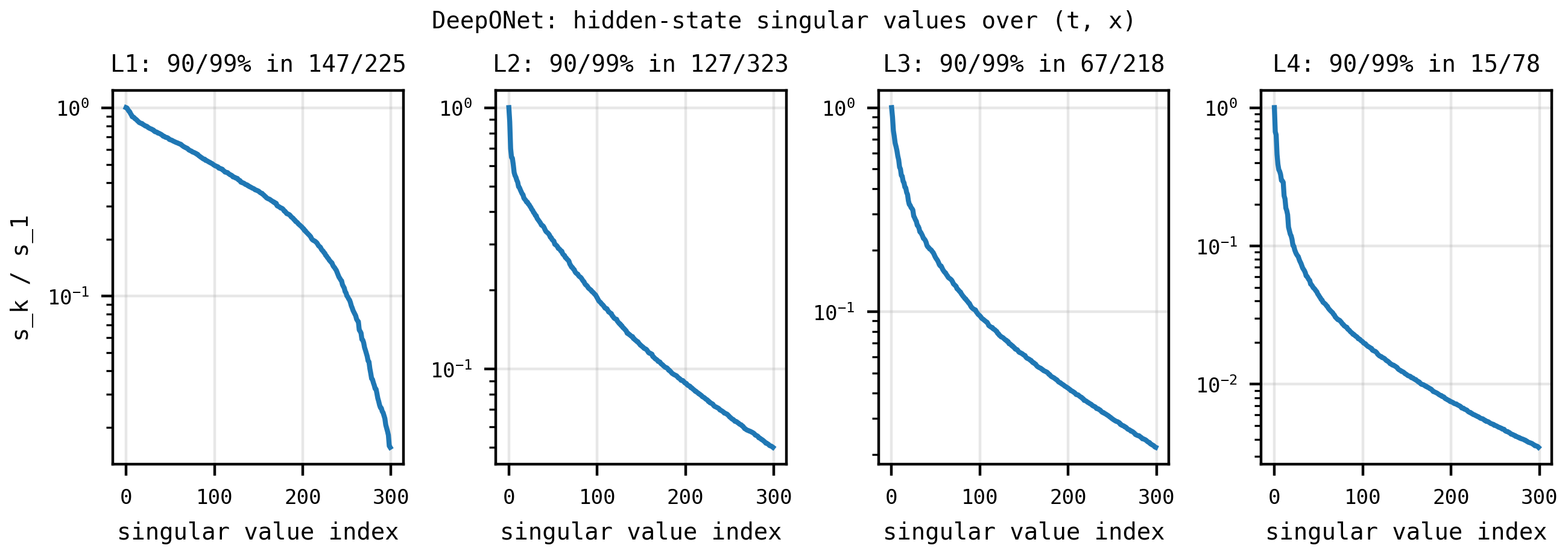}\\[2pt]\figor{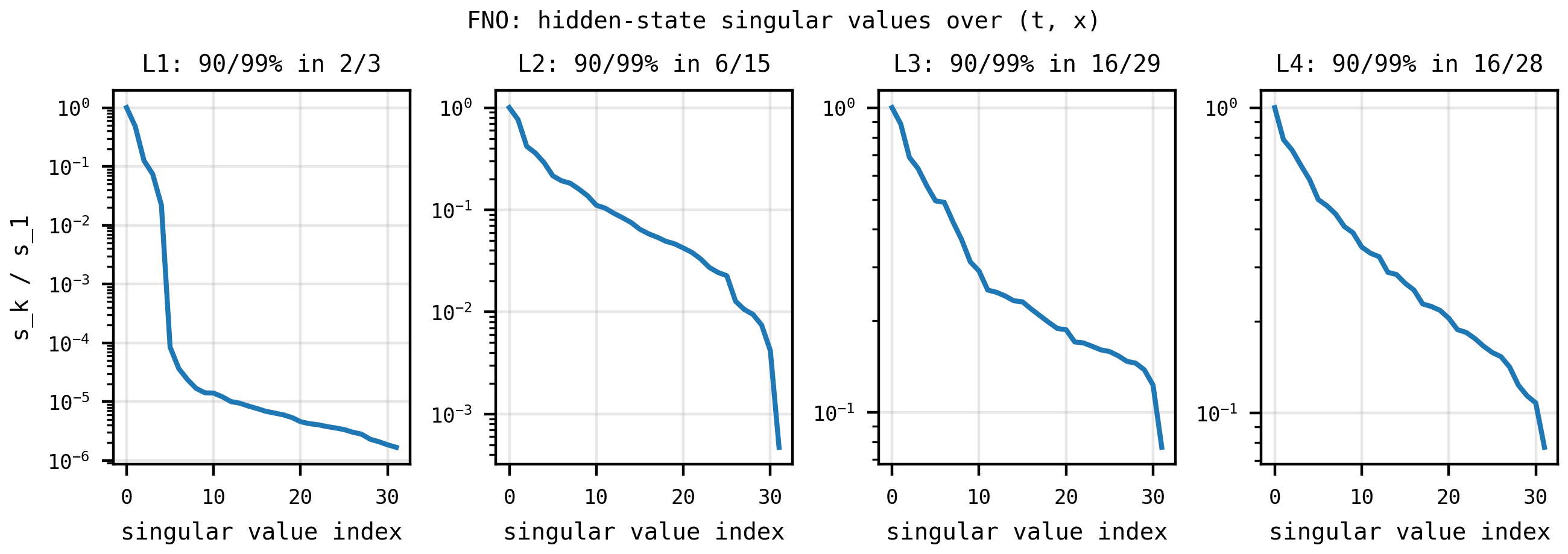}
\caption{Singular values of the hidden states over $(t,\bx)$ per layer for the three
baselines (normalized to the largest). In the \lrnr{} the pre-activations of every layer lie in
an $r=50$-dimensional space by construction; its hidden states are compared in
Table~\ref{tab:cellrank}.}
\label{fig:bl_svd}
\end{figure}

For each trained baseline we recorded the hidden states of every layer at 2000 fixed points and
the 81 times and computed the singular values of the $(81\times2000)\times{\rm width}$ matrix
(Table~\ref{tab:bl_svd}, Fig.~\ref{fig:bl_svd}), using the most accurate run of each baseline.
The DeepONet's 608-wide INR needs 147, 127, 67 and 15 directions for 90\% of the energy in its
four layers and 225, 323, 218 and 78 for 99\%, far from the \lrnr{}'s scale of 50. The
hypernetwork-SIREN uses 40, 31, 31 and 22 of its 48 directions, and the FNO's 32 channels need 2, 6, 16 and 16 directions for 90\%. In the \lrnr{} every
pre-activation lies in the 50-dimensional column space of $\bU_\ell$ by construction, and
Section~\ref{sec:coef} showed that the models use all 50 directions. The accurate baselines are
so narrow at this budget that their width bounds their hidden-state
rank, while the one baseline with a wide INR spreads its hidden states over hundreds of
directions and is the least accurate.

\begin{table}[h]\centering
\begin{adjustbox}{max width=\linewidth}\begin{tabular}{lrrrrrrrrr}\toprule
 & & \multicolumn{4}{c}{90\% energy, layer} & \multicolumn{4}{c}{99\% energy, layer} \\
\cmidrule(lr){3-6}\cmidrule(lr){7-10}
Model (best run) & Width & 1 & 2 & 3 & 4 & 1 & 2 & 3 & 4 \\\midrule
Hyper-SIREN & 48 & 40 & 31 & 31 & 22 & 47 & 46 & 46 & 37 \\
FNO & 32 & 2 & 6 & 16 & 16 & 3 & 15 & 29 & 28 \\
DeepONet & 608 & 167 & 137 & 42 & 8 & 236 & 259 & 103 & 38 \\
LRNR (by construction) & 2000 & \multicolumn{4}{c}{$\le 50$} & \multicolumn{4}{c}{$\le 50$} \\
\bottomrule\end{tabular}\end{adjustbox}
\caption{Hidden-state rank: the number of singular values carrying 90\% and 99\% of the energy of the $(81\times2000)\times{\rm width}$ hidden-state matrix in each layer, for the most accurate run of each baseline (hypernetwork-SIREN global $3\times10^{-4}$ at 2{,}337{,}600 steps; FNO global $3\times10^{-3}$ and DeepONet global $3\times10^{-4}$ at 97{,}400 steps).}\label{tab:bl_svd}\end{table}

\subsubsection*{Hidden-state dimension per cell}
\begin{table}[h]\centering
\begin{adjustbox}{max width=\linewidth}\begin{tabular}{llrrrrrrrrr}\toprule
 & & & \multicolumn{4}{c}{Per cell (median), layer} & \multicolumn{4}{c}{Whole domain, layer} \\
\cmidrule(lr){4-7}\cmidrule(lr){8-11}
Model & Training batches & Width & 1 & 2 & 3 & 4 & 1 & 2 & 3 & 4 \\\midrule
\lrnr{} & Local, adaptive & 2000 & 124 & 398 & 492 & 113 & 1942 & 1990 & 1964 & 640 \\
\lrnr{} & Global, adaptive & 2000 & 114 & 321 & 524 & 217 & 1940 & 1983 & 1985 & 1445 \\
\lrnr{} & Local, uniform & 2000 & 117 & 378 & 467 & 131 & 1922 & 1988 & 1955 & 745 \\
\lrnr{} & Global, uniform & 2000 & 110 & 327 & 508 & 175 & 1924 & 1985 & 1978 & 1197 \\
\addlinespace
Hyper-SIREN & Local, adaptive & 48 & 37 & 45 & 45 & 32 & 48 & 48 & 48 & 44 \\
Hyper-SIREN & Global, adaptive & 48 & 39 & 46 & 47 & 39 & 48 & 48 & 48 & 47 \\
Hyper-SIREN & Local, uniform & 48 & 36 & 40 & 44 & 34 & 48 & 48 & 48 & 45 \\
Hyper-SIREN & Global, uniform & 48 & 38 & 46 & 47 & 38 & 48 & 48 & 48 & 46 \\
\addlinespace
Hyper-SIREN & Local, adaptive & 96 & 57 & 78 & 71 & 50 & 96 & 96 & 94 & 84 \\
Hyper-SIREN & Global, adaptive & 96 & 56 & 89 & 92 & 66 & 96 & 96 & 96 & 92 \\
Hyper-SIREN & Local, uniform & 96 & 55 & 72 & 66 & 50 & 96 & 96 & 93 & 83 \\
Hyper-SIREN & Global, uniform & 96 & 57 & 90 & 95 & 76 & 96 & 96 & 96 & 90 \\
\addlinespace
Hyper-SIREN & Local, adaptive & 192 & 62 & 101 & 102 & 77 & 192 & 190 & 178 & 168 \\
Hyper-SIREN & Global, adaptive & 192 & 83 & 178 & 188 & 187 & 192 & 192 & 192 & 191 \\
Hyper-SIREN & Local, uniform & 192 & 63 & 96 & 81 & 67 & 192 & 190 & 174 & 154 \\
Hyper-SIREN & Global, uniform & 192 & 65 & 160 & 183 & 133 & 192 & 192 & 192 & 183 \\
\addlinespace
Hyper-SIREN & Local, adaptive & 384 & 59 & 104 & 114 & 103 & 379 & 380 & 329 & 302 \\
Hyper-SIREN & Global, adaptive & 384 & 84 & 320 & 373 & 376 & 384 & 384 & 384 & 384 \\
Hyper-SIREN & Local, uniform & 384 & 58 & 97 & 113 & 102 & 376 & 380 & 327 & 308 \\
Hyper-SIREN & Global, uniform & 384 & 105 & 366 & 378 & 378 & 384 & 384 & 384 & 382 \\
\bottomrule\end{tabular}\end{adjustbox}
\caption{Hidden-state dimension of models trained on the periodic acoustics problem: the number of singular directions needed to approximate a hidden layer's states to 1\% relative residual, sampled at 64 random points of a cell at all 81 training times (median over the 100 cells of a $10\times10$ tiling), and at 2{,}000 random points of the whole domain at all 81 times. The \lrnr{}s are the four-stage ladders; the hypernetwork-SIRENs are single 400{,}000-step runs at their better learning rate.}\label{tab:cellrank}\end{table}
The dimension relevant to the per-cell reduction in Section~\ref{sec:fast} is that of one
cell's hidden states over time, not that of the whole domain. Table~\ref{tab:cellrank} measures it
in the same way for the \lrnr{} and the hypernetwork-SIREN: the number of singular directions that
approximate a hidden layer's states to 1\% relative residual, on one cell of a $10\times10$ tiling
at all 81 training times, and on the whole domain. In every model a cell needs far fewer
directions than the domain; the \lrnr{}'s middle layers need about 400 to 500 of their 2000 per
cell and nearly all of them on the domain.

The training batches make the larger difference, and in the hypernetwork-SIREN the effect is
significant. Trained with local batches, its hidden states need fewer directions on each cell. At width 384 its
layers need 59 to 114 directions per cell, 15--30\% of the width, while the same network trained
with global batches needs 84 to 376, up to 98\% of the width in its deeper layers; at width 192
the figures are 62--102 against 83--188. The effect grows with width. At width 48, where the
network has little room, local and global training differ by a few directions. It appears under
both samplers and is equally strong or slightly stronger with uniform sampling than with the
adaptive one (58--113 against 59--114 per cell at width 384, 63--96 against 62--102 at width 192),
so it comes from the locality of the batches rather than from the refinement-driven choice of
cells. Local training also lowers the dimension over the whole domain, to 302--329 of 384 in the
last two layers, where global training uses all 384. In the \lrnr{}, by contrast, the per-cell
dimension of the first three layers is almost the same under local and global batches; local
batches lower only the last hidden layer, per cell (113 against 217 with adaptive sampling) and
over the domain (640 against 1445). The low dimension of a cell's hidden states is therefore not
specific to the \lrnr{}, and local mini-batching has a significant effect of the same
type on a generic INR's hidden states on each cell, a structure that
Section~\ref{sec:theory} shows to be available to any network with analytic activations. What is
specific to the \lrnr{} is that its pre-activations lie in an $r$-dimensional space by
construction, that its time dependence is captured by one or two hyper-modes
(Section~\ref{sec:hyper}), and that the per-cell reduction of its trained models reproduces them
(Section~\ref{sec:fast}).

\section{Conclusion and discussion}
\label{sec:discussion}

\subsubsection*{Conclusion}
Six time-dependent problems were fitted by the same low-rank neural representation with the
same training procedure using spatially local mini-batches to 1.1--2.3\% relative error on the solver's adaptive points, with a
tenth of the parameters of a dense network of the same width. The procedure's central
assumption, that a local projector of small rank exists at every hidden layer, is not
a happenstance of these problems but a property of analytic networks
(Section~\ref{sec:theory}), and the trained models demonstrate this: all 50 directions of the rank
are used globally, yet one cell's hidden states over the whole horizon reduce to 67--300
directions, and the per-cell leaves built on them are as accurate as or more accurate than the full model on the large cells. The LRNR coefficients depend on time through one or two hyper-modes along which the solution moves coherently
through its own evolution or re-times its fronts. Under the local sampler that the \lrnr{}
is trained with, baselines of the same size lose a factor of three or more, a hypernetwork-SIREN
46 times larger still falls short of the \lrnr{}, and at the matched size the hidden states of
the baselines show no comparable structure. At larger widths local training also has a
significant effect on the hypernetwork-SIREN: its hidden states need 15--30\% of the width on each
cell where global training uses up to 98\% of it, so local mini-batching partly induces the
per-cell structure that the \lrnr{} is based on, even in a generic INR.

\subsubsection*{Data access and cost per step: Global vs local}
At a fixed number of steps, global batches were more accurate for every model here, by a factor
of 3.5 to 5 for the \lrnr{}. The comparison counts steps, however, not the cost of a step or of
getting its data, and there are training setups in which a local step is much cheaper or a global
one is not available. Solver output is stored and produced by region. Adaptive-mesh data come as
patches, large three-dimensional or long-horizon data sets do not fit in memory, and a global
batch then needs random access across whole snapshots, while a local batch reads one contiguous
patch of one snapshot. When the data are partitioned across the processes of a parallel solver,
or produced in situ while the solver runs, each process holds only its own subdomain, and global
batches require moving data between processes that local batches do not. A local step also
uses only the points of one cell, and since a cell's hidden states are low-dimensional
(Sections~\ref{sec:theory} and~\ref{sec:fast}), it could in principle be computed with a reduced
per-cell network rather than the full width. Finally, the refinement-driven choice of cells
directs the steps to where the solution is difficult to represent, which already lowered the local \lrnr{}'s error
by about a fifth against uniform cells. In such settings a model that loses little under local
batches, as the \lrnr{} does and the baselines do not, may be worth the loss in accuracy per step.
We did not measure these costs; how much a local step saves, and whether it outweighs the factor
in accuracy, depends on the data layout and the hardware and is an open research topic.

\subsubsection*{Limitations and open questions}
Apart from the projector comparison, every result in this paper comes from a single training
run; the ladders are long enough that stage-to-stage variation is visible
(Section~\ref{sec:results}). Over five seeds the error of one configuration varies by about one
percentage point at width 1000 and by several at width 500 (Table~\ref{tab:proj_seeds}), so
single-run differences of that size are not resolved. The baseline comparison is on one problem. Only the hypernetwork-SIREN and the \lrnr{} were
trained at more than one width, and in those sweeps each was given two learning rates rather than a
sweep at the long budget. The learning-rate sweep of the \lrnr{} covers 20{,}000 steps only, and it shows
that the \lrnr{}, like the baselines, trains under local sampling only at rates of
$3\times10^{-4}$ or below. The per-cell reduction over-fits on small cells, and
the remedy proposed (no fine-tune, or a far shorter one) was identified but not tuned.
The feature-degree comparison on the acoustics problem suggests that raw coordinates may do
better than the degree-9 lift on periodic problems: in paired runs without the projector they
lowered the error by about 10\% at width 1000 over three seeds. The models reported here kept degree 9. The number of tiling levels and the cell size were fixed by design and not ablated; whether one level suffices is an open problem.
Global batches are more accurate per step for the \lrnr{} as for the baselines (1.4\% against
4.9\% after four stages), and the cost of a step under the two kinds of batches was not compared,
so the case made here for local mini-batching is that the \lrnr{} enables it, not
that it provides an advantage in accuracy. Finally, all problems are
two-dimensional and time is the only parameter; the extension of the sampler and
the projectors to parametric families is straightforward in principle and is
left to future work.

We propose the per-cell projector as a way to make the local structure explicit during training; it reached this aim only weakly here, and making it responsive to local features is an open research topic. The per-cell projection penalty, as formulated, is met mostly by directions shared across cells,
because each layer reads its input through a row basis that every cell shares
(Section~\ref{sec:training}). Projectors that encode local information would have to be denied this
route, for instance by building each cell's projector from its own hidden states, as the
FastLRNR's sampled bases are built. Fitting the projected pass itself to the data would then make
the projectors part of a local model and could give the per-cell reduction directly, without the
sampling step. Whether such a model keeps the accuracy of the plain one, and whether the weak
cell-adapted component that the current penalty leaves in the first and last hidden layers can be strengthened this way, are open problems. So is the question raised by the comparison of input features,
whether the degree-9 lift helps on any of the problems.

\section*{Acknowledgments}
The authors acknowledge the Research Infrastructure Services (RIS) group at Washington University
in St.~Louis for providing the computational resources and services needed to generate the
research results presented in this paper.
This work was partially supported by Institute of Information \& communications
Technology Planning \& Evaluation (IITP) grant funded by the Korea government (MSIT) (No. RS-2026-25587486).  K.~Lee acknowledges partial support
from the U.S. National Science Foundation under grant IIS 2338909. Claude
(Anthropic) was used to assist in conducting the experiments and preparing the
manuscript; all results were supervised by the authors, who take responsibility
for them.

\bibliographystyle{siamplain}
\bibliography{refs}

\appendix

\section{Proofs of the locality argument}
\label{sec:proofs}

\begin{proof}[Proof of Theorem~\ref{thm:width}]

Write $\bc=\bc_0+\bh$ and $z_{0,i}=\bu_i^{\top}\bc_0$. Entry $i$ of $\sigma(\bUbar\bc)$ is
$\sigma(z_{0,i}+\bu_i^{\top}\bh)$, and since $|\bu_i^{\top}\bh|\le\rho$,
\begin{equation}
 \sigma(\bu_i^{\top}\bc) = \sum_{k=0}^{K}\frac{\sigma^{(k)}(z_{0,i})}{k!} (\bu_i^{\top}\bh)^k + r_i(\bh),
 \quad |r_i(\bh)|\le\tau_K(\rho).
\end{equation}
By the multinomial theorem $(\bu_i^{\top}\bh)^k=\sum_{|\gamma|=k}\binom{k}{\gamma}\bu_i^{\gamma}\bh^{\gamma}$,
where $\gamma\in\mathbb{N}^q$ is a multi-index, $\bu_i^{\gamma}=\prod_j u_{ij}^{\gamma_j}$ and
$\bh^{\gamma}$ likewise. Collecting the monomials,
\begin{equation}
 \sigma(\bUbar\bc) = \sum_{|\gamma|\le K} \bh^{\gamma} \bv_{\gamma} + \br(\bh),
 \quad (\bv_{\gamma})_i=\frac{\sigma^{(|\gamma|)}(z_{0,i})}{|\gamma|!}
    \binom{|\gamma|}{\gamma}\bu_i^{\gamma},
 \quad \|\br(\bh)\|_2\le\sqrt{M} \tau_K(\rho).
\end{equation}
The vectors $\bv_{\gamma}\in\R^M$ do not depend on $\bh$: they are fixed by $\bUbar$, $\bc_0$ and
$\sigma$. Hence the polynomial part lies in $L_K=\mathrm{span}\{\bv_{\gamma}:|\gamma|\le K\}$,
whose dimension is at most the number of multi-indices, $\binom{q+K}{q}$, and every $\bs\in S$
is within $\sqrt{M} \tau_K(\rho)$ of $L_K$. Taking $K=K(n)$ makes $\dim L_K\le n$, and the width
is an infimum over all $n$-dimensional subspaces, so \eqref{eq:widthbound} follows. The sine
bound is Lagrange's remainder with $|\sigma^{(K+1)}|\le\omega^{K+1}$; the analytic bound is
Cauchy's estimate $|\sigma^{(k)}(z_0)|\le C_\sigma k!/\varrho^{k}$ summed as a geometric tail.
\end{proof}

\begin{proof}[Proof of Proposition~\ref{prop:layers}]

(a) By the mean value theorem applied entrywise, $\sigma(\bUbar\bc)-\sigma(\bUbar\bc^\ast)=\diag(\sigma'(\bxi)) \bUbar(\bc-\bc^\ast)$
for some intermediate arguments $\bxi$, and $\|\diag(\sigma'(\bxi))\|_2\le\omega$; then
$\bc'-\bc'^{\ast}=(\bD\bV'^{\top}(\bz-\bz^\ast);\bzero)$. (b) The next pre-activation changes by
$\bUbar'(\bD\bV'^{\top}(\bI-\bP)\bz;\bzero)$ and $\sigma$ is $\omega$-Lipschitz; $\bV'^{\top}(\bI-\bP)=\bzero$ when
$\mathrm{range}(\bV')\subset\mathrm{range}(\bP)$.
\end{proof}

\section{Stage ladders and training histories}
\label{sec:recipe}

\subsubsection*{Shared configuration}
All six models use the settings of Table~\ref{tab:hyper}. The five problems other than Euler
were trained with one configuration and no per-problem tuning: the same INR, sampler, loss
weights and schedule, with a $2\times128$ identity hypernetwork. The Euler model came first, and
the shared configuration grew out of it. It differs only in its $3\times5$ softplus
hypernetwork and in the way its ladder was run.

\subsubsection*{Execution of the ladders}
The ladders start with cosine stages and move to constant-rate stages once the per-stage gain of the cosine stages becomes small, as described in Section~\ref{sec:training}. The five shared-configuration
ladders follow one pattern. A single cosine stage from scratch is followed by constant-rate
stages, each averaged over its snapshots, and then by several stages that over-sample the end
frames. The final model is an average over boosted stages. On acoustics, advection and Burgers the
constant-rate phase lasts about fifteen stages and the boost seven to nine. On shock--bubble and
inclusions the constant-rate phase lasts five stages and the boost six to nine. On advection,
whose ladder was still improving, the final model is the average of the last boosted stage
alone.

The Euler ladder was the first to be run and is less regular. Its first twelve stages are cosine
stages, the second of them re-initialized. The tempered frame sampler is used from stage 13 on.
The snapshots of stage 16 are averaged, a further cosine stage starts from that average, and
three constant-rate stages follow, whose snapshot average is the final model, at 1.1\%. This ladder did not use the end-frame boost.

\begin{figure}[tp]
\centering\stackh{0.31}
\figor{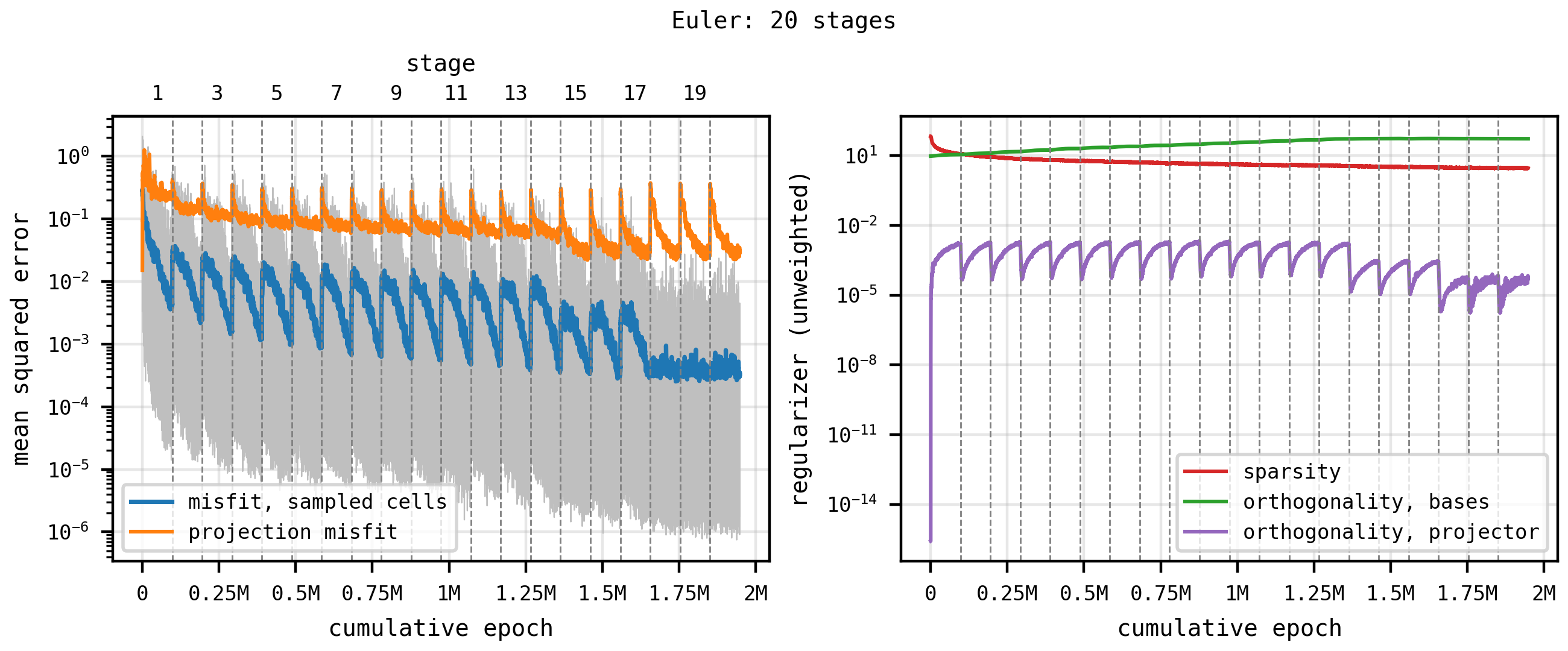}\\[2pt]\figor{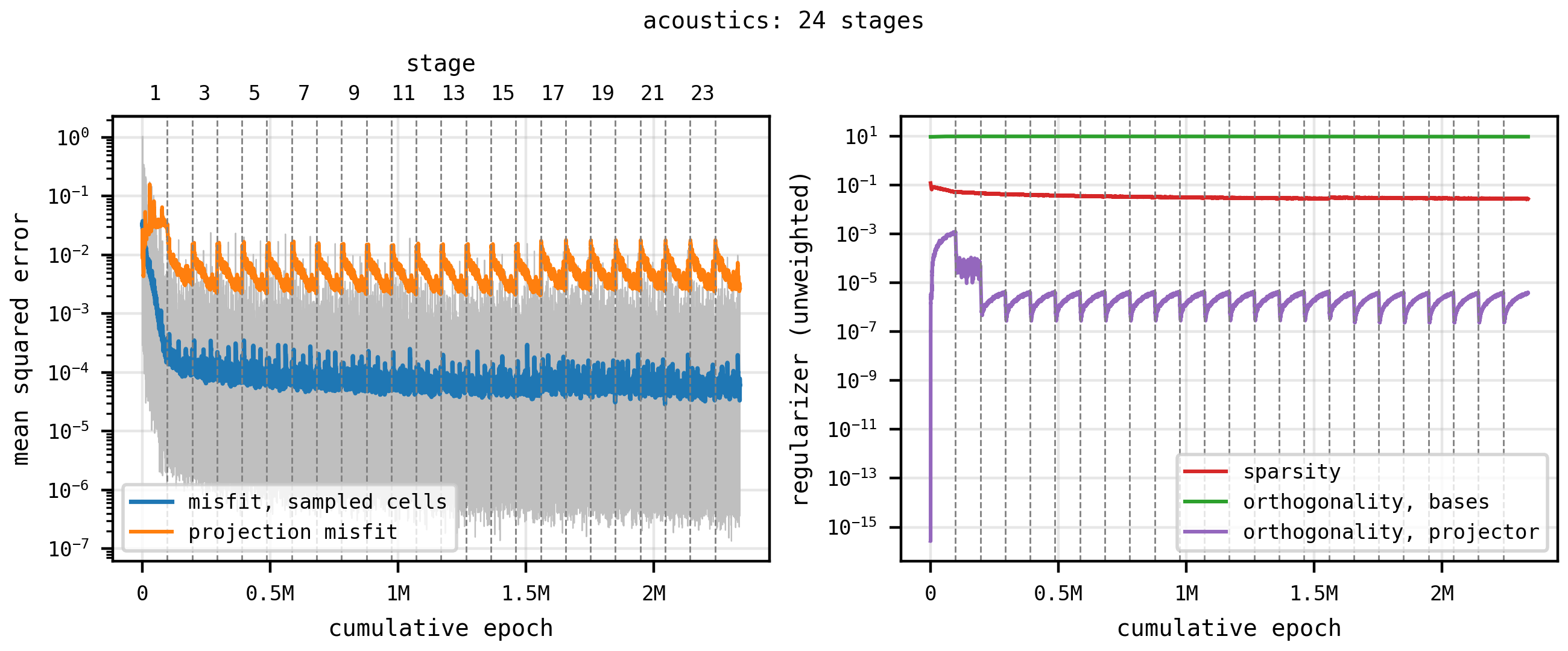}
\caption{Training histories over the whole ladders (Euler and acoustics; advection and
Burgers in Fig.~\ref{fig:loss2}, shock--bubble and inclusions in Fig.~\ref{fig:loss3}): the sampled-cell misfit (gray raw, blue smoothed) and the
projection misfit (orange), and the three regularizers (unweighted); dashed lines mark the stage
boundaries. The learning rate follows the stage ladders above: cosine stages decay from
$10^{-4}$, constant-rate stages run at $10^{-5}$.}
\label{fig:loss}
\end{figure}
\begin{figure}[tp]
\centering\stackh{0.31}
\figor{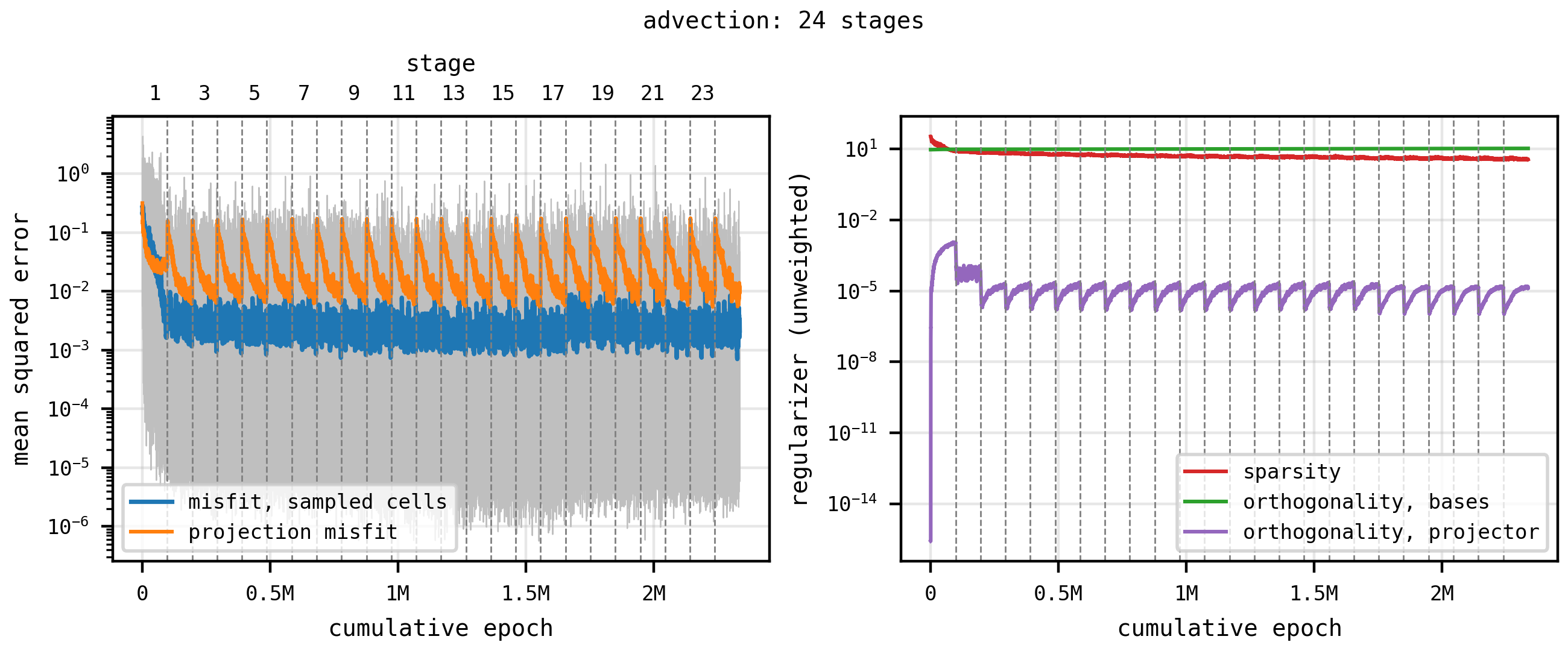}\\[2pt]\figor{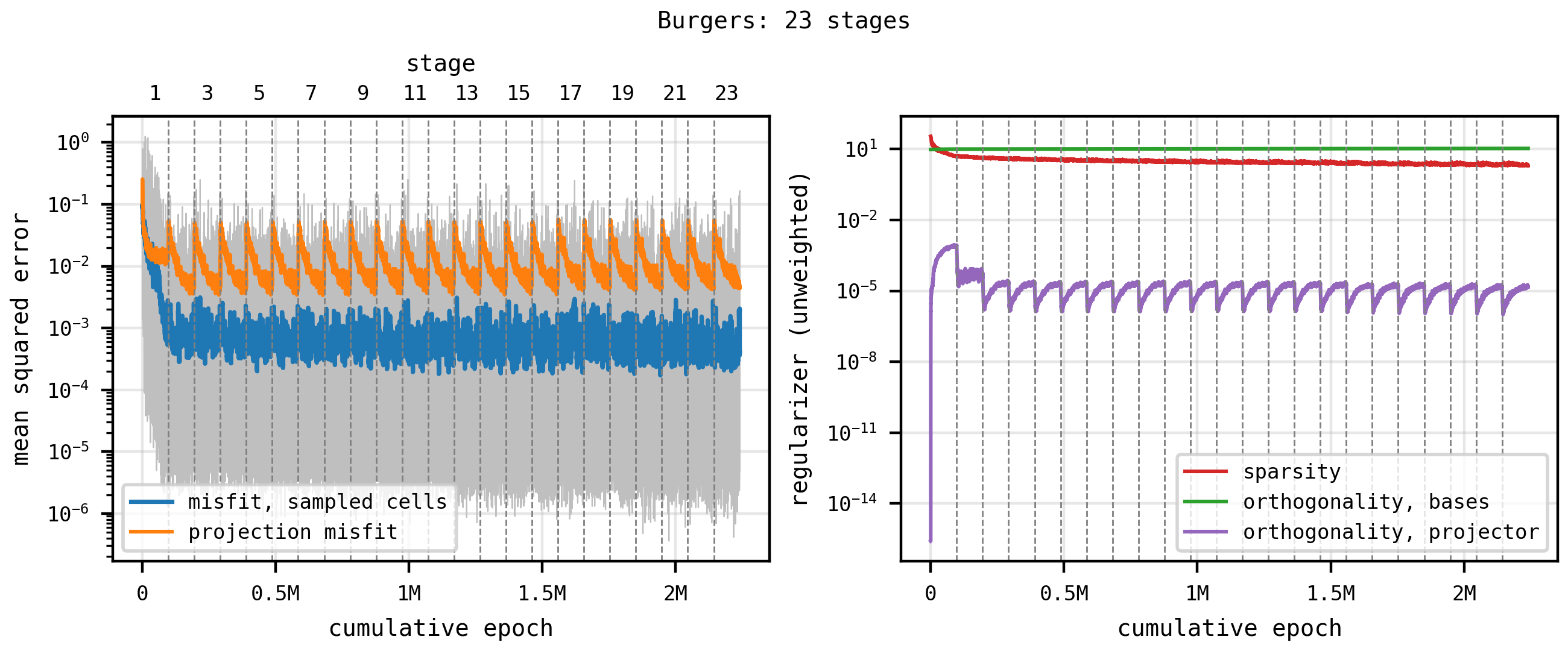}
\caption{Training histories of the advection and Burgers ladders (as Fig.~\ref{fig:loss}).}
\label{fig:loss2}
\end{figure}
\begin{figure}[tp]
\centering\stackh{0.31}
\figor{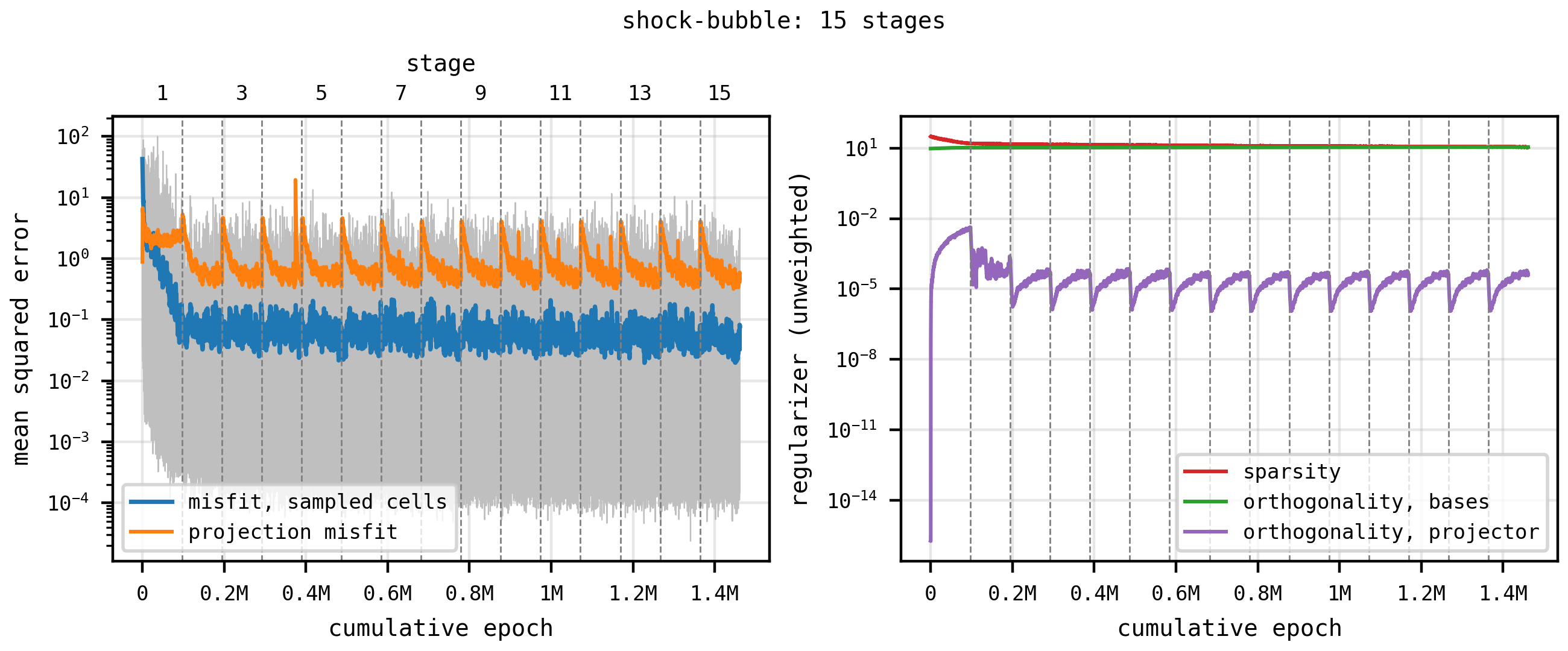}\\[2pt]\figor{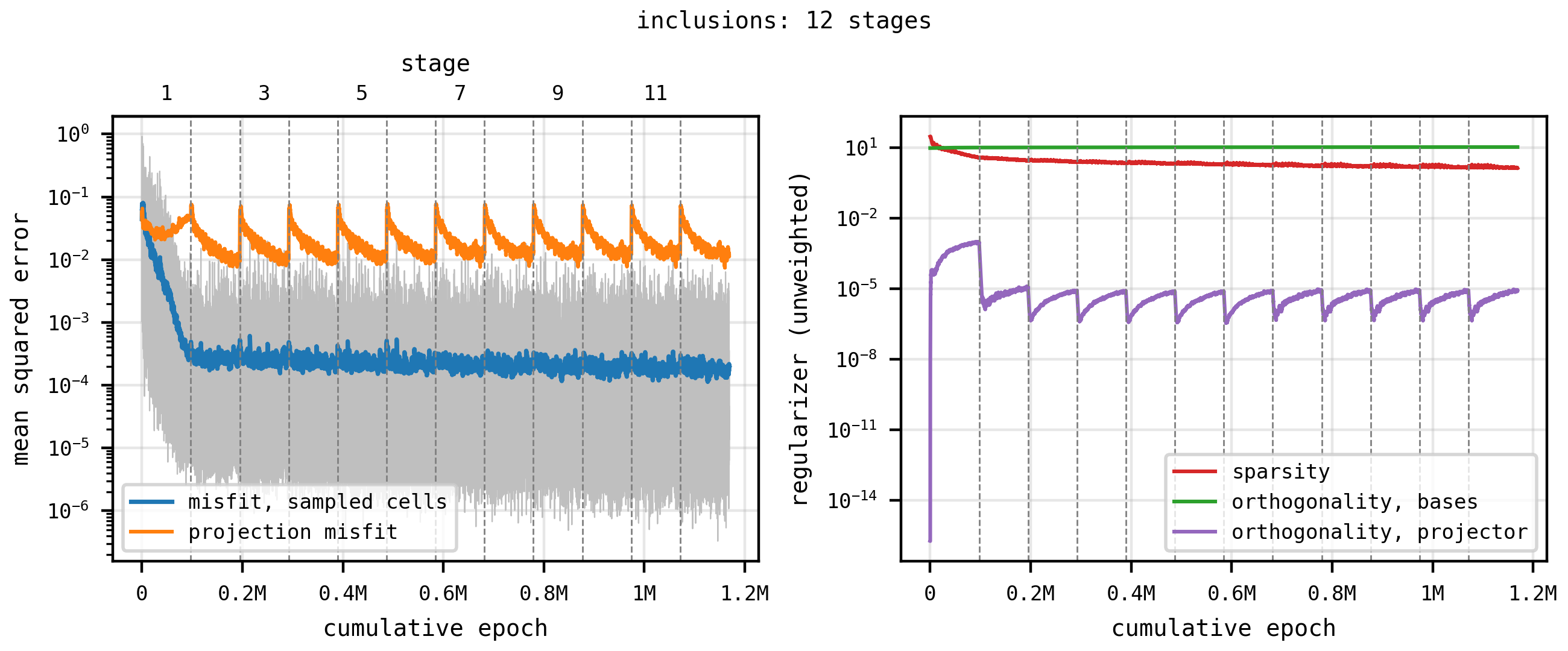}
\caption{Training histories of the shock--bubble and inclusions ladders (as Fig.~\ref{fig:loss}).}
\label{fig:loss3}
\end{figure}

\subsubsection*{Observations from the training histories}
Figures~\ref{fig:loss}--\ref{fig:loss3} show four consistent patterns.
\begin{itemize}[nosep]
 \item \emph{Most of the progress is early.} The sampled-cell misfit, averaged over a stage,
   falls by a factor of 2 to 17 over a ladder, most of it in the first three stages. On Euler it
   falls from $1.25\times10^{-3}$ to $7.5\times10^{-5}$ over 20 stages. The rest is a slow descent,
   which the constant-rate stages continue without spikes at the restarts.
 \item \emph{The projector is not a local basis.} The projection misfit settles
   within two stages at $10^{-3}$ to $10^{-2}$ and then follows the data misfit with a sawtooth at
   every restart, because the projector is refitted each time the INR changes. It remains well above the data misfit, which is the quantitative form of the finding that the learned
   projector is not a local basis (Section~\ref{sec:training}).
 \item \emph{The coefficient spectra keep reordering.} The sparsity penalty decreases steadily
   over every ladder, on Euler from 11.6 to 2.7, so the coefficients move toward geometric decay
   throughout training.
 \item \emph{The bases do not become orthonormal.} At its weight of $10^{-7}$ the orthogonality
   penalty on the bases stays at an unweighted 10 to 50 and on Euler grows from 11 to 53 across
   the ladder. Training does not drive the bases toward orthonormality, which agrees with the
   ablation in which enforcing it cost 34\% (Appendix~\ref{sec:design}). The projector's own
   orthonormality penalty, by contrast, stays at $10^{-5}$ to $10^{-6}$.
\end{itemize}

\section{Design choices and ablations}
\label{sec:design}

The training recipe in Section~\ref{sec:training} was settled by about sixty controlled
comparisons on the Euler problem, together with a smaller set on a parametric acoustics family
whose conclusions also held. Two observations summarize them. First, every change that moved
the Euler error by more than ten percent concerns where and when the data are visited (the
schedule, the sampler, the staging and the averaging) rather than the capacity of the model:
width beyond 2000 and rank beyond 50 did not help on these time-only problems. Second, the
components that keep the model low-dimensional (the low-rank factorization, the sparsity
penalty and the small hypernetwork) cost no accuracy at these ranks: rank 50 matches rank 200,
and the log-domain sparsity penalty is a useful regularizer even though it does not by itself
make the coefficient spectrum decay (Section~\ref{sec:coef}).

Each item below gives the comparison and the resulting choice. Errors are root-mean-square errors on
the Euler problem unless given in percent, in which case they are area-weighted relative $L^2$
errors; step counts refer to training steps.
\begin{itemize}
 \item \emph{Learning-rate schedule.} Cosine annealing with a decaying regularizer weight, against a
   constant rate with cyclic regularization, lowered the error from 0.323 to 0.230 after 22{,}000
   steps and from 0.219 to 0.102 after 97{,}400. This was the largest single effect, and cosine
   annealing was adopted.
 \item \emph{Staging.} No single-cycle run matched the staged model at an equal budget (0.071 against
   0.062, and 0.110 against 0.083). Restarting with a peak rate of $5\times10^{-5}$ lowered the
   error by 17\%, while $2.5\times10^{-5}$ was worse. The recipe uses ladders of restarted stages,
   with the peak rate halved once.
 \item \emph{Snapshot averaging.} Averaging the snapshots of a stage lowered the error by 20--22\% once
   a ladder had flattened, but raised it while the ladder was still descending, and averaging all ten snapshots was more accurate than averaging the last five. Averaging is used in the constant-rate stages.
 \item \emph{End-frame over-sampling.} On advection one stage took the error from 2.9 to 2.3\% (at
   $t=T$ from 4.7 to 2.5\%), and it lowered all three ladders by about 30\%. The final stages
   over-sample the end frames, where the error lies.
 \item \emph{Sampling.} Selecting cells by their current error was less accurate than uniform selection and did not
   find the Kelvin--Helmholtz billows. Using the solver's refinement as the prior lowered the error
   by 19\% after 22{,}000 steps, weighting frames by their point density lowered the error on the
   billows by 27\%, and tempering the frame weights to $p\propto\sqrt{N}$ took the error from 0.052
   to 0.038 while keeping the billow bands. The sampler draws by the solver's refinement,
   tempered.
 \item \emph{Polynomial features.} Chebyshev features replaced the quadratic monomials $(x,y,x^2,y^2,xy)$ of the earliest models because they are far better conditioned: on the Euler points the Gram matrix of the degree-3 features has condition number 23, against 9409 for monomials, whose $x^2$ and $x^3$ columns correlate at 0.986. In early Euler runs degree 9 was more accurate than degree 5 in every region, and in 399{,}000-step runs degrees 13 and 17 resolved the Kelvin--Helmholtz roll-ups that degree 9 did not; at 200{,}000 steps degree 17 reached 0.0123 in density over the Kelvin--Helmholtz box, against 0.0295 for the 97{,}400-step degree-9 model of that study. In a 22{,}000-step screen, Chebyshev degree 15 was worse than degree 9 (0.286 against 0.229), more so with a high first-layer frequency. On periodic acoustics, degrees 1
   (raw coordinates), 3 and 9 reached 8.2, 11.4 and 11.0\% after one stage, and after a second,
   averaged stage degree 1 reached 4.8\% against 7.3\% for degree 9. Degree 9 was kept for the
   models of this paper; on the periodic problem raw coordinates do better, since the sine input
   layer learns its own frequencies.
 \item \emph{Activation.} All-sine layers lowered the error against a mix of 25\% sine and 75\% $\tanh$ neurons by 11\%
   overall and 31\% on the billows after 22{,}000 steps (0.091 against 0.102 after 97{,}400). The
   INR uses sine activations throughout.
 \item \emph{First-layer frequency.} Raising $\omega_0$ from 30 to 60 on Euler changed the error by
   $+5$ to $+8\%$ on a single frame and was neutral on the full model; on the parametric family
   the values 3, 10, 30 and 100 were compared and 10 was best. The time-only problems use 30.
 \item \emph{Hypernetwork.} On a five-frame window, the $2\times128$ identity-output hypernetwork was more accurate than the $3\times5$ softplus one on the billows (0.023 against 0.097); a $3\times256$ network was 25\%
   worse and a $3\times512$ network diverged, and at rank 30 width 16 had a 26\% lower error than width 128.
   A larger hypernetwork did not lower the error, the identity output is marginally better, and the
   hypernetwork is kept small.
 \item \emph{Rank.} After 22{,}000 steps the error stopped improving beyond rank 30; after 97{,}400
   steps rank 50 matched rank 200 (0.102 against 0.104) and rank 20 reached 0.139. On advection
   rank 100 was 3.5 times worse after one stage. The models use rank 50.
 \item \emph{Width.} On Euler width 4000 reached 0.0135 against 0.0137 for width 2000, at twice the
   cost; on acoustics, advection and Burgers width 4000 reached the same floor in a third of the
   stages but did not go below it. The models use width 2000.
 \item \emph{Orthonormal bases.} At weight $10^{-4}$ the orthogonality penalty left both the bases
   (penalty 9.7) and the error (0.230) unchanged. At $10^{-2}$ it halved the penalty to 4.7 at an
   error of 0.237, and a per-layer rank allocation of the trained model kept about twenty fewer
   ranks. Enforcing orthonormality drove the penalty to 0.07 but raised the error by 34\%, and
   truncating the coefficients by index raised the error as much as before. The recipe keeps only a soft
   penalty, at weight $10^{-7}$.
 \item \emph{Sparsity penalty.} A weight of $10^{-3}$ was more accurate than $10^{-2}$ (0.190 against 0.209 at rank
   200) and left a smaller spectral tail; $10^{-2}$ saved 125 of the 1000 ranks of the rank-200
   model at 8\% in error. A plain (non-logarithmic) penalty collapsed the output. The
   scale-invariant log-domain penalty is kept at $10^{-3}$.
 \item \emph{Spatial projector.} Learned projectors did not become a usable local basis (they captured
   11\% of the hidden-state energy). As a regularizer, rank 300 at weight $10^{-2}$ had a 3--4\% lower error than rank 800 at $10^{-3}$, in half the time on one run. Removing the penalty altogether changes the
   error by less than the seed-to-seed spread (Table~\ref{tab:proj_seeds}). The penalty is kept in
   the recipe; it is met mostly through directions shared by all cells (Section~\ref{sec:training}),
   and local bases are built by sampling instead (Section~\ref{sec:fast}).
 \item \emph{Rank reduction.} A rank-20 model fine-tuned from a rank-50 one reached 0.156, against
   0.335 when trained from scratch; re-basing through an SVD was lossy even at full rank; and on
   the parametric family dropping coefficients by their contribution did not raise the error down to half the coefficients. Rank is reduced by dropping coefficients, not by an SVD.
\end{itemize}

\section{Baseline comparison protocol}
\label{sec:bl_protocol}
All baseline runs share one implementation, so that the three baselines
share the data pipeline, the samplers, the optimizer, the evaluation and the hidden-state
analysis; only the model class differs.

\subsubsection*{Data}
The periodic-acoustics trajectory as the \lrnr{} saw it: 81 frames of AMRClaw output at
$t_k = k/80$, each a scattered set of cell centers (the points of every patch not covered by a
finer one) with the three conserved variables $(p, u, v)$, 10{,}000--22{,}000 points per frame,
on the periodic square $[-1,1]^2$. The two INR baselines are trained on the solver points without interpolation. For the FNO, which needs a uniform grid, every frame is mapped once onto a
$64\times64$ grid by nearest neighbor (each grid node takes the value of the closest solver
point), and the FNO is trained on and evaluated through that grid.

\subsubsection*{Parameter matching}
The \lrnr{} of record for the problem has $1{,}267{,}756$ parameters:
$1{,}203{,}000$ in the bases of its five layers and $64{,}756$ in the $2\times128$
hypernetwork. The budget for each baseline was $1{,}251{,}500$ (1.3\% below the exact count,
which is immaterial here), and each architecture was grown
along its natural width until the next step would exceed it:
\begin{itemize}[nosep]
 \item Hypernetwork-SIREN: hypernetwork $1\to128\to128\to n_w$ with $\tanh$ hidden
   units emitting all $n_w$ weights and biases of a SIREN $2\to M\to M\to M\to M\to3$ with
   $\sin(30 \cdot)$ on the first layer and $\sin(\cdot)$ after; widths stepped by 8. The
   largest that fits is $M=48$ ($n_w=7{,}491$; $964{,}531$ parameters, 77\% of the budget;
   the next width, 56, overshoots). The generated weights are initialized SIREN-style through
   the last hypernetwork layer's bias (uniform in $\pm1/n_{\rm in}$ for the first layer,
   $\pm\sqrt{6/n_{\rm in}}/30$ after) with that layer's weights scaled by $10^{-2}$.
 \item DeepONet: network of $t$ $1\to128\to128\to 3p$ ($\tanh$), INR of $\bx$
   $2\to M\to M\to M\to M\to p$ with the same sine activations as the SIREN, output
   $u_i(\bx,t)=\sum_{j} b_{ij}(t) \tau_j(\bx)+c_i$, $p=100$; widths stepped by 16;
   $M=608$ fits ($1{,}25\times10^6$).
 \item FNO: lift $3\to c$ of the channels $(x,y,t)$ on the grid, four spectral
   layers with 12 retained modes in each direction and a $1\times1$ convolution bypass, GELU
   between them, projection $c\to128\to3$; channels stepped by 4; $c=32$ fits
   ($1{,}188{,}611$).
\end{itemize}

\subsubsection*{Samplers}
Two, applied identically to every model.
\emph{Global}: a frame is drawn with probability $\propto\sqrt{N_k}$ ($N_k$ its point count,
the \lrnr{}'s tempered frame density), then 4{,}096 of its points uniformly at random; the FNO
is scored on its whole grid.
\emph{Local}: the same frame draw, then one cell of the $10\times10$ tiling of the domain with
probability $\propto\sqrt{n_{k,c}}$ ($n_{k,c}$ the points of frame $k$ in cell $c$), and all of
that cell's points (about 250), the draw the \lrnr{}'s trainer makes
(Section~\ref{sec:training}), so that
the data seen per step and over a run match the \lrnr{}'s in distribution. For the FNO,
which cannot take a cell as input, ``local'' means the loss is scored on a random
$16\times16$ patch of the $64\times64$ grid (256 nodes, matching the cell's point count)
while the operator sees the whole grid.

\subsubsection*{Optimization}
Adam with cosine annealing from the stated rate to zero over the run, mean-squared error on
the sampled points, gradient-norm clipping at $10^3$, one draw per step. A run is recorded as
\emph{diverged} when its loss becomes non-finite or exceeds $10^3$ times its initial value, and
it is stopped there. Four budgets were run: (i) the \emph{sweep}, 20{,}000 steps at
$10^{-2},3\times10^{-3},10^{-3},3\times10^{-4},10^{-4}$ under both samplers, 30 runs
(Table~\ref{tab:baselines}); (ii) the best global configuration of each model at the
\lrnr{}'s stage length, 97{,}400 steps (hyper-SIREN 9.7\%, DeepONet 70.0\%, FNO 15.6\%); (iii) the
hyper-SIREN at the \lrnr{}'s full 24-stage budget, $2{,}337{,}600$ steps, in one cosine cycle;
(iv) the \emph{fair} comparison, local sampling, 400{,}000 steps at the two best local rates
of the sweep, $10^{-4}$ and $3\times10^{-4}$ (Table~\ref{tab:baselines}). The \lrnr{}
references are its own ladder: stage 1 (97{,}400 steps, 11.0\%), stage 4 (389{,}600 steps,
4.9\%), the final model (24 stages, 1.7\%), and a 20{,}000-step cosine stage run for the
sweep's budget (32.7\%).

\subsubsection*{Evaluation}
The same metric as everywhere in this paper: the relative $L^2$ error on the domain with
cell-area weights (each solver point weighted by the square of its nearest-neighbor spacing),
rms over the five frames $k=0,20,40,60,80$, all three variables pooled. The INRs
are evaluated at the solver points; the FNO's grid output is read at the solver
points by nearest neighbor, the inverse of the mapping used to build its training grid. The
intermediate evaluations printed during training use the same metric.

\subsubsection*{Hidden-state analysis}
After training, each model is evaluated at the 81 frame times on a fixed set of 2{,}000
uniformly random points (the FNO on its grid), the post-activation state of every layer is
collected into an $(81\times2{,}000)\times{\rm width}$ matrix per layer, centered, and its
singular values computed; the counts for 90\% and 99\% of the energy are reported with the
ratios $s_{50}/s_1$ and $s_{100}/s_1$.

\subsubsection*{Hardware and cost}
All baseline runs ran on 16-thread CPU nodes: 20{,}000 steps take 2 (hyper-SIREN), 2--12 (DeepONet) and 17 (FNO)
minutes; the 400{,}000-step fair runs 40 minutes, 1 hour and 6.5 hours respectively; the
$2.3\times10^6$-step hyper-SIREN 3.5 hours. The \lrnr{} stages ran on GPUs (2--4 hours per
stage).

\subsubsection*{Matched and unmatched quantities}
Parameters (to within the width step), the sampler and the data seen per step (fair runs),
the step budget, the optimizer family and the evaluation are matched. Not matched, and stated
where they bear on a number: the \lrnr{}'s stages restart the learning rate and average
snapshots, which the single-cycle baseline runs do not; a global baseline step sees 4{,}096
points where a local step sees about 250, so the global sweeps see roughly $16\times$ more
data per step; and the FNO's grid is a resampling of the data that the other models do not see.

\section{Additional figures and tables}
\label{sec:morefigs}

\begin{figure}[tp]
\centering
\figor{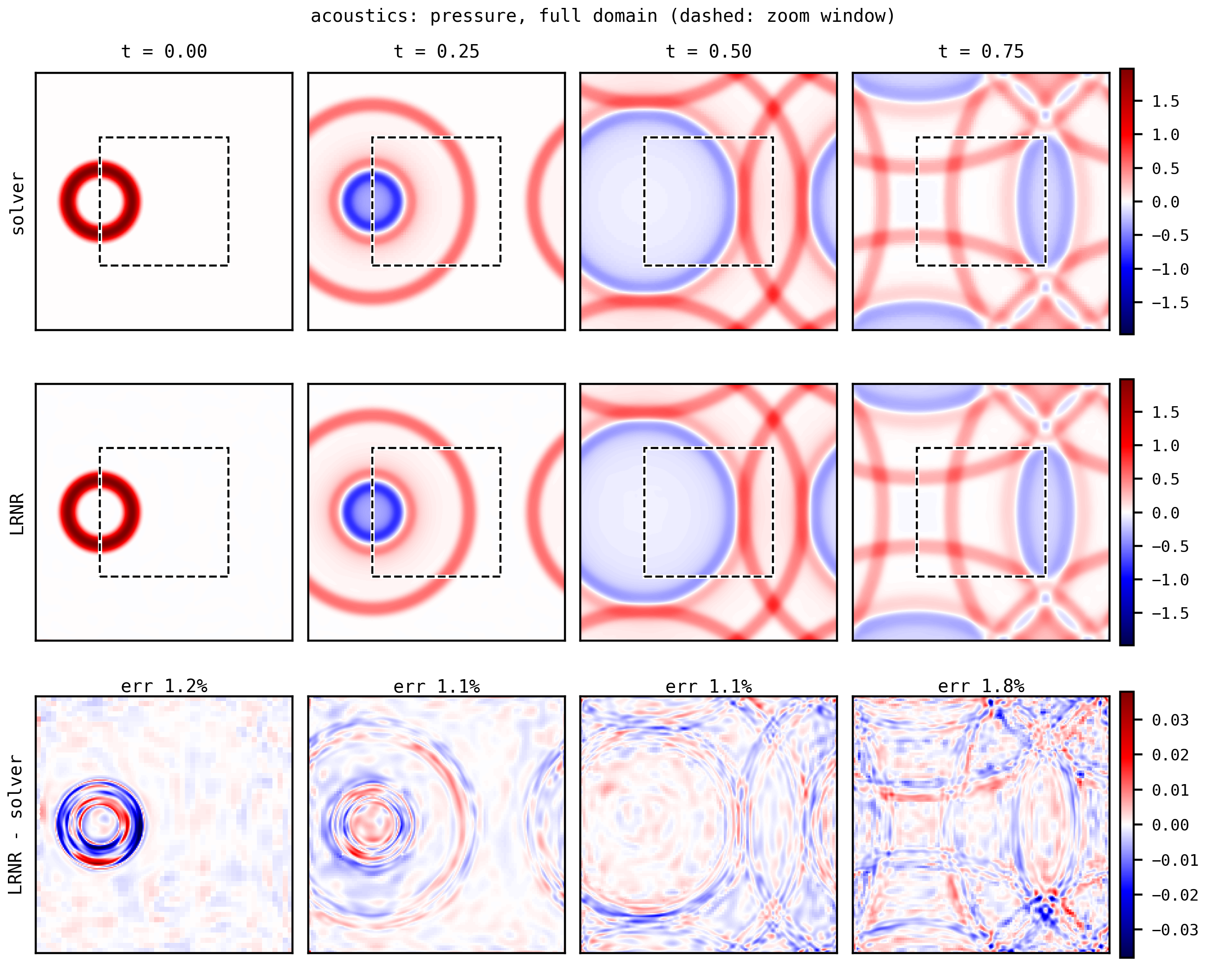}
\caption{Periodic acoustics, pressure: as Fig.~\ref{fig:story_euler}.}
\label{fig:story_acoustics}
\end{figure}

\begin{figure}[tp]
\centering
\figor{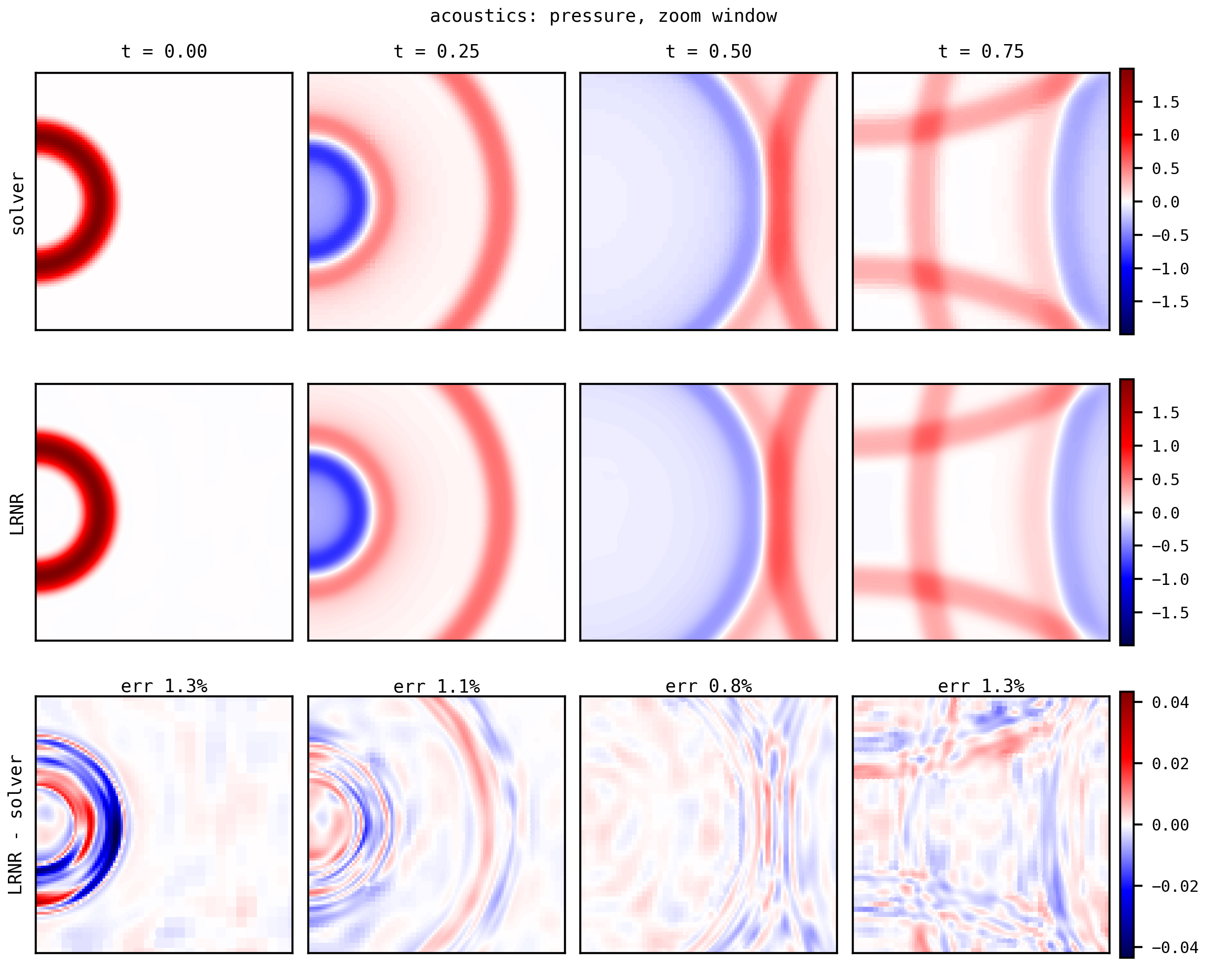}
\caption{As Fig.~\ref{fig:story_acoustics}, inside the central window $[-0.5,0.5]^2$ where the
periodic images of the ring cross.}
\label{fig:zoom_acoustics}
\end{figure}

\begin{figure}[tp]
\centering
\figor{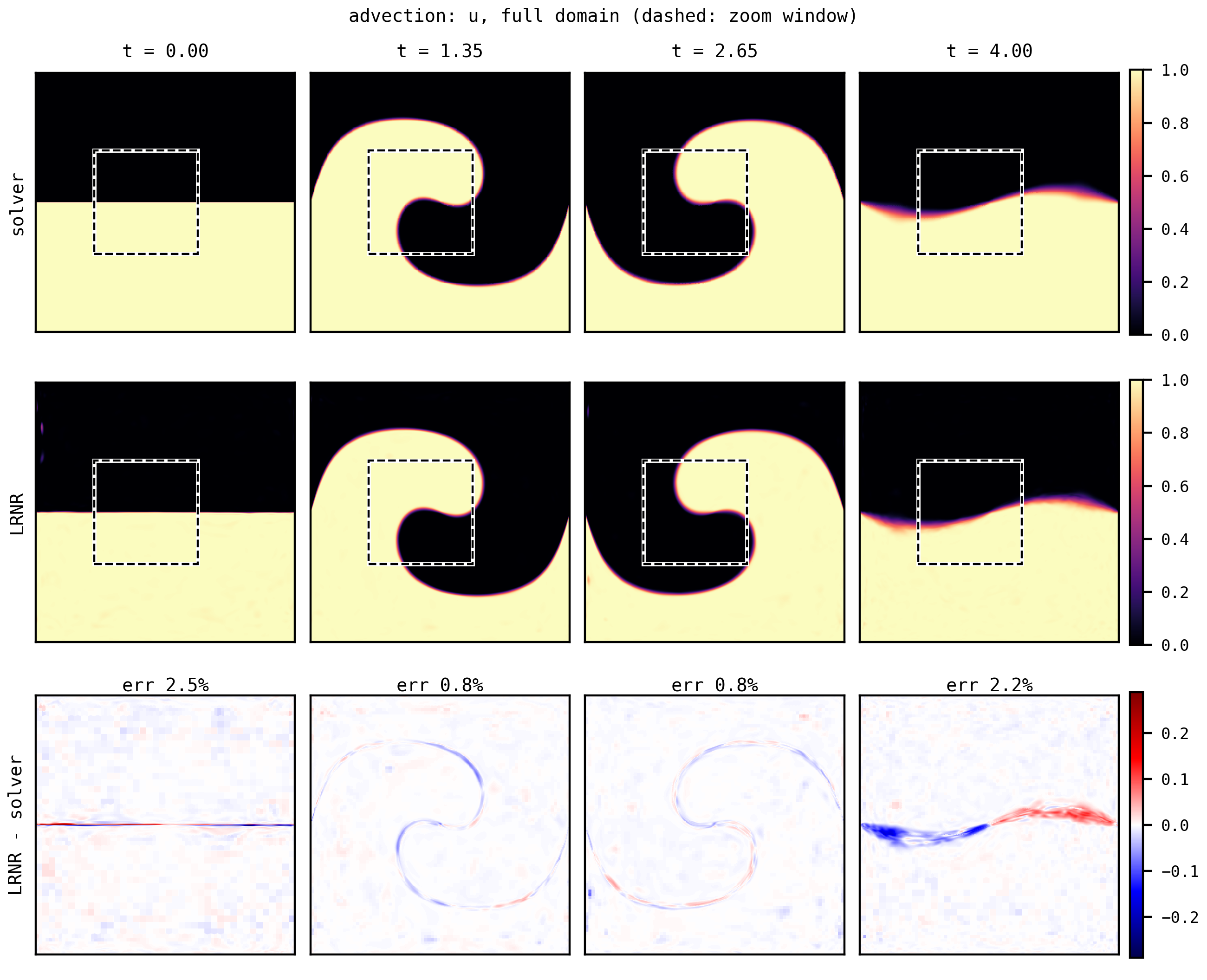}
\caption{Swirling advection: as Fig.~\ref{fig:story_euler}.}
\label{fig:story_advection}
\end{figure}

\begin{figure}[tp]
\centering
\figor{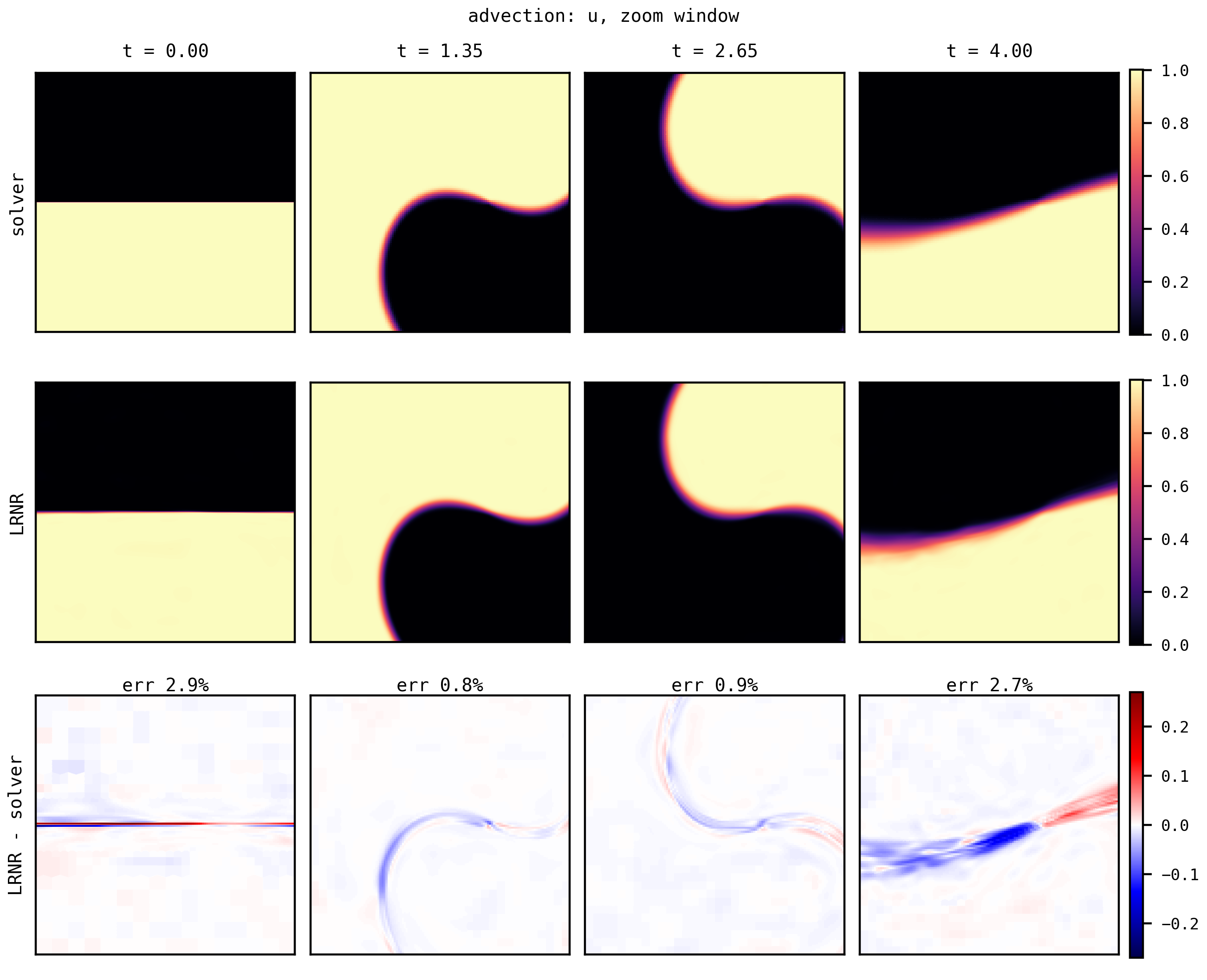}
\caption{As Fig.~\ref{fig:story_advection}, inside the $0.4$-wide
square holding the most solver points at the final time, the swirl center.}
\label{fig:zoom_advection}
\end{figure}

\begin{figure}[tp]
\centering
\figor{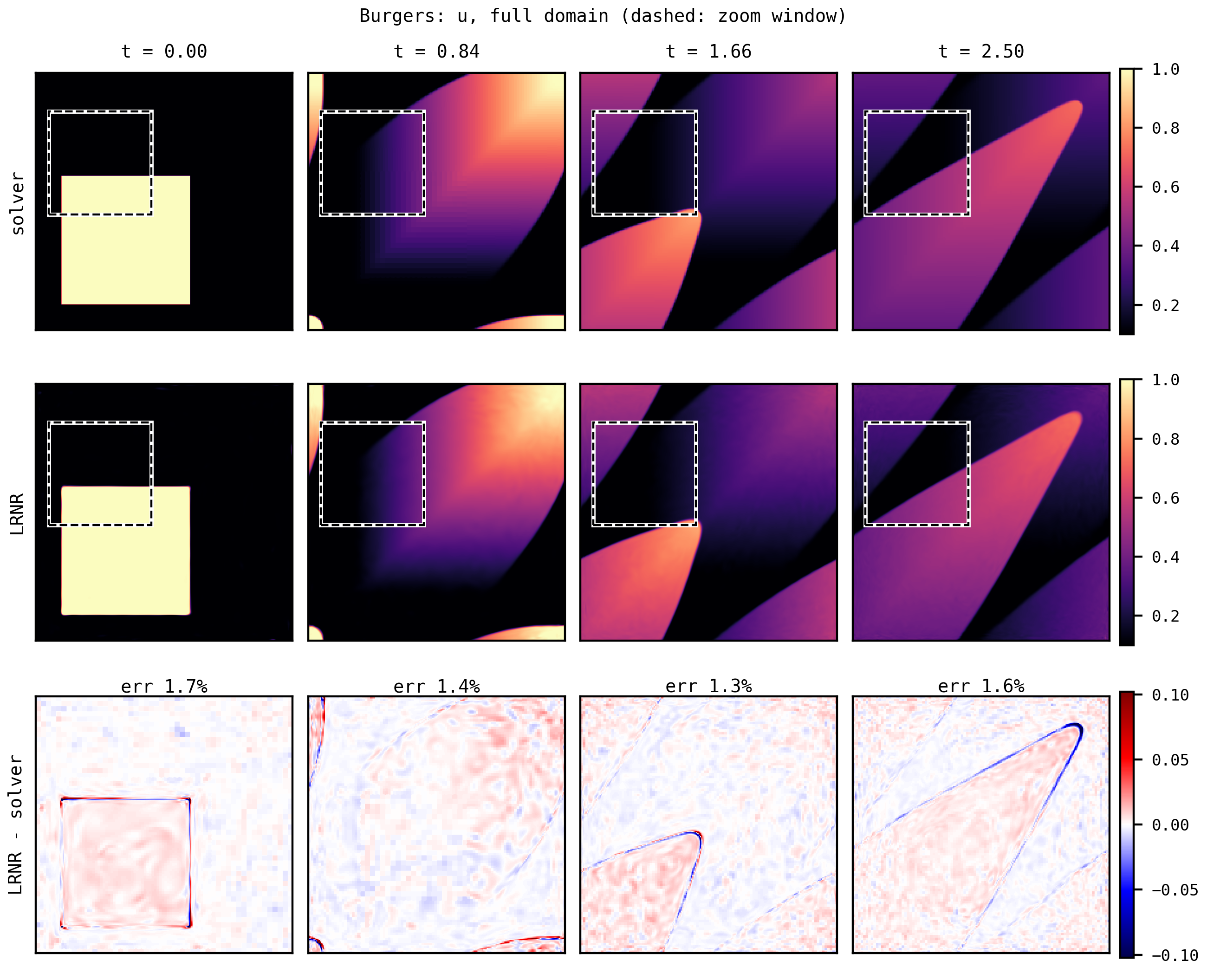}
\caption{Periodic Burgers: as Fig.~\ref{fig:story_euler}.}
\label{fig:story_burgers}
\end{figure}

\begin{figure}[tp]
\centering
\figor{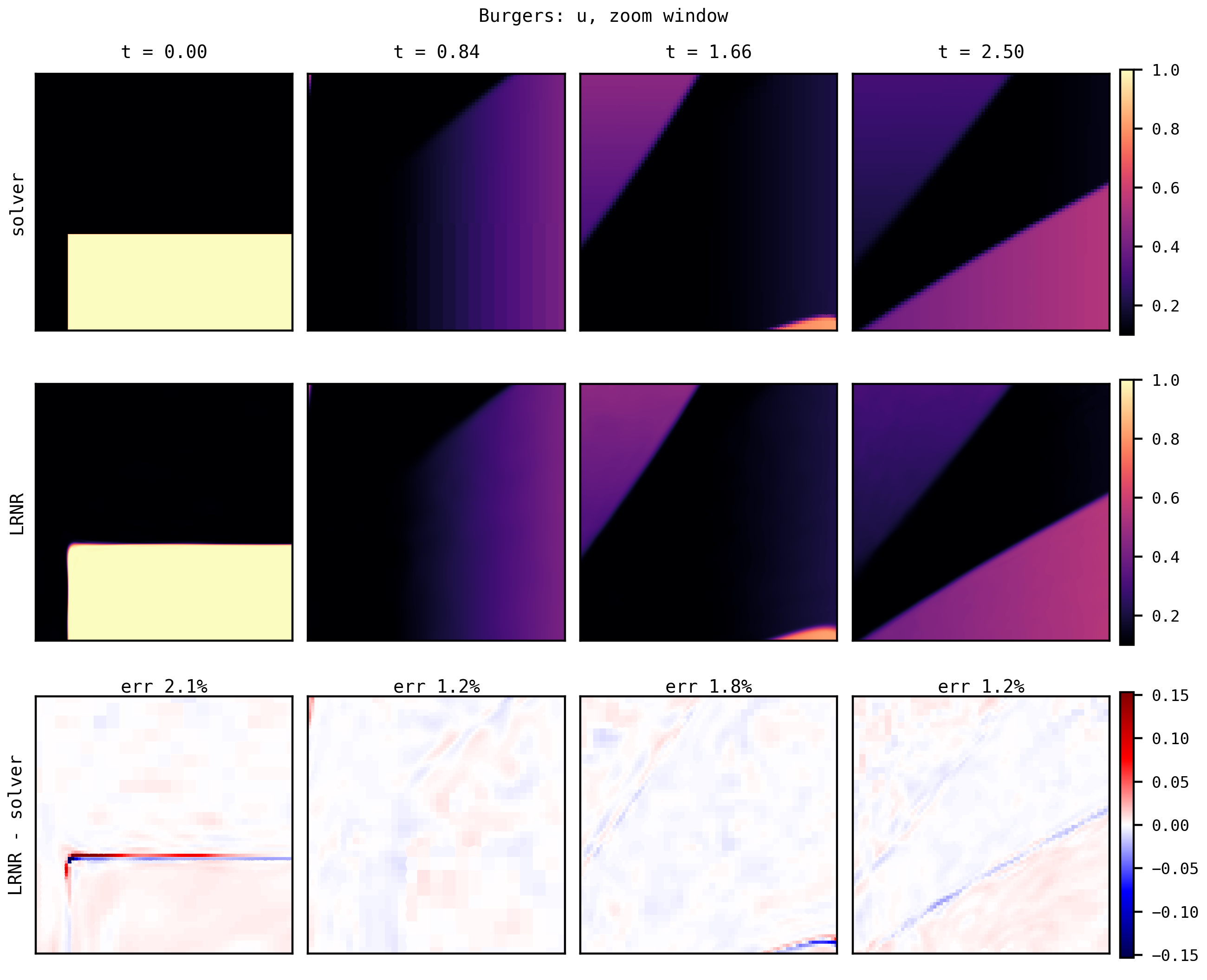}
\caption{As Fig.~\ref{fig:story_burgers}, inside the $0.4$-wide square
holding the most solver points at the final time, on the leading shock.}
\label{fig:zoom_burgers}
\end{figure}

\begin{figure}[tp]
\centering
\figor{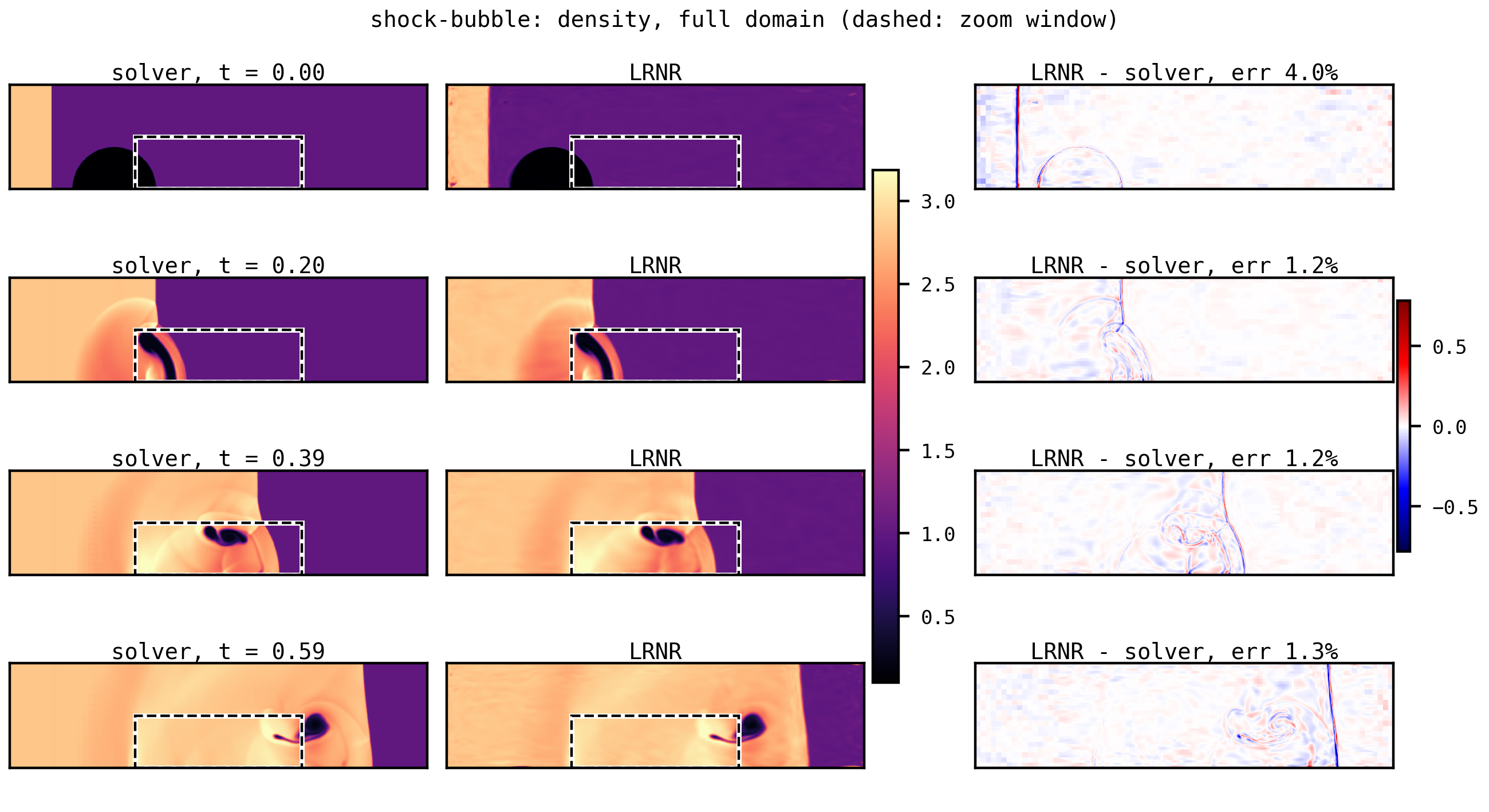}
\caption{Shock--bubble interaction, density, drawn at the physical aspect of the tube
$[0,2]\times[0,0.5]$ (the symmetry axis at the bottom): solver, \lrnr{} and their difference
(columns) at four times (rows) on the full domain; the dashed rectangle is the window of
Fig.~\ref{fig:zoom_shockbubble}.}
\label{fig:story_shockbubble}
\end{figure}

\begin{figure}[tp]
\centering
\figor{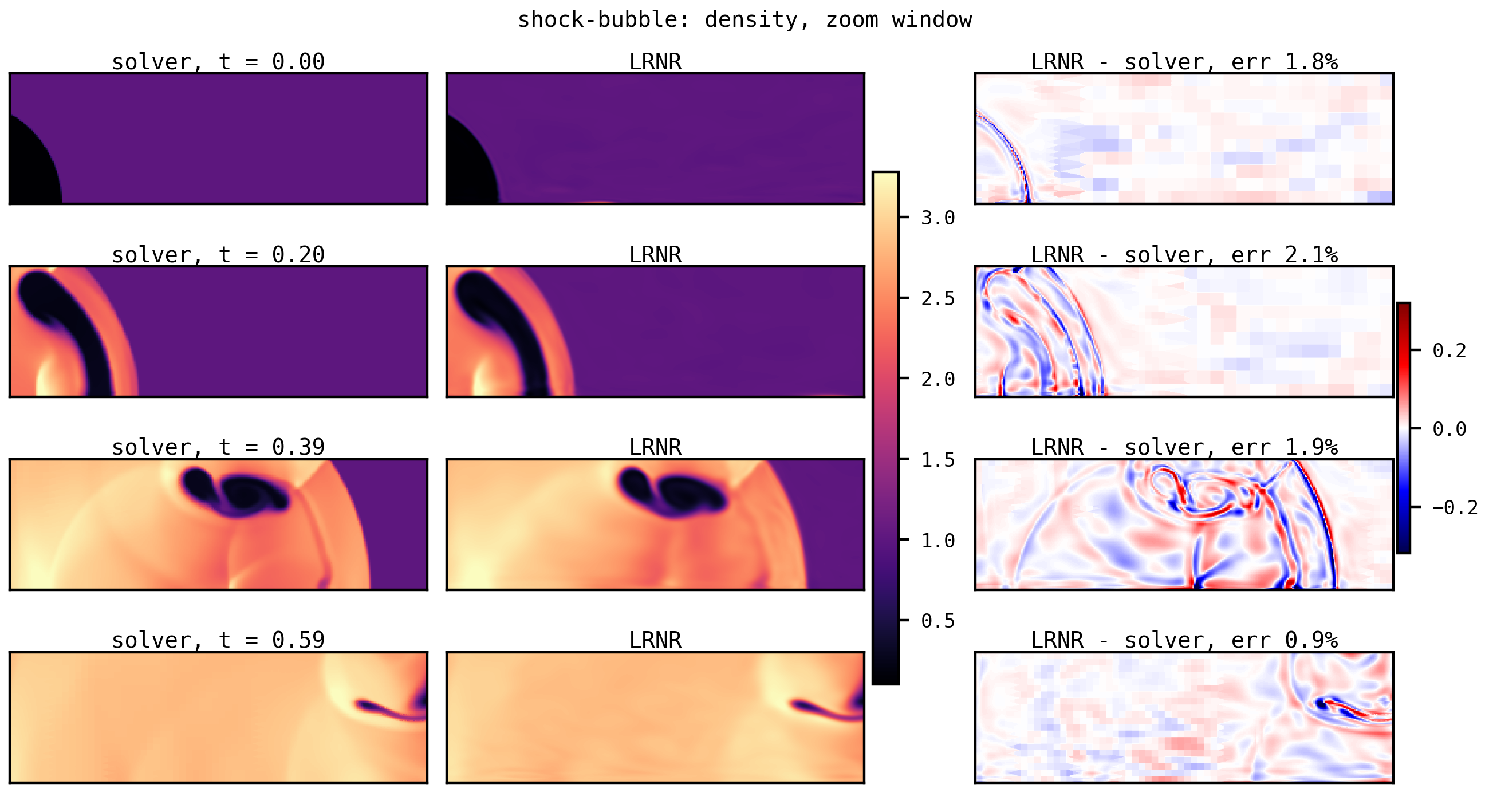}
\caption{As Fig.~\ref{fig:story_shockbubble}, inside the window $[0.6,1.4]\times[0,0.25]$ that
the vortex ring crosses.}
\label{fig:zoom_shockbubble}
\end{figure}

\begin{figure}[tp]
\centering\stackh{0.36}
\figor{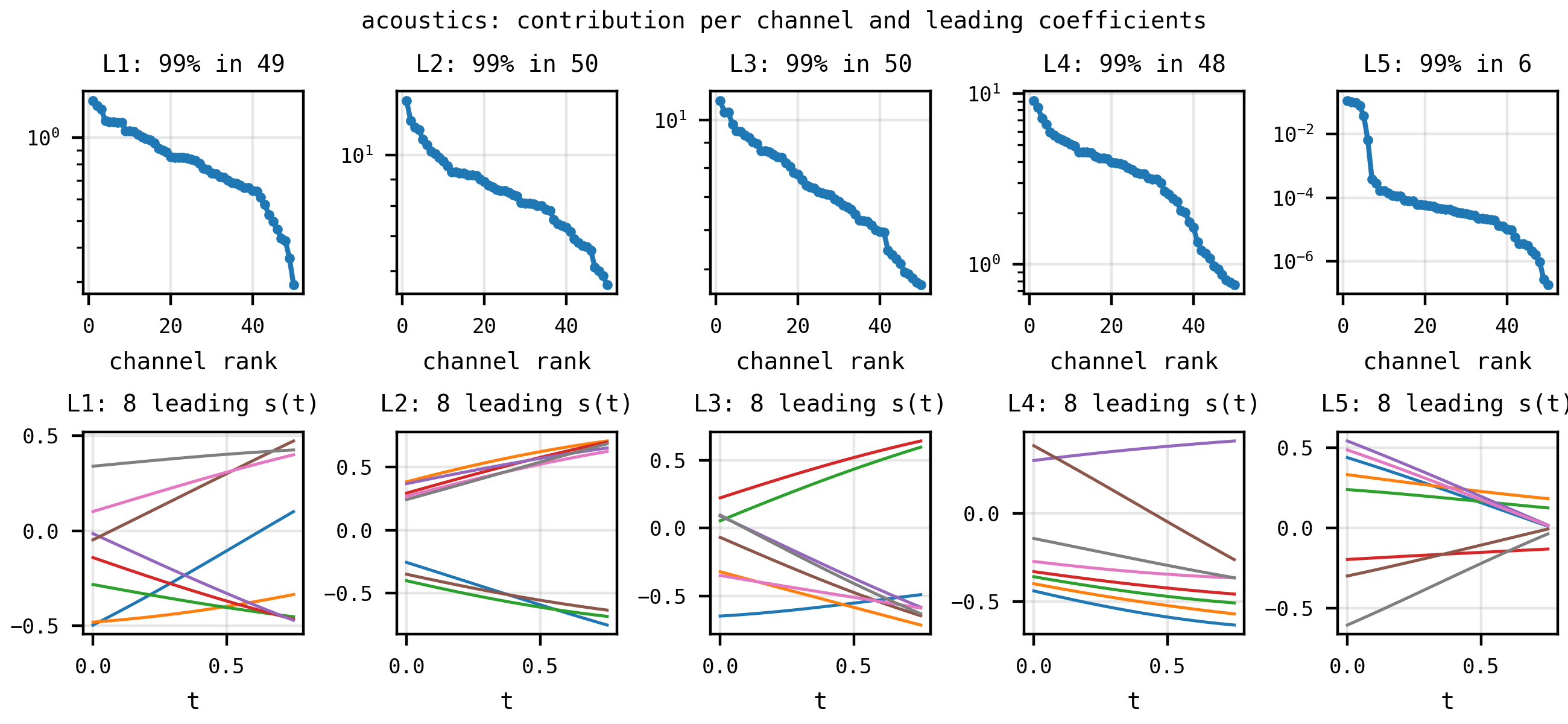}\\[4pt]\figor{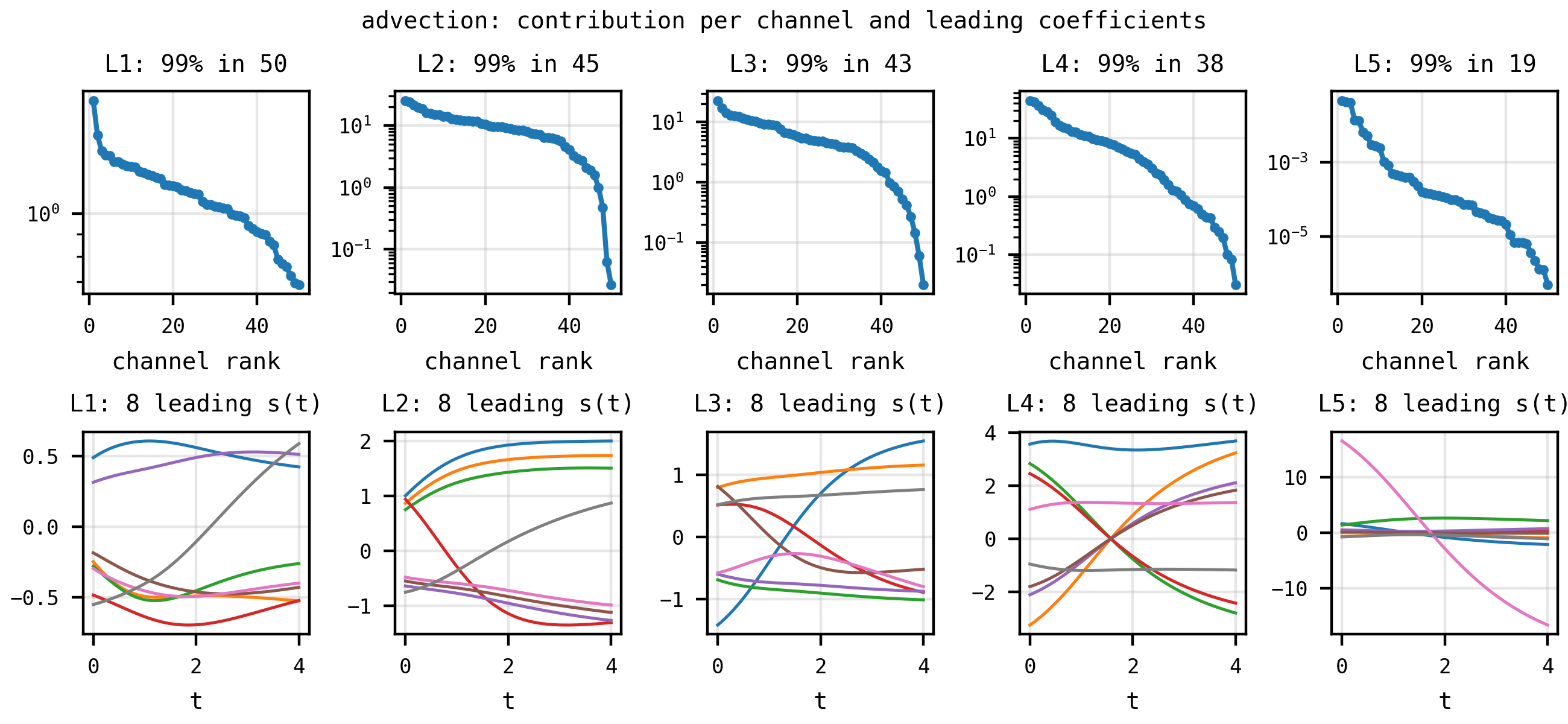}
\caption{As Fig.~\ref{fig:coef}, for the acoustics and advection models.}
\label{fig:coef_b}
\end{figure}

\begin{figure}[tp]
\centering\stackh{0.36}
\figor{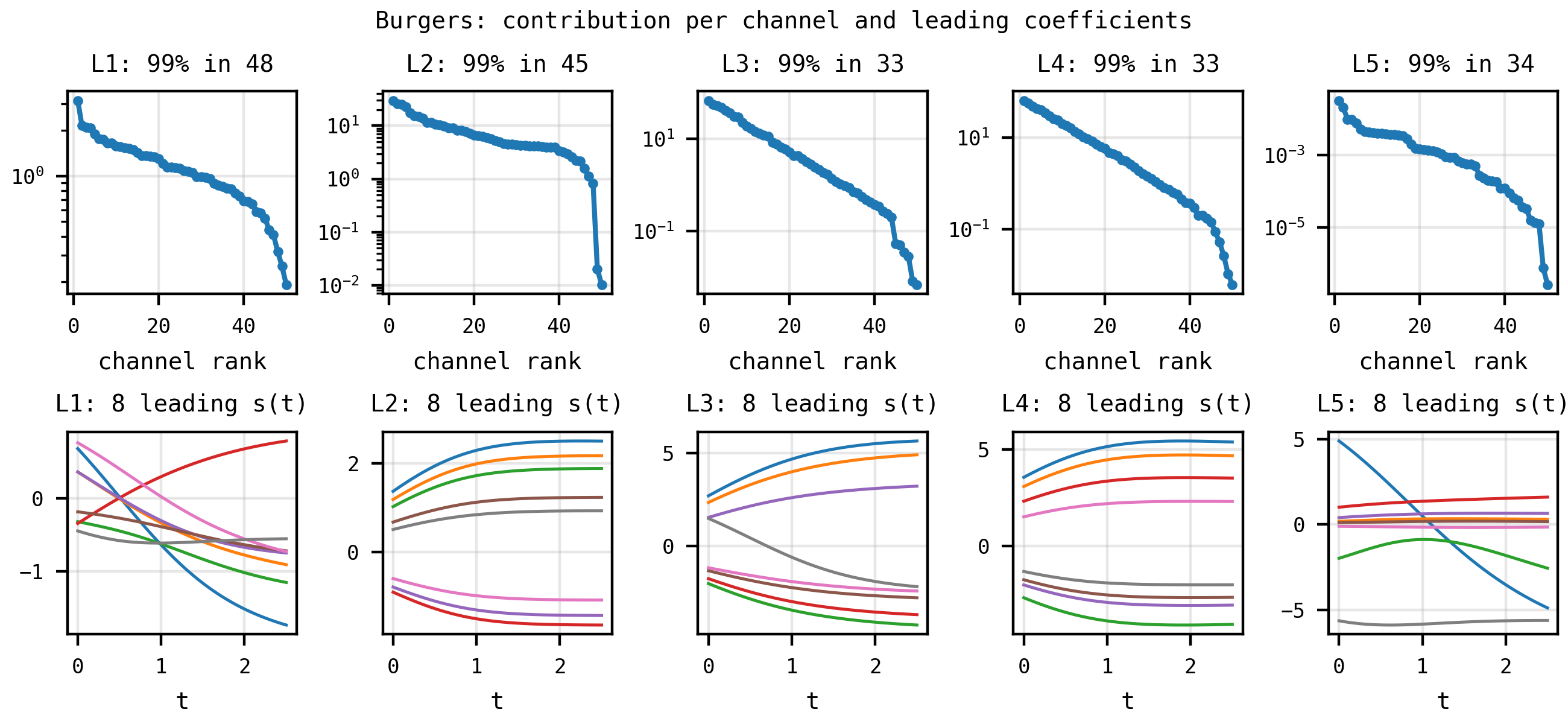}\\[4pt]\figor{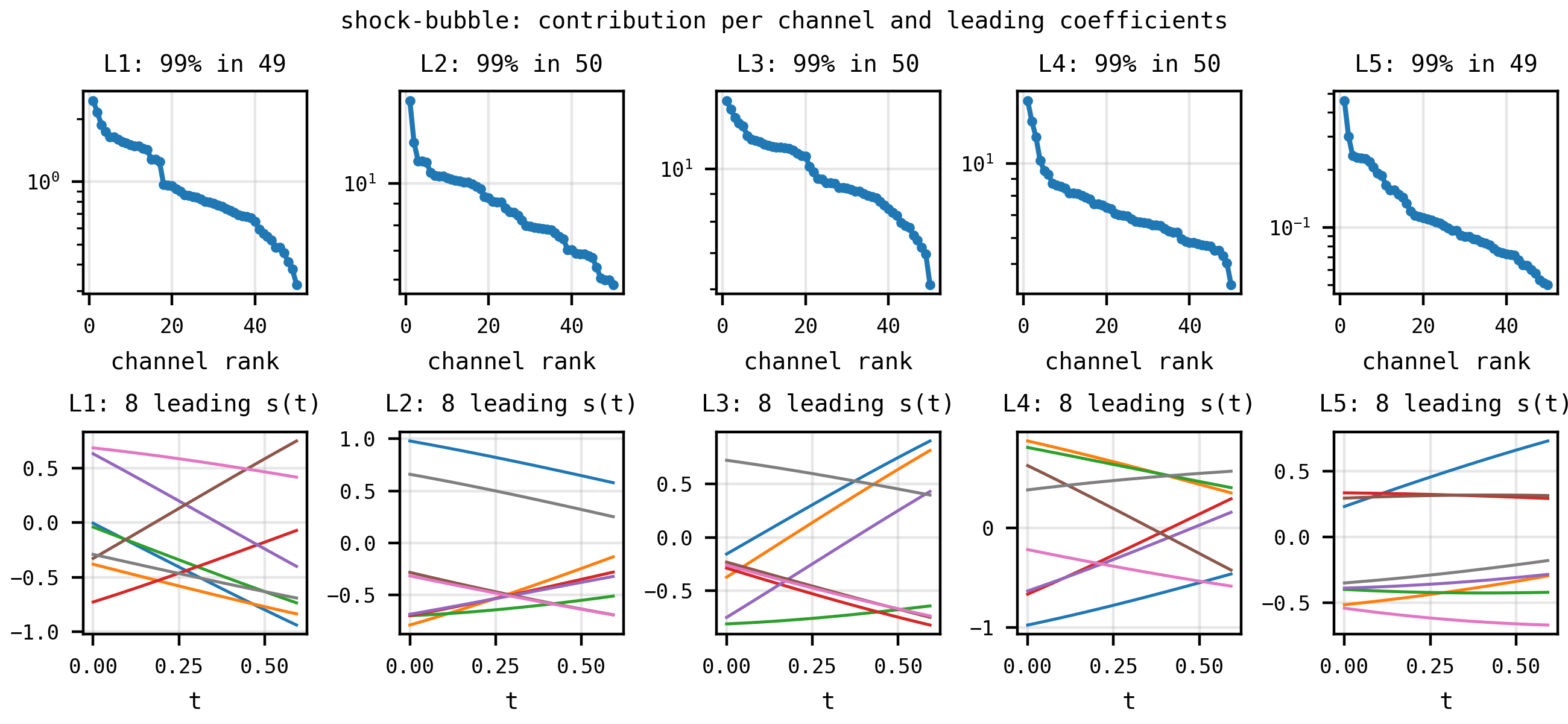}
\caption{As Fig.~\ref{fig:coef}, for the Burgers and shock--bubble models.}
\label{fig:coef_new}
\end{figure}

\begin{figure}[tp]
\centering\stackh{0.9}
\figor{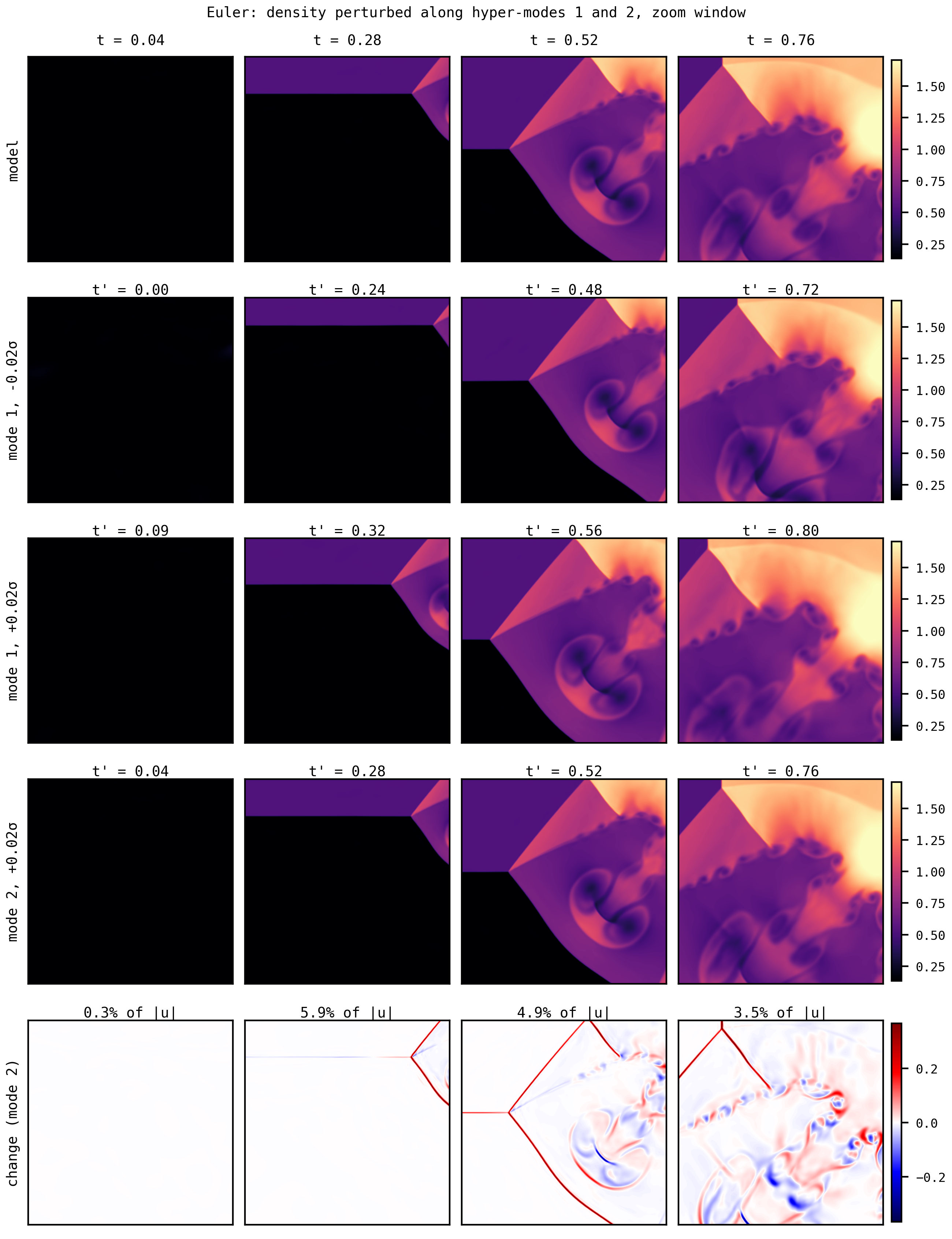}
\caption{As Fig.~\ref{fig:hm_perturb}, inside the Kelvin--Helmholtz window.}
\label{fig:hm_zoom_euler}
\end{figure}

\begin{figure}[tp]
\centering\stackh{0.9}
\figor{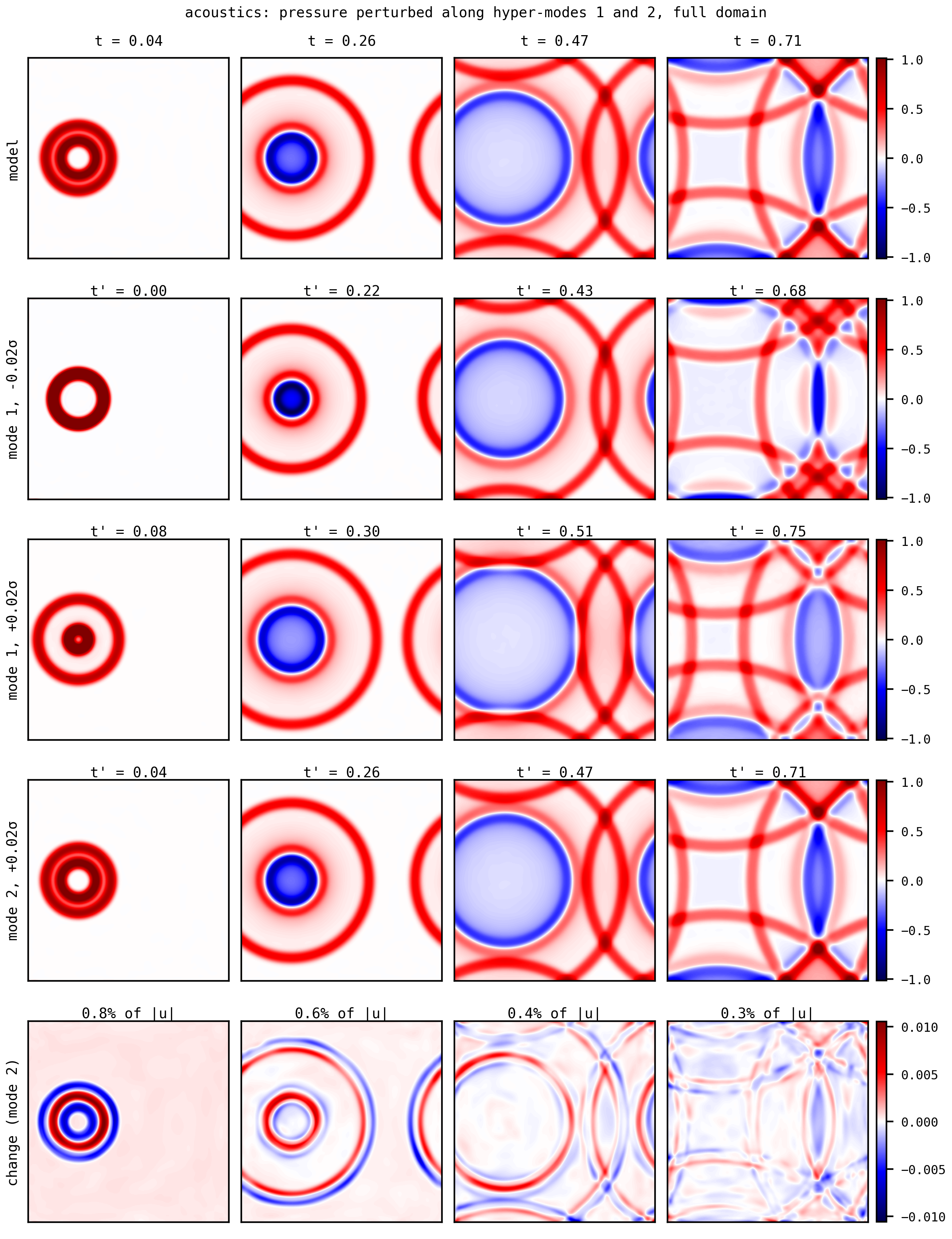}
\caption{Periodic acoustics, pressure, perturbed along the hyper-modes (layout of Fig.~\ref{fig:hm_perturb}).}
\label{fig:hm_acoustics}
\end{figure}

\begin{figure}[tp]
\centering\stackh{0.9}
\figor{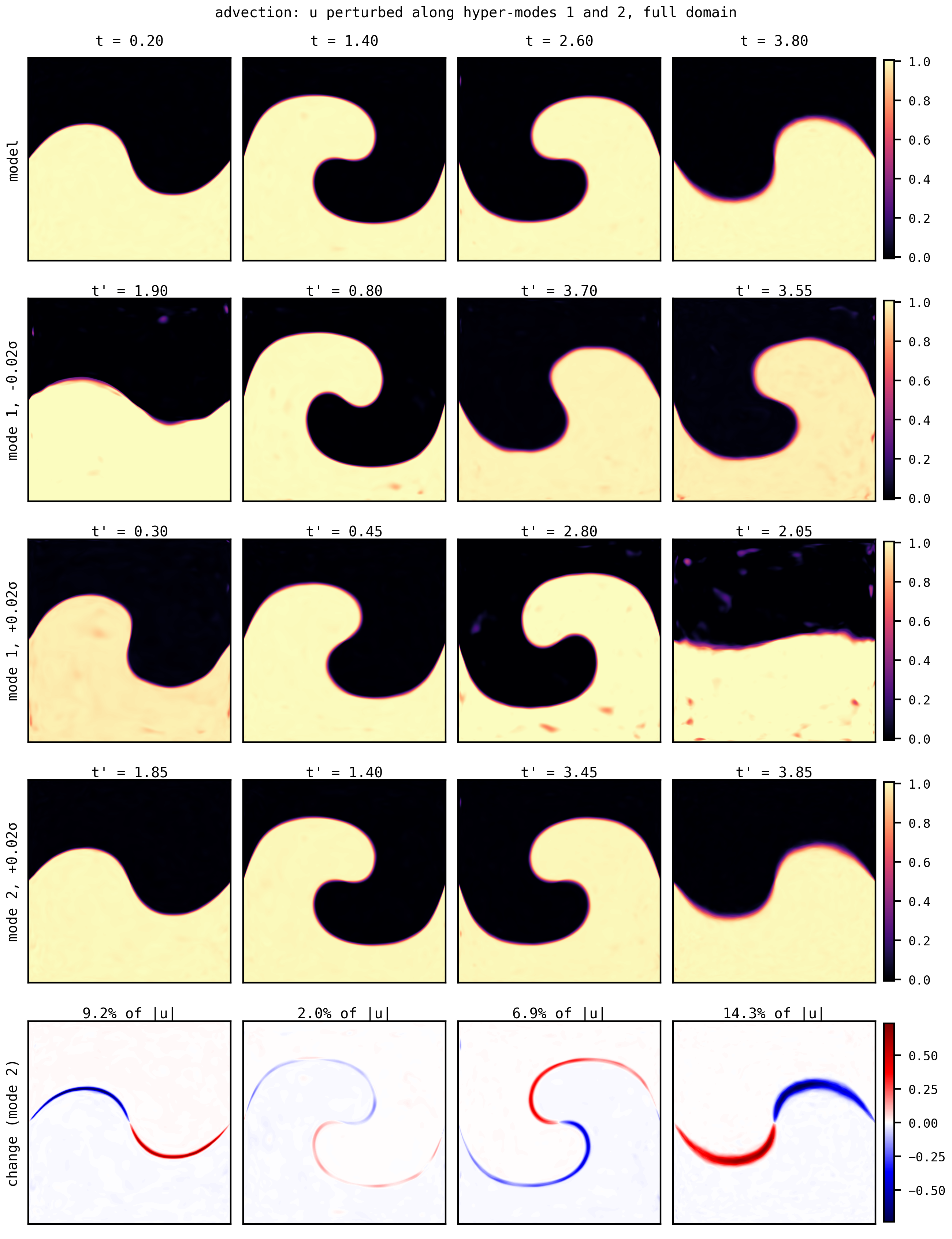}
\caption{Swirling advection, perturbed along the hyper-modes (layout of Fig.~\ref{fig:hm_perturb}).}
\label{fig:hm_advection}
\end{figure}

\begin{figure}[tp]
\centering\stackh{0.9}
\figor{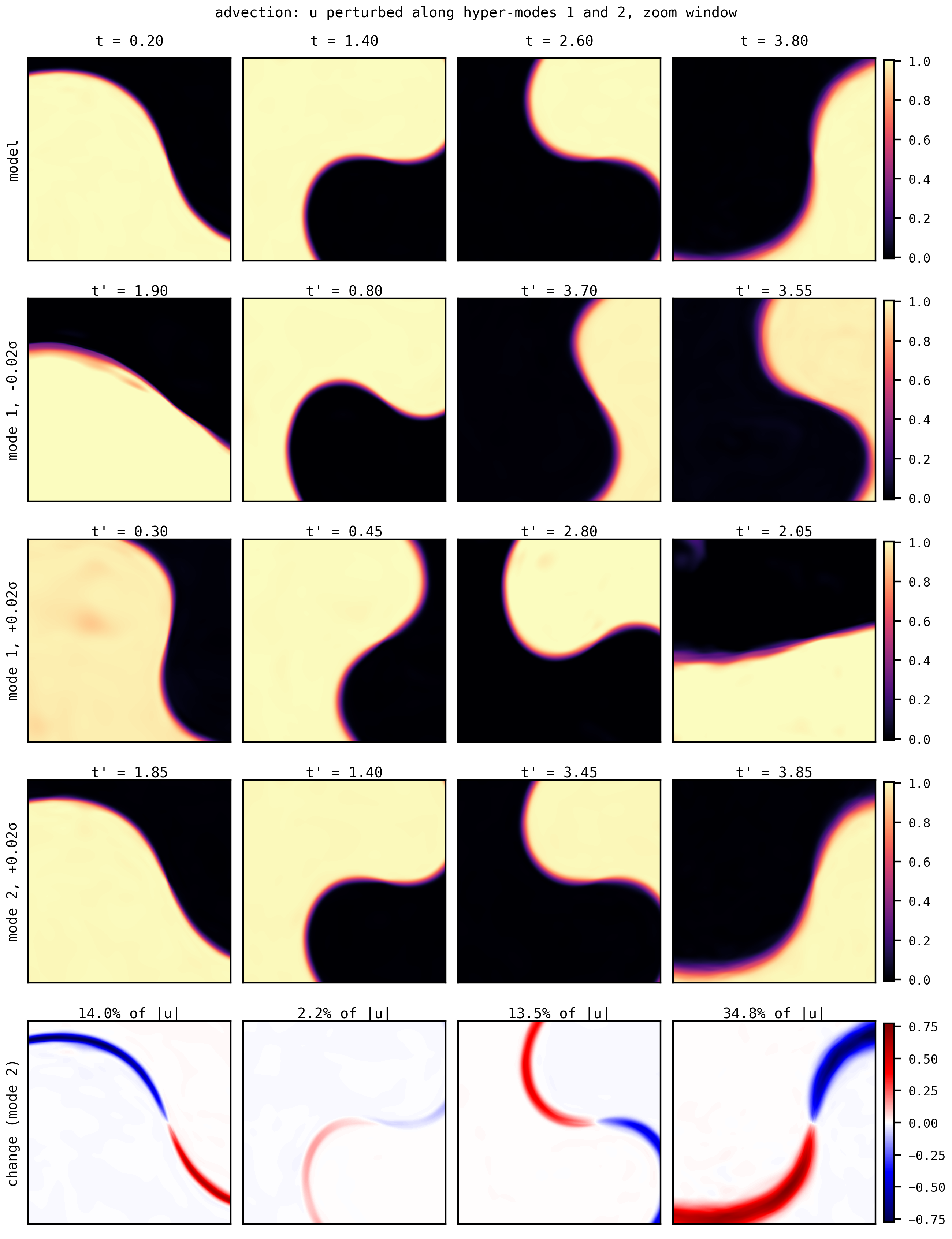}
\caption{As Fig.~\ref{fig:hm_advection}, inside the swirl-center window.}
\label{fig:hm_zoom_advection}
\end{figure}

\begin{figure}[tp]
\centering\stackh{0.9}
\figor{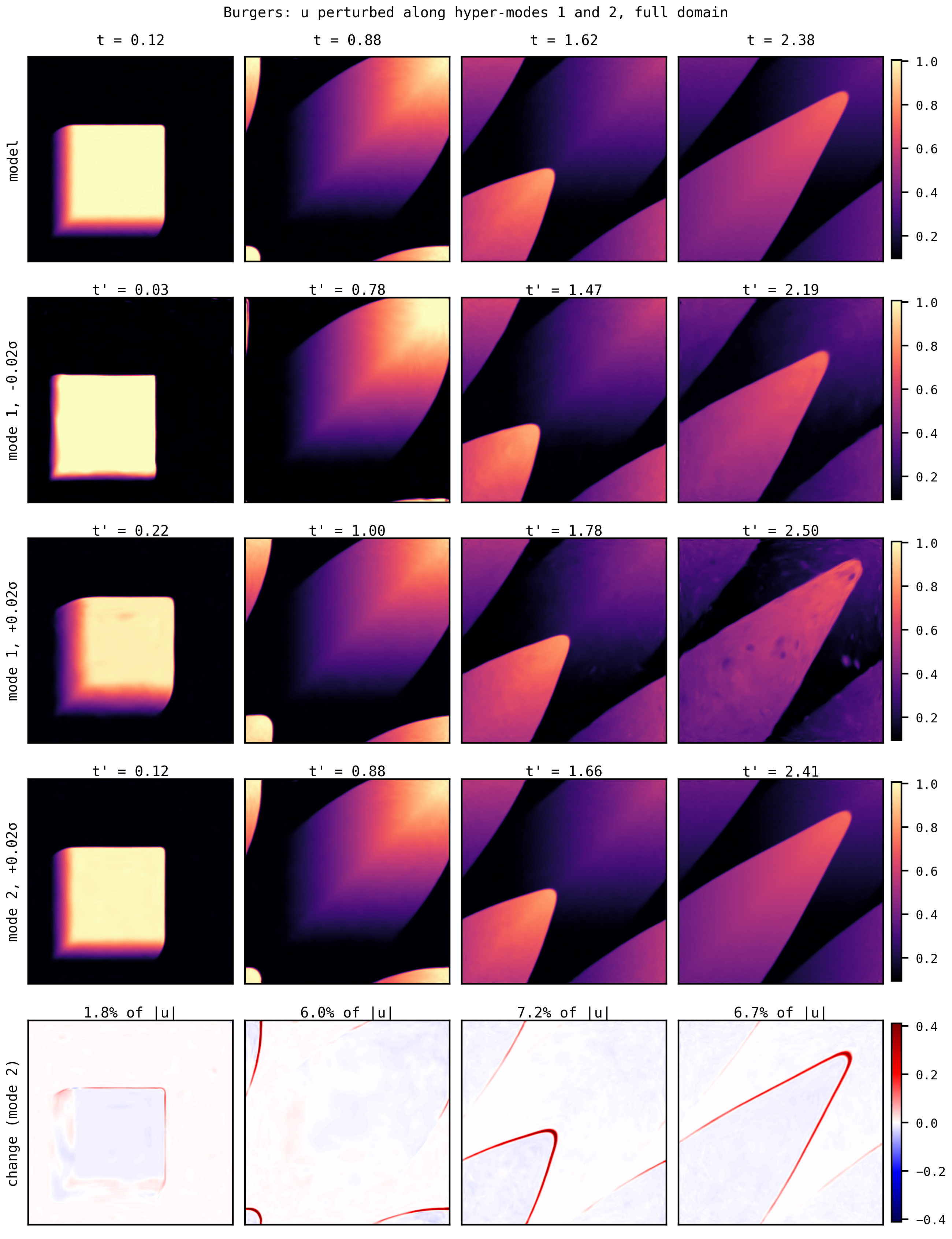}
\caption{Periodic Burgers, perturbed along the hyper-modes (layout of Fig.~\ref{fig:hm_perturb}).}
\label{fig:hm_burgers}
\end{figure}

\begin{figure}[tp]
\centering
\figor{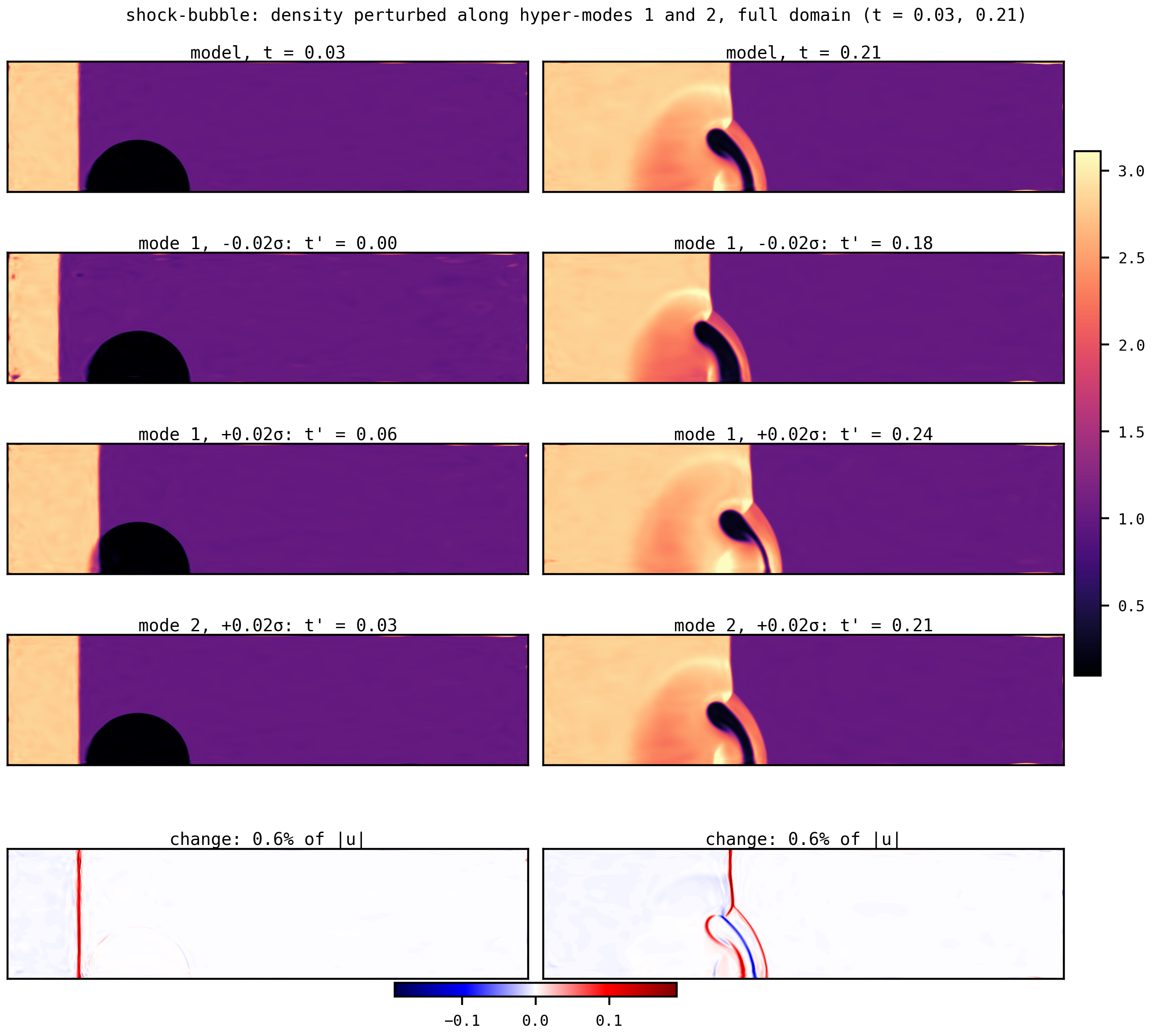}
\caption{Shock--bubble interaction, density, perturbed along the hyper-modes at the first two
times, drawn at the physical aspect of the tube. Each time is a column holding the five rows of
Fig.~\ref{fig:hm_perturb}: the model; $-0.02 \sigma_1\bv_1$ and $+0.02 \sigma_1\bv_1$ with the
nearest trajectory time $t'$; $+0.02 \sigma_2\bv_2$; and the change it makes.}
\label{fig:hm_shockbubble}
\end{figure}

\begin{figure}[tp]
\centering
\figor{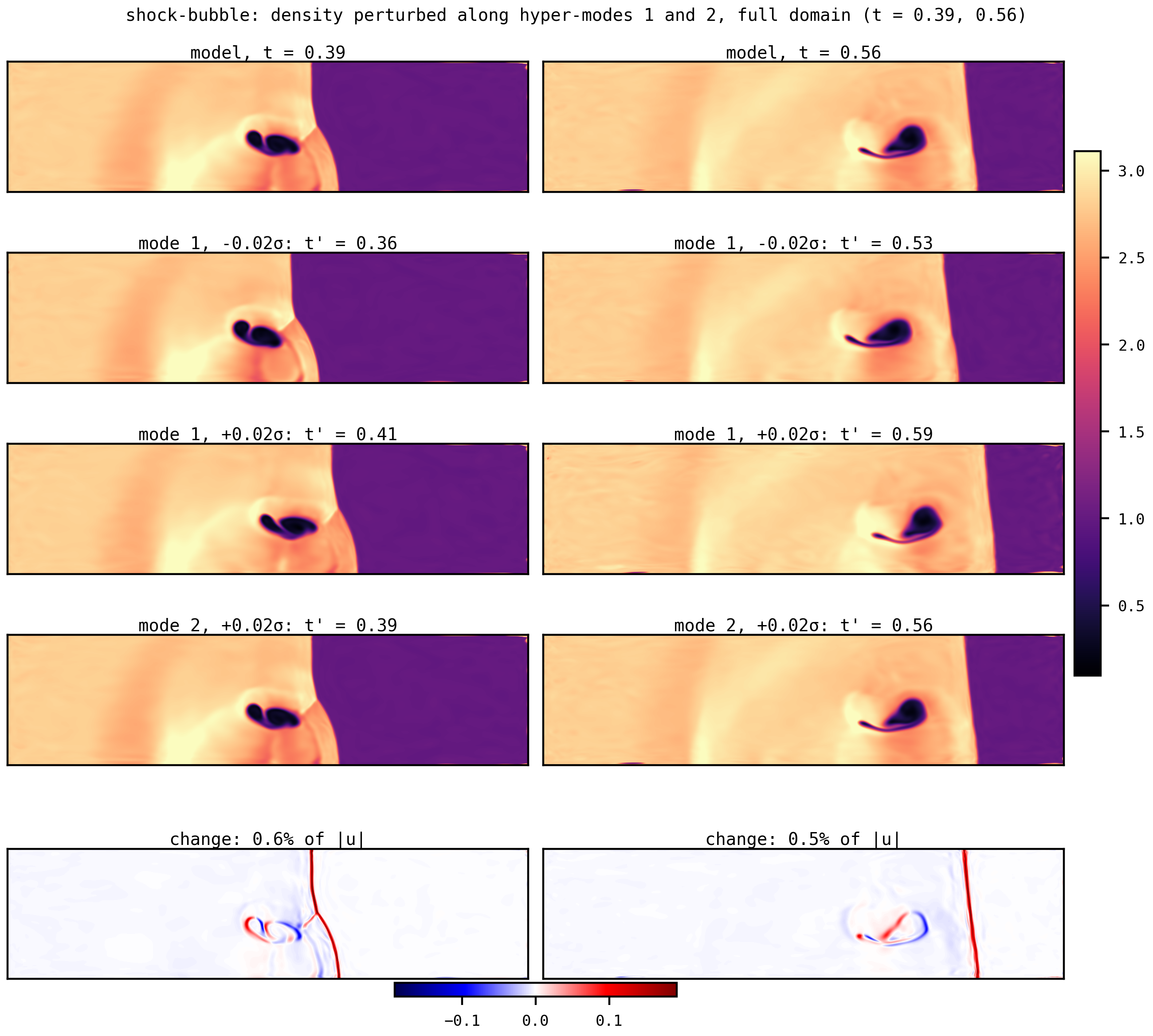}
\caption{As Fig.~\ref{fig:hm_shockbubble}, at the last two times.}
\label{fig:hm_shockbubble_b}
\end{figure}

\begin{figure}[tp]
\centering\stackh{0.9}
\figor{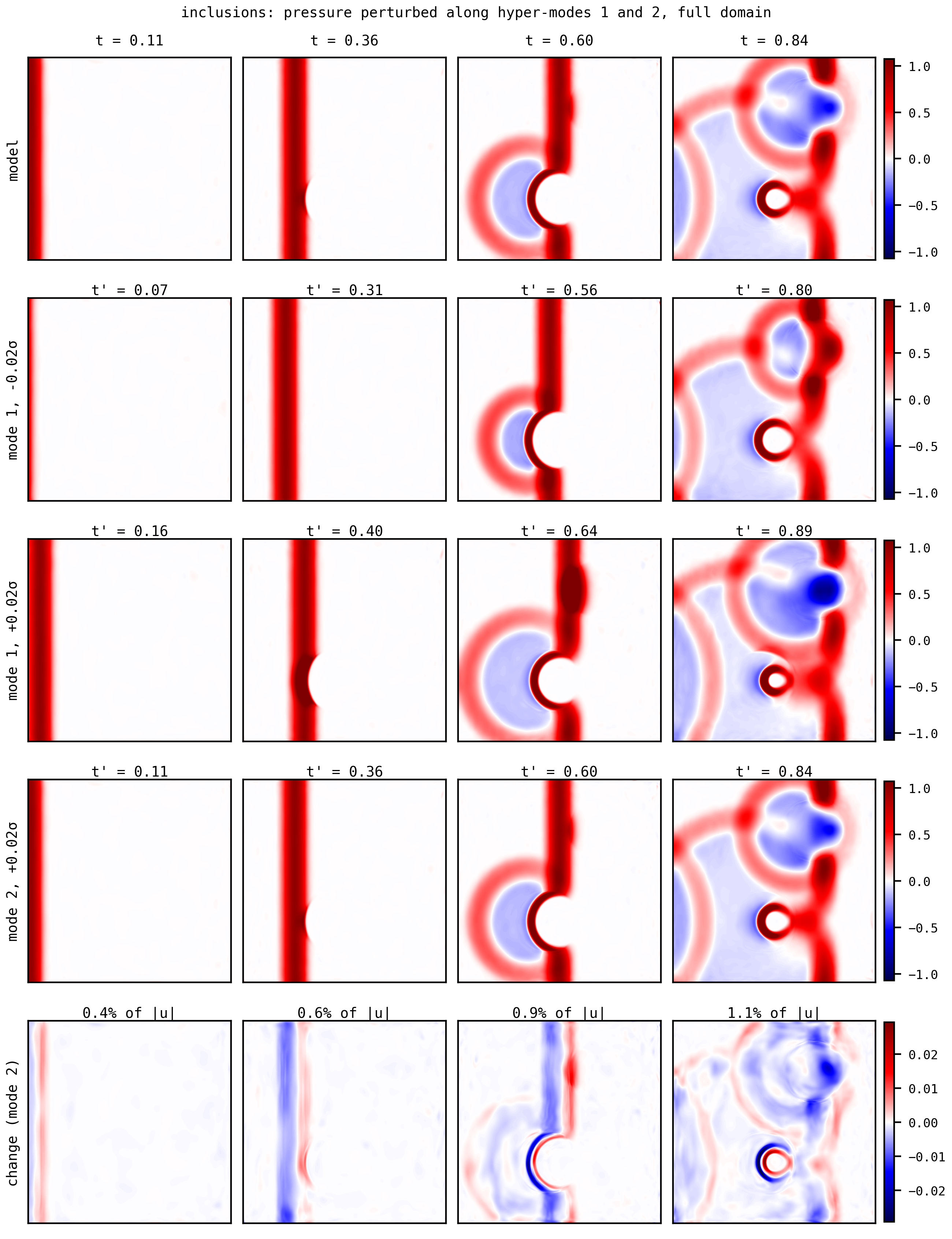}
\caption{Acoustics with inclusions, pressure, perturbed along the hyper-modes (layout of Fig.~\ref{fig:hm_perturb}).}
\label{fig:hm_inclusions}
\end{figure}

\begin{figure}[tp]
\centering\stackh{0.44}
\figor{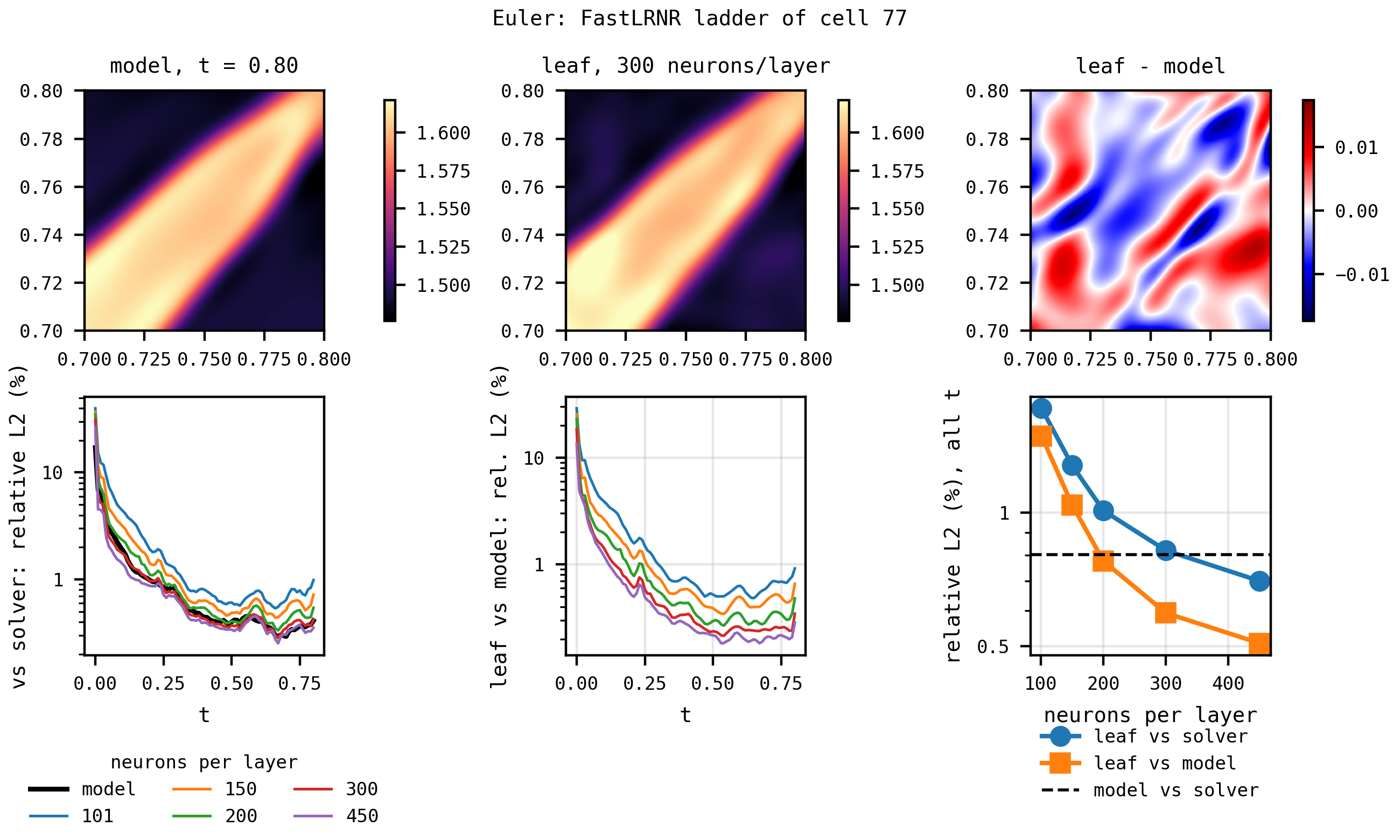}\\[4pt]\figor{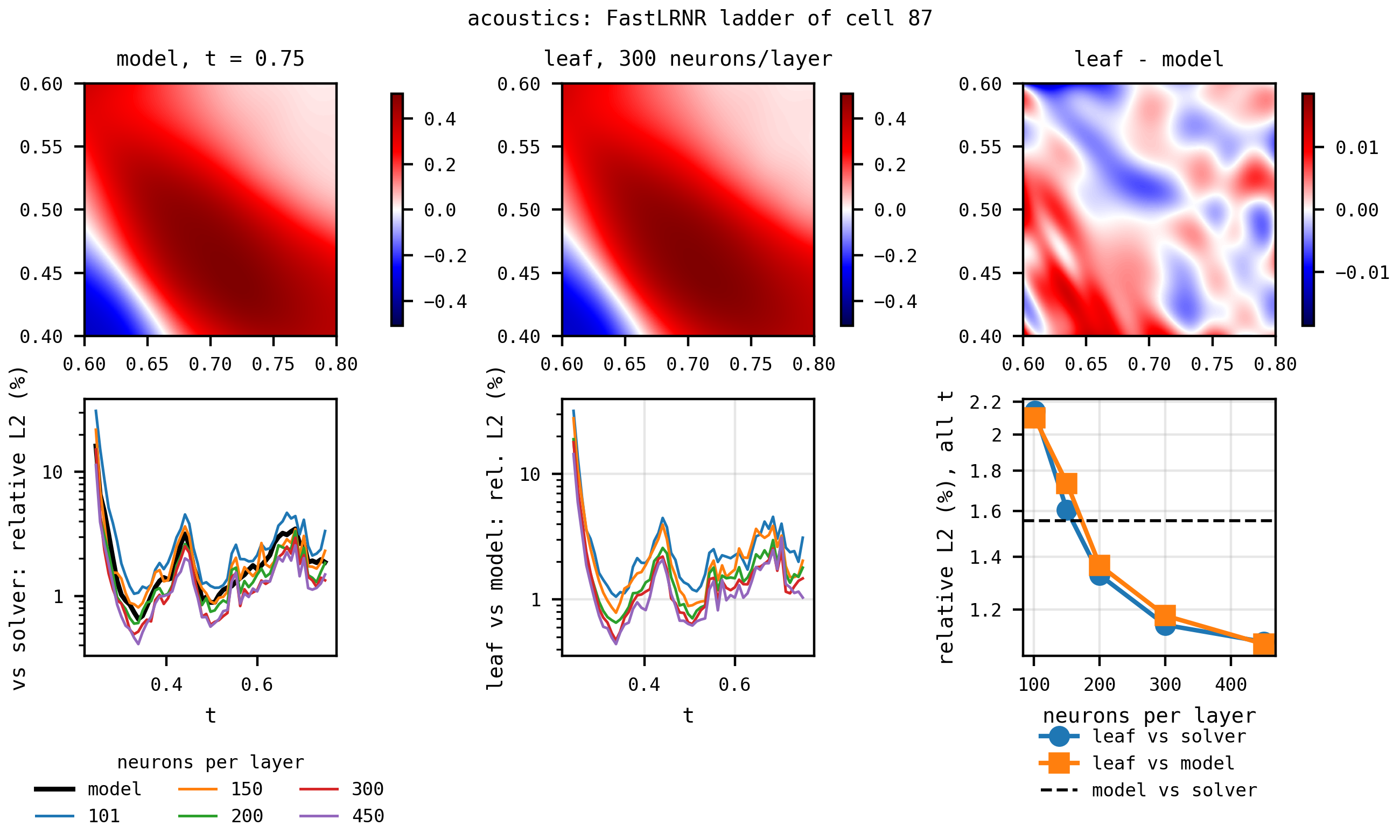}
\caption{The most complicated cell of the Euler and acoustics problems: the full model, the
300-neuron FastLRNR leaf and their difference at the final time (top row of each), and the error
against the solver and the agreement with the model over time for every width of the ladder,
with the ladder pooled over time (bottom row of each).}
\label{fig:fast_cells_a}
\end{figure}

\begin{figure}[tp]
\centering\stackh{0.44}
\figor{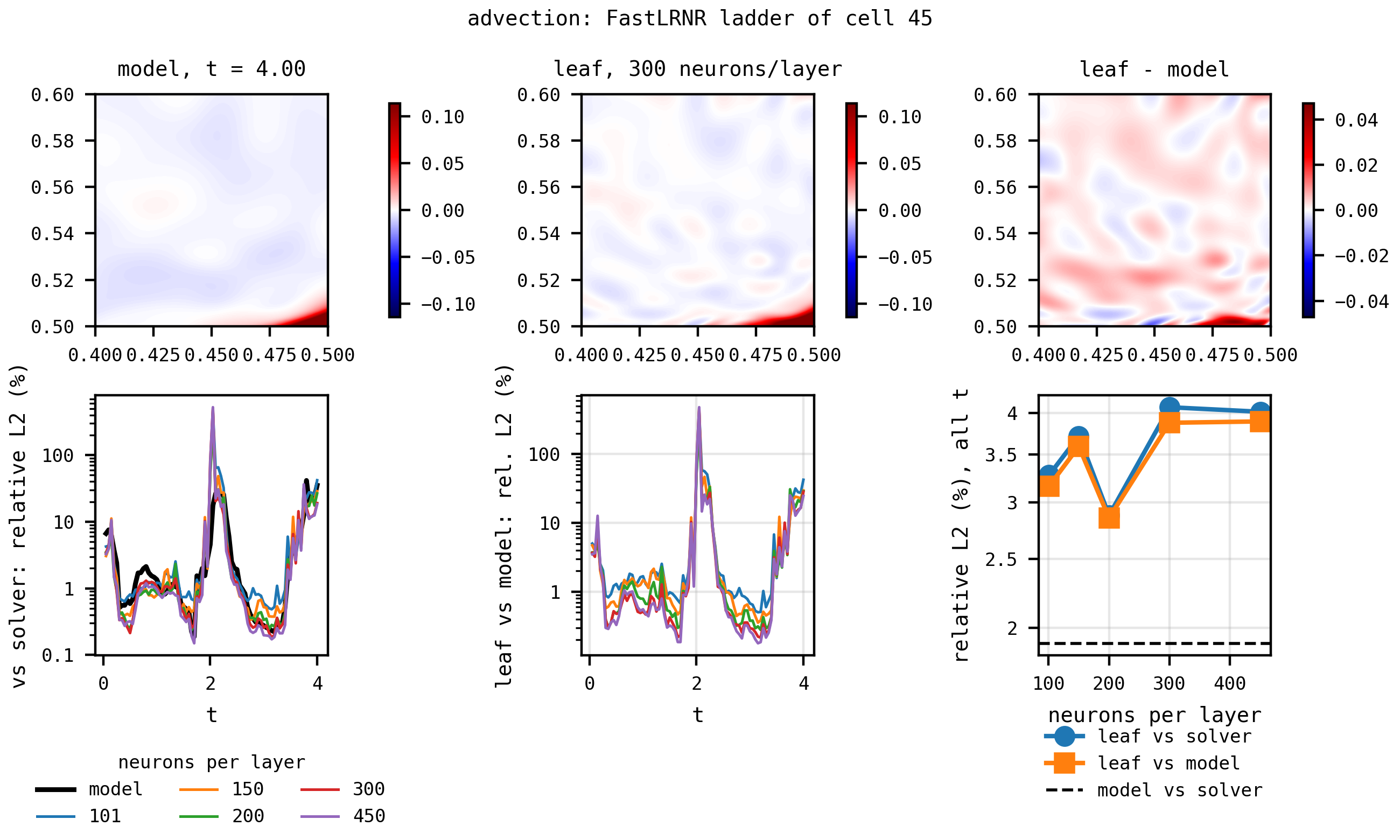}\\[4pt]\figor{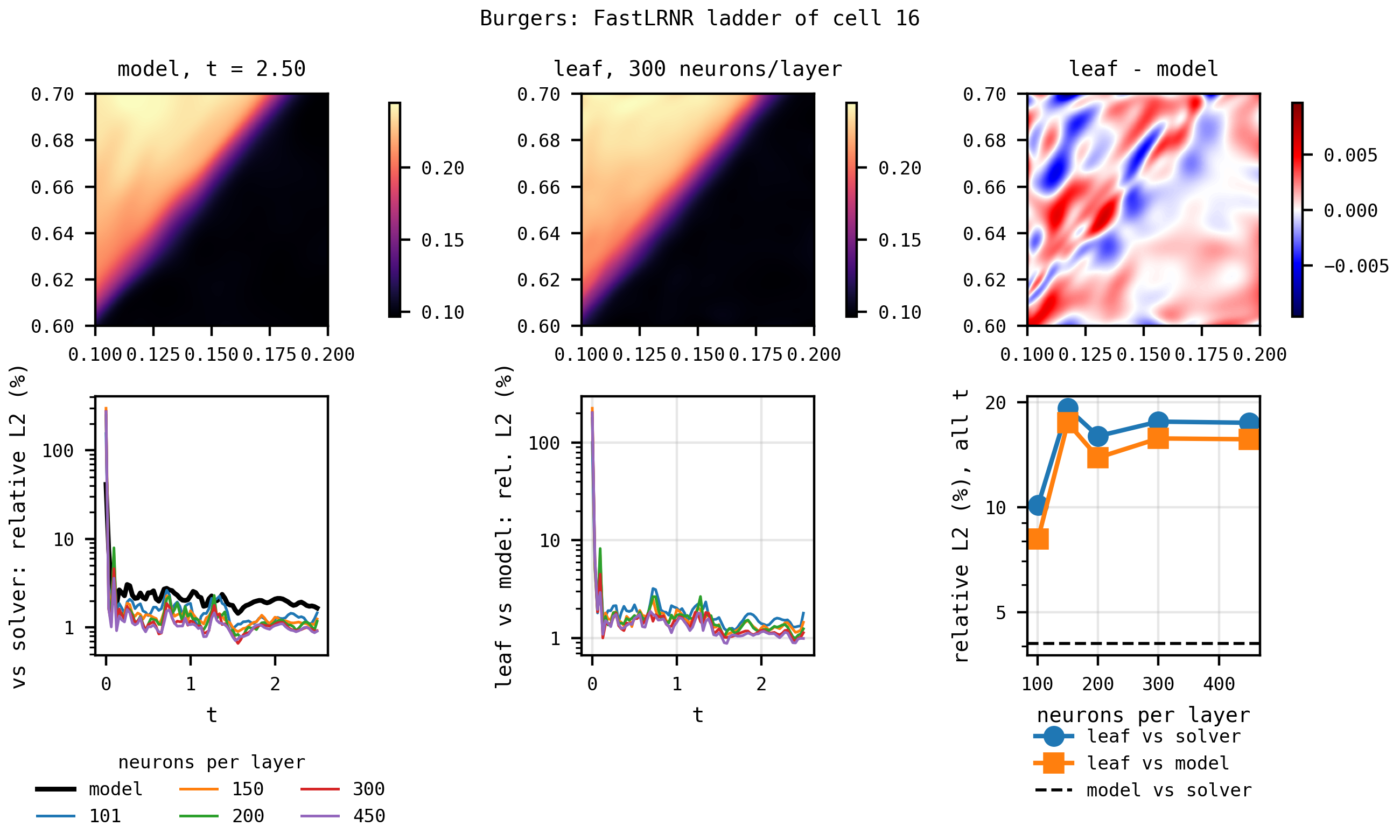}
\caption{As Fig.~\ref{fig:fast_cells_a}, for the advection and Burgers cells.}
\label{fig:fast_cells_b}
\end{figure}

\begin{figure}[tp]
\centering\stackh{0.44}
\figor{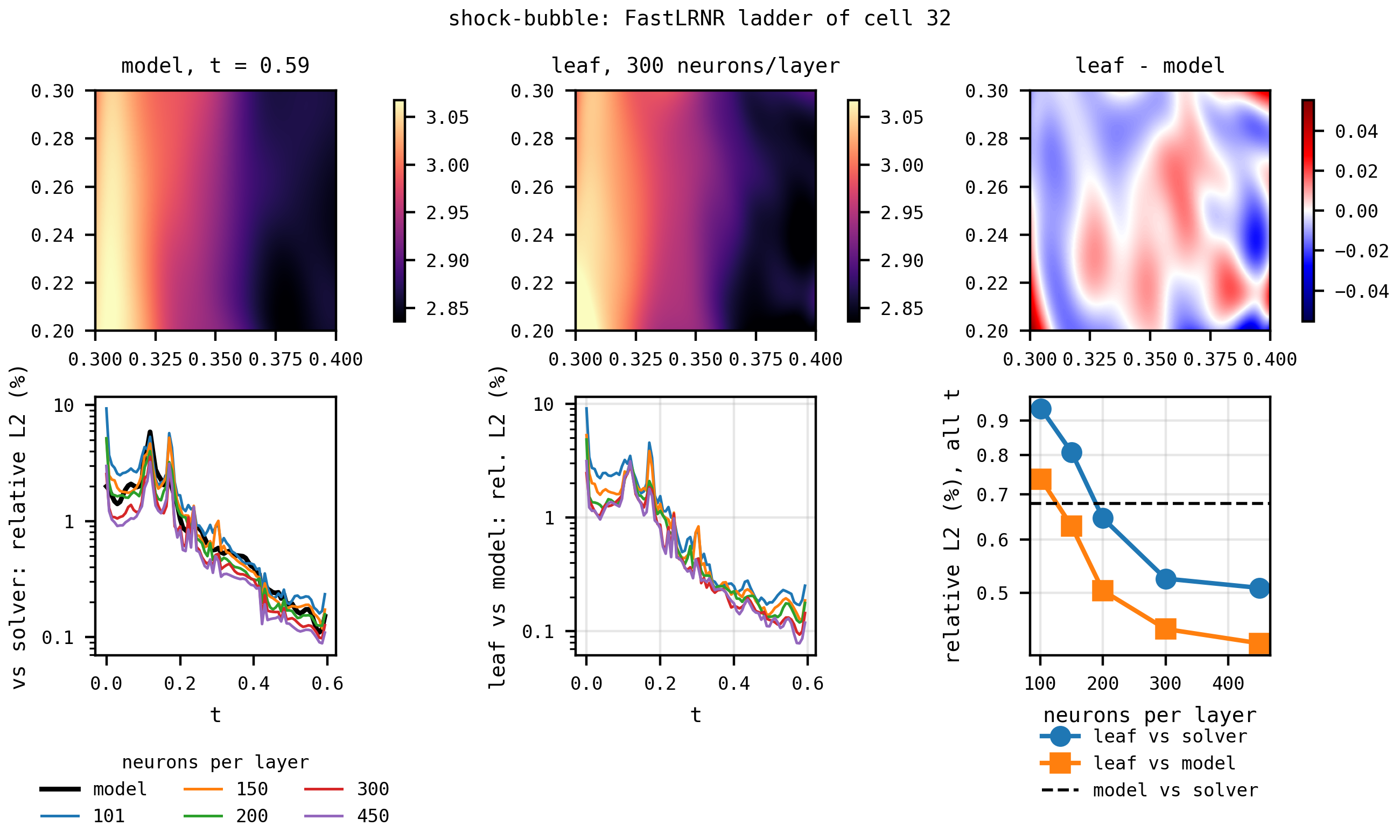}\\[4pt]\figor{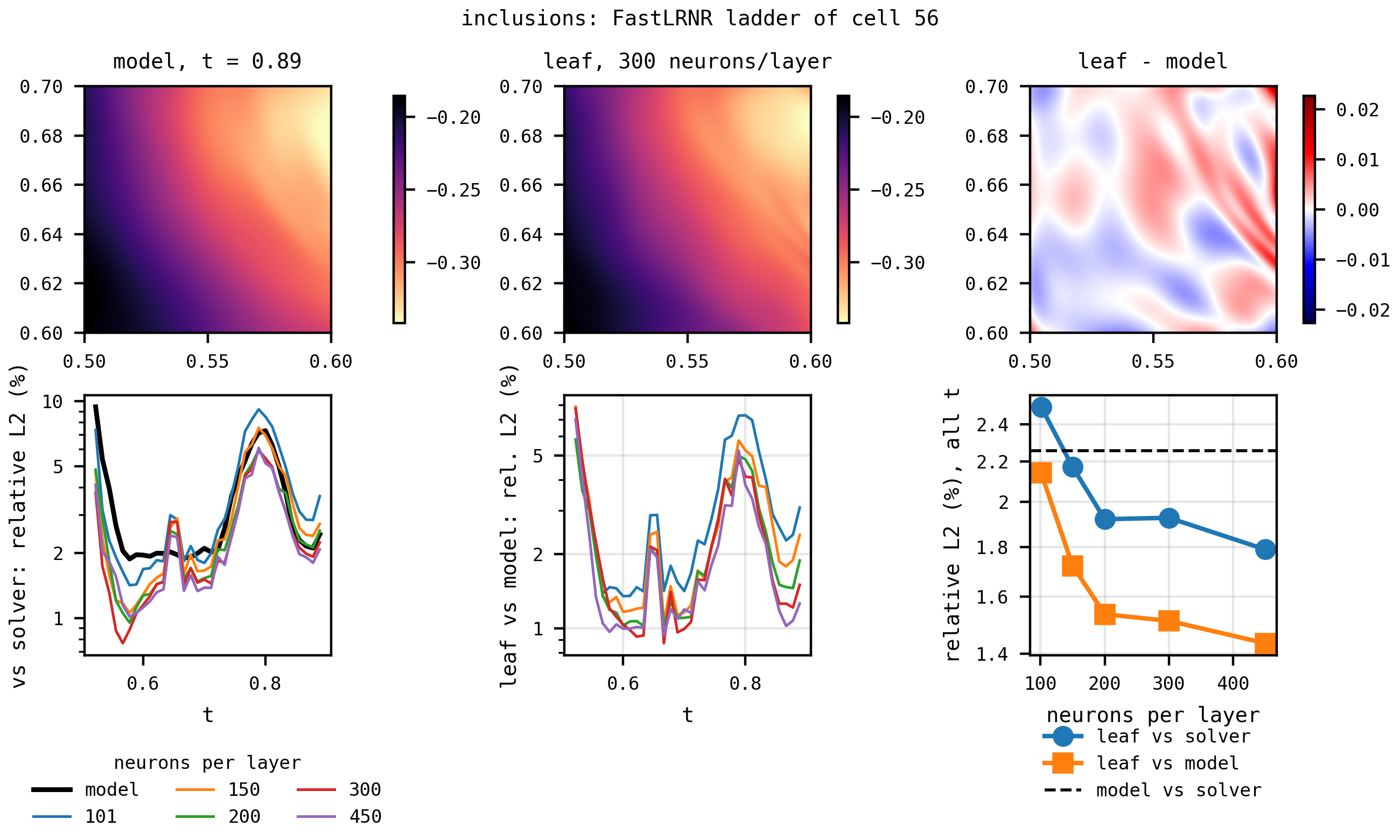}
\caption{As Fig.~\ref{fig:fast_cells_a}, for the shock--bubble and inclusions cells; the
per-time panels of the inclusions cell omit the times at which it is empty.}
\label{fig:fast_cells_c}
\end{figure}

\begin{table}[h]\centering
\begin{adjustbox}{max width=\linewidth}\begin{tabular}{llllll}\toprule
model, layer & keep 45 & 40 & 30 & 20 & 10 \\\midrule
Euler, layer 1 & 0.314 & 0.387 & 0.613 & 0.732 & 0.760 \\
Euler, layer 2 & 0.025 & 0.034 & 0.095 & 0.163 & 0.251 \\
Euler, layer 3 & 0.019 & 0.028 & 0.065 & 0.124 & 0.200 \\
Euler, layer 4 & 0.017 & 0.018 & 0.024 & 0.102 & 0.242 \\
Euler, layer 5 & 0.017 & 0.018 & 0.028 & 0.066 & 0.220 \\
acoustics, layer 1 & 0.931 & 1.059 & 1.117 & 1.133 & 1.038 \\
acoustics, layer 2 & 0.095 & 0.283 & 0.373 & 0.580 & 0.831 \\
acoustics, layer 3 & 0.080 & 0.144 & 0.297 & 0.717 & 0.814 \\
acoustics, layer 4 & 0.015 & 0.016 & 0.163 & 0.540 & 0.678 \\
acoustics, layer 5 & 0.015 & 0.015 & 0.050 & 0.052 & 0.179 \\
advection, layer 1 & 0.781 & 0.801 & 0.892 & 0.884 & 0.871 \\
advection, layer 2 & 0.035 & 0.041 & 0.086 & 0.148 & 0.299 \\
advection, layer 3 & 0.035 & 0.035 & 0.071 & 0.132 & 0.360 \\
advection, layer 4 & 0.035 & 0.035 & 0.037 & 0.110 & 0.320 \\
advection, layer 5 & 0.035 & 0.035 & 0.038 & 0.536 & 0.844 \\
Burgers, layer 1 & 0.123 & 0.695 & 0.806 & 0.779 & 0.777 \\
Burgers, layer 2 & 0.023 & 0.031 & 0.203 & 0.371 & 0.490 \\
Burgers, layer 3 & 0.020 & 0.020 & 0.023 & 0.075 & 0.350 \\
Burgers, layer 4 & 0.020 & 0.020 & 0.021 & 0.040 & 0.340 \\
Burgers, layer 5 & 0.020 & 0.041 & 0.032 & 0.425 & 1.117 \\
shock-bubble, layer 1 & 0.304 & 0.377 & 0.553 & 0.570 & 0.595 \\
shock-bubble, layer 2 & 0.043 & 0.056 & 0.088 & 0.125 & 0.225 \\
shock-bubble, layer 3 & 0.051 & 0.063 & 0.093 & 0.156 & 0.293 \\
shock-bubble, layer 4 & 0.055 & 0.114 & 0.164 & 0.272 & 0.341 \\
shock-bubble, layer 5 & 0.065 & 0.159 & 0.290 & 0.414 & 0.549 \\
inclusions, layer 1 & 0.160 & 0.839 & 0.987 & 0.968 & 0.986 \\
inclusions, layer 2 & 0.018 & 0.019 & 0.052 & 0.214 & 0.518 \\
inclusions, layer 3 & 0.018 & 0.018 & 0.029 & 0.215 & 0.557 \\
inclusions, layer 4 & 0.018 & 0.019 & 0.031 & 0.292 & 0.536 \\
inclusions, layer 5 & 0.018 & 0.021 & 0.049 & 0.553 & 0.963 \\
\bottomrule\end{tabular}\end{adjustbox}
\caption{Layer-wise truncation: relative $L^2$ error when the weakest channels of one layer are dropped and the other layers are left intact.}\label{tab:trunc_layer}\end{table}

\begin{table}[h]\centering
\begin{adjustbox}{max width=\linewidth}\begin{tabular}{lrr|rr|rr|rr}\toprule
 & & & \multicolumn{2}{c|}{Solver-matched, 60k} & \multicolumn{2}{c|}{Model-only, 60k} & \multicolumn{2}{c}{Solver-matched, 15k} \\
Cell & Neurons & Model RMSE & Agree. & RMSE & Agree. & RMSE & Agree. & RMSE \\\midrule
Euler 77 & 101 & 0.0130 & 1.2\% & 0.0264 & 1.3\% & 0.0276 & -- & -- \\
Euler 77 & 150 & 0.0130 & 1.1\% & 0.0216 & 1.0\% & 0.0223 & -- & -- \\
Euler 77 & 200 & 0.0130 & 0.8\% & 0.0180 & 0.7\% & 0.0189 & -- & -- \\
Euler 77 & 300 & 0.0130 & 0.6\% & 0.0154 & 0.5\% & 0.0165 & -- & -- \\
Euler 77 & 450 & 0.0130 & 0.5\% & 0.0136 & 0.4\% & 0.0149 & -- & -- \\
Acoustics 87 & 101 & 0.0026 & 2.6\% & 0.0040 & 2.4\% & 0.0044 & -- & -- \\
Acoustics 87 & 150 & 0.0026 & 2.2\% & 0.0031 & 2.0\% & 0.0037 & -- & -- \\
Acoustics 87 & 200 & 0.0026 & 1.7\% & 0.0026 & 1.5\% & 0.0033 & -- & -- \\
Acoustics 87 & 300 & 0.0026 & 1.5\% & 0.0023 & 1.3\% & 0.0032 & -- & -- \\
Acoustics 87 & 450 & 0.0026 & 1.4\% & 0.0021 & 1.2\% & 0.0030 & -- & -- \\
Advection 45 & 101 & 0.0203 & 7.7\% & 0.0658 & 7.8\% & 0.0700 & 9.2\% & 0.0714 \\
Advection 45 & 150 & 0.0203 & 7.7\% & 0.0679 & 7.4\% & 0.0673 & 6.3\% & 0.0553 \\
Advection 45 & 200 & 0.0203 & 7.2\% & 0.0630 & 6.7\% & 0.0613 & 5.4\% & 0.0460 \\
Advection 45 & 300 & 0.0203 & 9.3\% & 0.0802 & 9.8\% & 0.0872 & 6.5\% & 0.0549 \\
Advection 45 & 450 & 0.0203 & 10.6\% & 0.0914 & 10.4\% & 0.0898 & 6.4\% & 0.0561 \\
Burgers 16 & 101 & 0.0088 & 12.4\% & 0.0309 & 7.7\% & 0.0214 & -- & -- \\
Burgers 16 & 150 & 0.0088 & 27.2\% & 0.0588 & 24.2\% & 0.0525 & -- & -- \\
Burgers 16 & 200 & 0.0088 & 22.1\% & 0.0491 & 19.6\% & 0.0458 & -- & -- \\
Burgers 16 & 300 & 0.0088 & 24.0\% & 0.0538 & 23.8\% & 0.0532 & -- & -- \\
Burgers 16 & 450 & 0.0088 & 24.8\% & 0.0533 & 21.5\% & 0.0488 & -- & -- \\
\bottomrule\end{tabular}\end{adjustbox}
\caption{The fine-tune variants at the held-out times: agreement with the model (relative $L^2$) and error against the solver's points (RMSE) for the solver-matched 60{,}000-step fine-tune of Table~\ref{tab:fast}, the same fine-tune matched to the model only, and the solver-matched fine-tune stopped at 15{,}000 steps (run for the advection cell). Model RMSE is the full model's held-out error on the cell.}\label{tab:fast_variants}\end{table}

\begin{table}[h]\centering
\begin{adjustbox}{max width=\linewidth}\begin{tabular}{llrrrrrr}\toprule
 & & Neurons & \multicolumn{2}{c}{Agreement} & \multicolumn{2}{c}{Leaf RMSE} & \\
\cmidrule(lr){4-5}\cmidrule(lr){6-7}
Cell & Variant & per layer & Frames & Held-out $t$ & Frames & Held-out $t$ & Floats \\\midrule
Euler 77 & Ladder default (over 1.5, rank 67) & 101/101/101/101 & 1.50\% & 1.24\% & 0.0331 & 0.0264 & 59,293 \\
Euler 77 & Over 2.0, rank 50 & 100/100/100/100 & 1.39\% & 1.10\% & 0.0334 & 0.0249 & 58,736 \\
Euler 77 & Over 1.2, rank 83 & 100/100/100/100 & 1.99\% & 1.65\% & 0.0392 & 0.0329 & 58,736 \\
Euler 77 & Fine-tune all (col/row/bias) & 101/101/101/101 & 0.67\% & 0.66\% & 0.0225 & 0.0161 & 59,293 \\
Euler 77 & 150k steps, lr $10^{-4}$ & 101/101/101/101 & 1.34\% & 1.20\% & 0.0308 & 0.0244 & 59,293 \\
Euler 77 & Uneven 150/100/75/75 & 150/101/75/75 & 1.55\% & 1.29\% & 0.0363 & 0.0271 & 59,831 \\
Acoustics 87 & Ladder default (over 1.5, rank 67) & 101/101/101/101 & 2.02\% & 2.60\% & 0.0046 & 0.0040 & 63,083 \\
Acoustics 87 & Over 2.0, rank 50 & 100/100/100/100 & 1.93\% & 2.37\% & 0.0039 & 0.0035 & 62,488 \\
Acoustics 87 & Over 1.2, rank 83 & 100/100/100/100 & 2.26\% & 3.00\% & 0.0048 & 0.0045 & 62,488 \\
Acoustics 87 & Fine-tune all (col/row/bias) & 101/101/101/101 & 1.50\% & 2.34\% & 0.0031 & 0.0034 & 63,083 \\
Acoustics 87 & 150k steps, lr $10^{-4}$ & 101/101/101/101 & 1.74\% & 2.29\% & 0.0037 & 0.0032 & 63,083 \\
Acoustics 87 & Uneven 150/100/75/75 & 150/101/75/75 & 1.81\% & 2.49\% & 0.0041 & 0.0036 & 62,688 \\
\bottomrule\end{tabular}\end{adjustbox}
\caption{Spending about 100 neurons per layer on the Euler and acoustics cells: six reductions (EIM oversampling factor and basis rank, what the fine-tune touches, its length, and an uneven allocation), with the agreement with the model and the error against the solver at the sampled frames and at the 20 held-out times. The model's held-out RMSE is 0.0130 (Euler) and 0.0026 (acoustics).}\label{tab:fast100}\end{table}

\begin{table}[h]\centering
\begin{adjustbox}{max width=\linewidth}\begin{tabular}{llccccccc}\toprule
 & & \multicolumn{5}{c}{20{,}000 steps, learning rate} & \multicolumn{2}{c}{400{,}000 steps, learning rate} \\
\cmidrule(lr){3-7}\cmidrule(lr){8-9}
Model & Sampler & $10^{-4}$ & $3\times10^{-4}$ & $10^{-3}$ & $3\times10^{-3}$ & $10^{-2}$ & $10^{-4}$ & $3\times10^{-4}$ \\\midrule
Hyper-SIREN & Global, adaptive & 21.9\% & 15.6\% & 99.5\% & 97.1\% & div. & 5.8\% & 5.1\% \\
Hyper-SIREN & Global, uniform & 14.6\% & 14.0\% & 95.6\% & 95.6\% & div. & 4.8\% & 5.6\% \\
Hyper-SIREN & Local, adaptive & 78.3\% & 82.5\% & 96.2\% & div. & div. & 13.7\% & 96.2\% \\
Hyper-SIREN & Local, uniform & 79.5\% & 95.7\% & div. & div. & div. & 21.7\% & -- \\
DeepONet & Global, adaptive & 83.6\% & 79.1\% & 82.8\% & div. & div. & -- & -- \\
DeepONet & Global, uniform & 81.4\% & 75.2\% & 75.4\% & div. & div. & 50.2\% & 44.8\% \\
DeepONet & Local, adaptive & 86.0\% & 90.5\% & 91.6\% & div. & div. & 74.9\% & 75.0\% \\
DeepONet & Local, uniform & 85.7\% & 89.1\% & 90.2\% & div. & div. & 73.7\% & 74.5\% \\
FNO & Global, adaptive & 27.0\% & 16.6\% & 15.6\% & 15.4\% & 15.5\% & -- & -- \\
FNO & Global, uniform & 31.4\% & 16.1\% & 15.6\% & 15.4\% & div. & 15.4\% & 15.4\% \\
FNO & Local, adaptive & 77.2\% & 38.7\% & 21.2\% & 42.3\% & 95.7\% & 15.9\% & 15.5\% \\
FNO & Local, uniform & 78.0\% & 39.8\% & 20.4\% & 25.9\% & 95.7\% & 16.2\% & 15.5\% \\
\midrule
\lrnr{} & Local, adaptive & 29.0\% & 63.6\% & 95.9\% & 95.9\% & div. & -- & -- \\
\lrnr{} & Local, uniform & 39.1\% & 77.8\% & 95.7\% & 95.7\% & div. & -- & -- \\
\bottomrule\end{tabular}\end{adjustbox}
\caption{The runs of Table~\ref{tab:baselines} at the 80 times halfway between the training frames (Table~\ref{tab:heldout}): relative $L^2$ error (area-weighted, root mean square over the 80 times). The \lrnr{}'s ladder reaches 3.1\% after 389{,}600 steps and 1.8\% at the end.}\label{tab:baselines_heldout}\end{table}

\begin{table}[h]\centering
\begin{adjustbox}{max width=\linewidth}\begin{tabular}{r*{7}{r}}\toprule
 & \multicolumn{4}{c}{Adaptive sampling} & \multicolumn{3}{c}{Uniform sampling} \\
\cmidrule(lr){2-5}\cmidrule(lr){6-8}
 & \multicolumn{2}{c}{Global} & \multicolumn{2}{c}{Local} & \multicolumn{2}{c}{Global} & Local \\
\cmidrule(lr){2-3}\cmidrule(lr){4-5}\cmidrule(lr){6-7}\cmidrule(lr){8-8}
Width & $10^{-4}$ & $3\times10^{-4}$ & $10^{-4}$ & $3\times10^{-4}$ & $10^{-4}$ & $3\times10^{-4}$ & $10^{-4}$ \\\midrule
48 & 5.8\% & 5.1\% & 13.7\% & 96.2\% & 4.8\% & 5.6\% & 21.7\% \\
96 & 2.8\% & 2.4\% & 9.0\% & 96.0\% & 2.8\% & 2.6\% & 11.6\% \\
192 & 1.5\% & 1.3\% & 6.1\% & 95.8\% & 1.3\% & 0.9\% & 6.4\% \\
384 & 0.9\% & 1.0\% & 4.9\% & 95.8\% & 1.1\% & 0.7\% & 6.8\% \\
\bottomrule\end{tabular}\end{adjustbox}
\caption{The runs of Table~\ref{tab:bl_width} at the 80 times halfway between the training frames (Table~\ref{tab:heldout}): relative $L^2$ error (area-weighted, root mean square over the 80 times).}\label{tab:bl_width_heldout}\end{table}

\begin{table}[h]\centering
\begin{adjustbox}{max width=\linewidth}\begin{tabular}{r*{6}{r}}\toprule
 & \multicolumn{3}{c}{Adaptive sampling} & \multicolumn{3}{c}{Uniform sampling} \\
\cmidrule(lr){2-4}\cmidrule(lr){5-7}
Width & Local & No projector & Global & Local & No projector & Global \\\midrule
250 & 31.0\% & 14.1\% & 2.3\% & 71.1\% & 71.6\% & 2.6\% \\
500 & 5.5\% & 5.4\% & 1.2\% & 11.1\% & 14.1\% & 1.1\% \\
1000 & 3.9\% & 4.0\% & 0.9\% & 5.8\% & 5.6\% & 0.9\% \\
2000 & 3.4\% & 3.0\% & 0.9\% & 5.1\% & 5.1\% & 0.8\% \\
\bottomrule\end{tabular}\end{adjustbox}
\caption{The runs of Table~\ref{tab:lrnr_width} at the 80 times halfway between the training frames (Table~\ref{tab:heldout}): relative $L^2$ error (area-weighted, root mean square over the 80 times). Global: the learning rate of Table~\ref{tab:lrnr_width}.}\label{tab:lrnr_width_heldout}\end{table}

\clearpage
\end{document}